\PassOptionsToPackage{table,dvipsnames}{xcolor}
\PassOptionsToPackage{colorlinks=true, linkcolor=blue, citecolor=blue, urlcolor=blue}{hyperref}

\documentclass[iicol]{sn-jnl}

\usepackage[most]{tcolorbox}
\usepackage{graphicx}%
\usepackage{multirow}%
\usepackage{amsthm}%
\usepackage{mathrsfs}%
\usepackage[title]{appendix}%
\usepackage{textcomp}%
\usepackage{manyfoot}%
\usepackage{algorithm}%
\usepackage{algorithmicx}%
\usepackage{algpseudocode}%
\usepackage{listings}%
\usepackage{subcaption}
\usepackage{tikz}
\usepackage{array}

\usepackage{threeparttable}
\tcbset{
  mydefinition/.style={
    colback=gray!10,    % background color
    colframe=black!70,  % frame color
    coltitle=black,     % title color
    rounded corners,
    boxrule=0.4pt,      % border line thickness
    fonttitle=\bfseries,
    left=4pt,
    right=4pt,
    top=4pt,
    bottom=4pt,
    before skip=6pt,
    after skip=6pt,
    width=\columnwidth, % fit to one column
    boxsep=0pt
  }
}

\usepackage[font=small]{caption}
\usepackage{amsmath,amsfonts}
\usepackage{subcaption}
\usepackage{stfloats}
\usepackage{url}
\usepackage{verbatim}
\usepackage{booktabs}
\usepackage{multirow}
\usepackage{stfloats}
\usepackage{tikz}
\usepackage{adjustbox}
\usepackage{tcolorbox}
\usepackage{graphicx}
\tcbuselibrary{skins}
\usetikzlibrary{shadows, shapes, positioning}
\usepackage{xcolor}  
\usepackage[colorlinks=true, linkcolor=blue, citecolor=blue, urlcolor=blue]{hyperref}
\usepackage{float}
\usepackage{svg}
\definecolor{wine-stain}{rgb}{0.5,0,0}
\hypersetup{
  colorlinks,
  linkcolor=wine-stain,
  linktoc=all
}
\usepackage{refcount}  % enables \getrefnumber
\definecolor{ForrestGreen}{RGB}{34, 139, 34}
\definecolor{MagentaPink}{RGB}{255, 0, 144}
\definecolor{darkteal}{RGB}{21,96,130}

\newcommand{\legendbox}[1]{%
  \begingroup
    \setlength{\fboxsep}{0pt}%
    \color{black}% reset
    \raisebox{0.25ex}{\colorbox{#1}{\phantom{\rule{1.2em}{1.2ex}}}}%
  \endgroup
}
\newcolumntype{M}{>{\arraybackslash}m{0.85cm}}

\definecolor{terminalbg}{HTML}{1E1E26}  % same feel as VS-Code / many terminals

\newtcolorbox{terminalbox}{
  enhanced,
  colback=terminalbg,      % background
  colframe=terminalbg,     % no visible border
  coltext=white,           % text colour
  boxrule=0pt,             % remove frame line
  left=4pt,right=4pt,top=4pt,bottom=4pt,  % inner padding
  sharp corners,
  fontupper=\ttfamily\scriptsize,         % small monospace
  width=\linewidth,                       % full-width box
  breakable=false,                        % keep your own line breaks
}

\usepackage{soul}
\definecolor{sg}{HTML}{00ff7f}
\definecolor{lb}{HTML}{c9e9ed} % Adjusted to a recognizable code
\definecolor{lg}{HTML}{9bfaa8}
\theoremstyle{thmstyleone}%
\theoremstyle{thmstyletwo}%

\theoremstyle{thmstylethree}%
\newtheorem{definition}{Definition}%

\begin{document}

\title{ \centering
    Multi-Robot Navigation in Social Mini-Games:\\ Definitions, Taxonomy, and Algorithms
}

\author*[1]{\fnm{Rohan} \sur{Chandra}}\email{rohanchandra@virginia.edu}

\author[2]{\fnm{Shubham} \sur{Singh}}\email{singh281@utexas.edu}

\author[3]{\fnm{Wenhao} \sur{Luo}}\email{wenhao@uic.edu}

\author[4]{\fnm{Katia} \sur{Sycara}}\email{sycara@andrew.cmu.edu}

\affil*[1]{\orgdiv{Dept. of Computer Science}, \orgname{University of Virginia}}
\affil[2]{\orgdiv{Dept. of Aerospace Engineering \& Engg. Mechanics}, \orgname{The University of Texas at Austin}}
\affil[3]{\orgdiv{Dept. of Computer Science}, \orgname{University of Illinois Chicago}}
\affil[4]{\orgdiv{Robotics Institute}, \orgname{Carnegie Mellon University}}


\abstract{The ``Last Mile Challenge'' has long been considered an important, yet unsolved, challenge for autonomous vehicles, public service robots, and delivery robots. A central issue in this challenge is the ability of robots to navigate constrained and cluttered environments that have high agency (e.g., doorways, hallways, corridor intersections), often while competing for space with other robots and humans. We refer to these environments as ``Social Mini-Games'' (SMGs). Traditional navigation approaches designed for MRN do not perform well in SMGs, which has led to focused research on dedicated SMG solvers. However, publications on SMG navigation research make different assumptions (on centralized versus decentralized, observability, communication, cooperation, etc.), and have different objective functions (safety versus liveness). These assumptions and objectives are sometimes implicitly assumed or described informally. This makes it difficult to establish appropriate baselines for comparison in research papers, as well as making it difficult for practitioners to find the papers relevant to their concrete application. Such ad-hoc representation of the field also presents a barrier to new researchers wanting to start research in this area. SMG navigation research requires its own taxonomy, definitions, and evaluation protocols to guide effective research moving forward. This survey is the first to catalog SMG solvers using a well-defined and unified taxonomy and to classify existing methods accordingly. It also discusses the essential properties of SMG solvers, defines what SMGs are and how they appear in practice, outlines how to evaluate SMG solvers, and highlights the differences between SMG solvers and general navigation systems. The survey concludes with an overview of future directions and open challenges in the field. Our project is open-sourced at \href{https://socialminigames.github.io/}{\textbf{https://socialminigames.github.io/}}.}

\keywords{Multi-Robot Navigation, Deadlocks, Social Navigation}

\maketitle

\section{Introduction}
\label{sec:Intro}

Multi-robot navigation (MRN) is an active and essential area of research in robotics and artificial intelligence with wide-ranging applications such as autonomous driving, warehouse logistics, swarms, delivery, service, and personalized home robots~\cite{drew2021multi,rasheed2022review,loizou2014multi,nathan2011review,orr2023multi,antonyshyn2023multiple,chen2021decentralized,raibail2022decentralized,huang2019collision,patwardhan2022distributing}. MRN builds upon the challenges of single-robot navigation by introducing the complexities of inter-robot, human-robot interactions, and collective decision-making. 
In multi-robot systems, the algorithms must account for the intentions, movements, and potential conflicts among multiple robots, including humans, making the navigation problem significantly more complex \cite{mavrogiannis2017socially, fan2020distributed, alzetta2020time}. 
\begin{figure*}[t]
    \centering
\begin{minipage}[t]{0.48\textwidth}
    \centering
    \includegraphics[height=5cm, width = \linewidth]{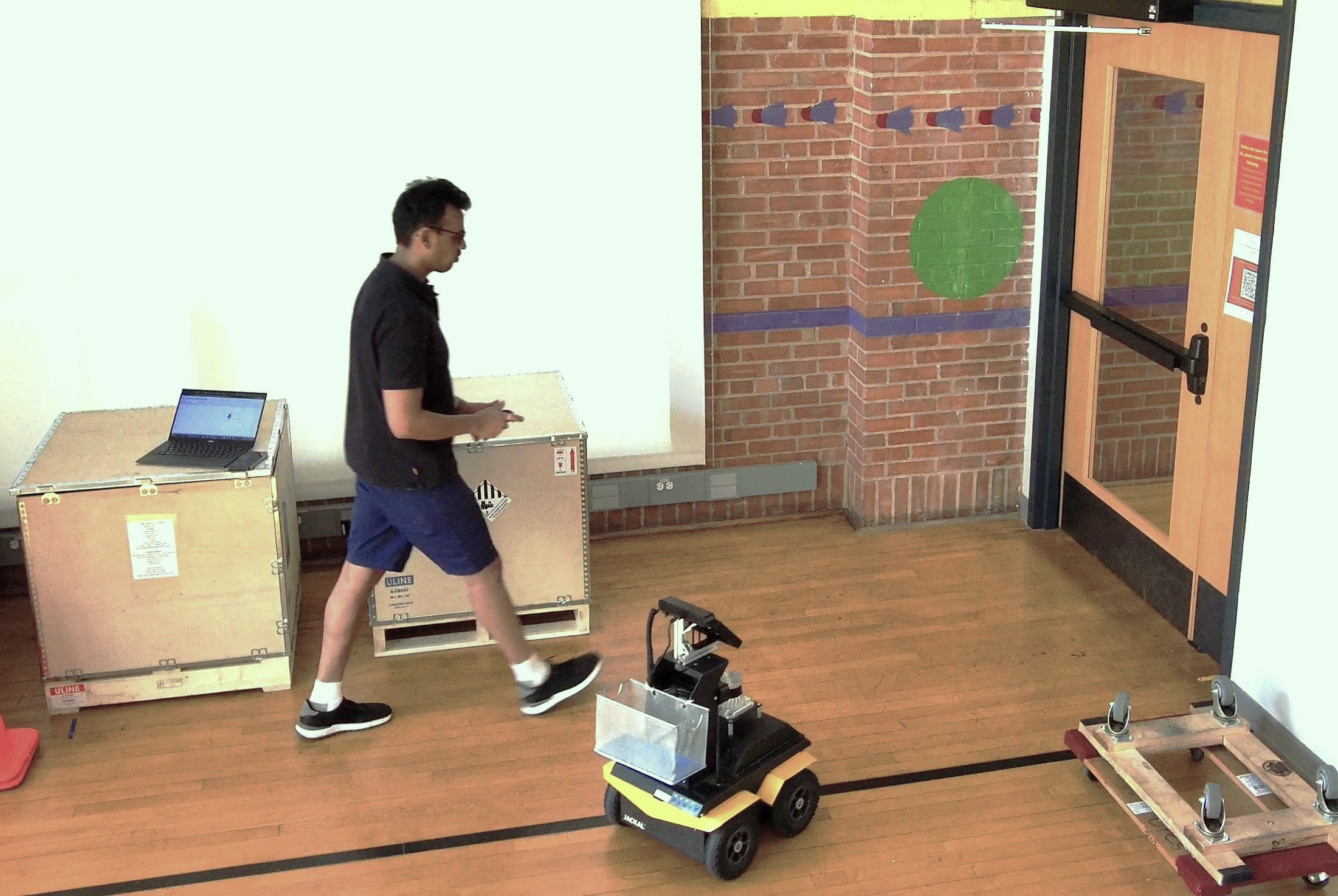}
    \subcaption{A Social Mini-Game (SMG) (taken with permission from~\cite{chandra2025deadlock}).}
    \label{fig:smg_mrn_a}
\end{minipage}
    \begin{minipage}[t]{0.48\textwidth}
        \centering
        \includegraphics[height=5cm, width = \linewidth]{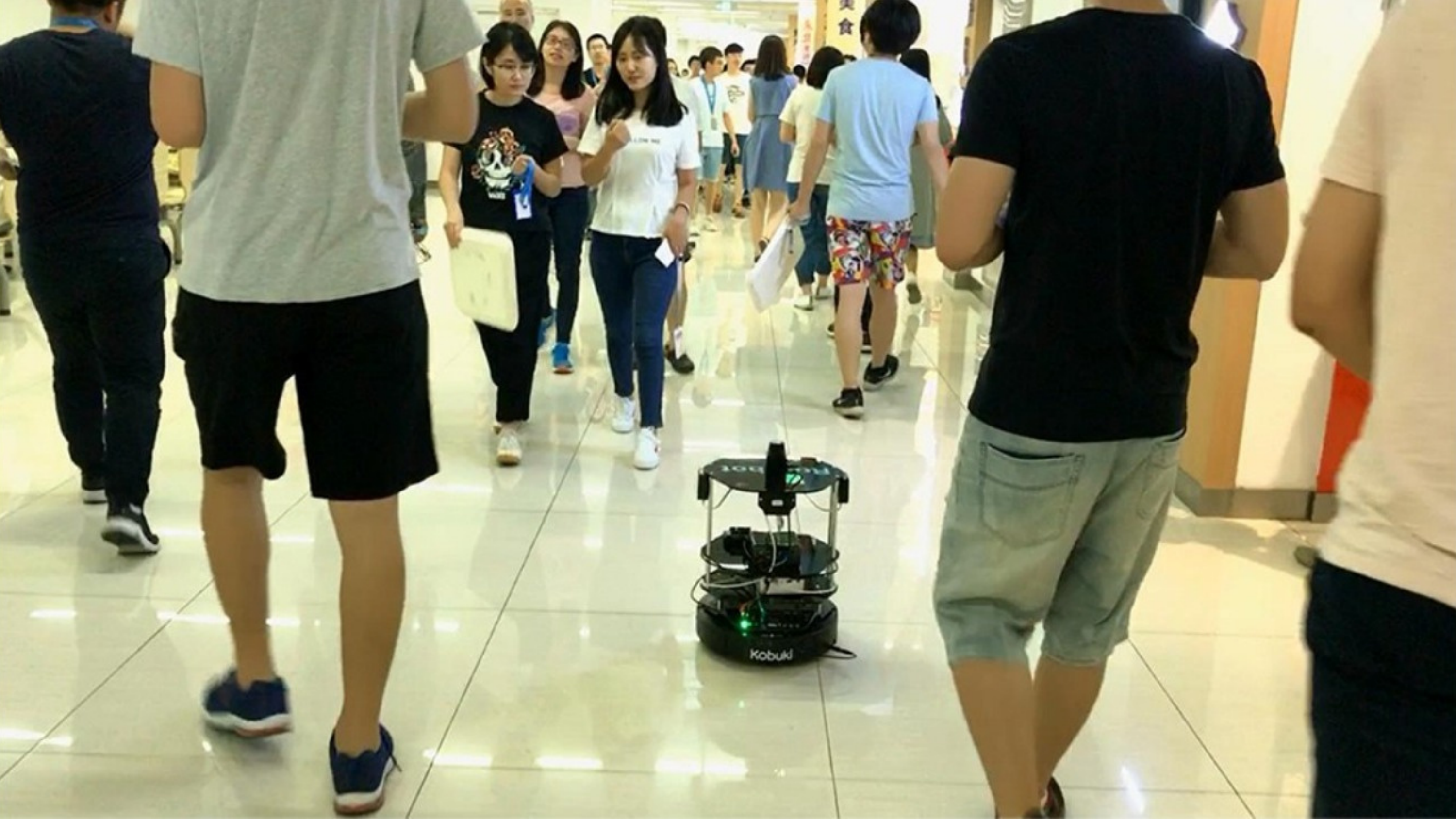}
        \subcaption{General multi-robot navigation (taken with permission from~\cite{fan2018crowdmove}).}
        \label{fig:smg_mrn_b}
    \end{minipage}
    \caption{Social Mini-Games \textit{(left)} differ from general dense navigation scenarios \textit{(right)} in that SMGs are more physically and geometrically constrained and robots (and humans) often compete for shared space.}
    \label{fig:smg_mrn}
\end{figure*}
In recent years, the research field of MRN has started to coincide with the momentum of robot deployments in shared human spaces, largely driven by advances in data, compute, and foundation models. As a consequence, research has flourished in MRN, and more specifically, MRN in constrained environments that arise in shared human spaces. In Figure~\ref{fig:smg_mrn}, we highlight two types of MRN in constrained scenarios. On the left, we depict {Social Mini-Games} (SMGs) \cite{chandra2025deadlock, chandra2024towards} on the right we depict ``general MRN''. A key distinguishing factor between these scenarios is the notion of \textit{agency}--that is, the capacity of an agent to take independent actions and influence the environment around them. \textbf{SMGs exhibit high agency} where small actions can significantly change the interaction dynamics, coordination requirements, and potential for deadlocks. General MRN, on the other hand, exhibit low to medium agency. 
High-agency SMGs have several unique challenges. First, without explicit scheduling, robots result in deadlocks, collisions, or non-smooth trajectories \cite{kandhasamy2020scalable,qian2015decentralized}. Second, humans can avoid collisions and deadlocks without having to deviate too much from their nominal speed or trajectory. For instance, when two individuals go through a doorway together, one person modulates their velocity by just enough to enable the other person to pass through first, while still adhering closely to their preferred speed. This type of behavior presents a significant challenge for robots, especially in high-agency SMGs.
Classical navigation approaches typically fail to cover all the aspects needed for SMGs~\cite{chandra2025deadlock}. So, new techniques, ideas, and systems are being investigated, often with roots deep within the larger parent field of MRN. This fact presents the crux of the issue: researchers studying SMG solvers from a MAPF perspective will build off of MAPF research, MARL-based SMG solvers will form their roots from the MARL community, and so on. Other communities build upon game theory and optimization. Each field has its own set of assumptions, limitations, objectives, and techniques. As a result, MRN in SMG research has fractured into these sub-communities without a clear, unified formulation. This categorization into sub-communities presents an entry barrier to new researchers in this field by making it difficult to navigate through and understand existing literature and to establish appropriate baselines for comparison. In addition, this makes it difficult for practitioners to find papers and solutions relevant to their concrete application.
This article aims to address this growing challenge by introducing a unified taxonomy to describe MRN in SMG problems, and by establishing common assumptions, properties, and measures for evaluating MRN in SMG algorithms. The unified taxonomy we present in this paper is our attempt to classify the currently studied SMG solver variants. We hope this terminology will serve as a common ground for future researchers, and will be used by them to describe their contributions succinctly and accurately.
\subsection{Differentiation from Existing Surveys}
Our survey focuses on multi-robot navigation in \textit{social mini-games}, and thus belongs to the broader field of social robot navigation. Social robot navigation, however, is vast and we will not attempt to summarize it; instead, we refer to many recent surveys on social navigation \cite{kruse2013human,rios2015proxemics,chik2016review,charalampous2017recent,cheng2018autonomous,gao2021evaluation,mavrogiannis2021core,mirsky2021prevention,moller2021survey,wang2022metrics}. 
Among these, \cite{gao2021evaluation} focuses on \textit{evaluating} social robot navigation algorithms, reviewing 177 recent papers to gather evaluation methods, scenarios, datasets, and metrics, using their findings to discuss shortcomings of existing research and to make recommendations on how to properly evaluate social navigation research. Another recent survey by Mavrogiannis et al.~\cite{mavrogiannis2021core} focuses on the core \textit{challenges} of social navigation with respect to different algorithms, human behavior models, and evaluation, but they focus on the general navigation case.
% Our work builds on the works of \cite{gao2021evaluation} and \cite{mavrogiannis2021core} and similar surveys to map the field. We contribute a crisp definition of social robot navigation based on discussions held at the 2022 Social Navigation Symposium, an overview of methodologies for research, and a taxonomy of the field which we use to examine existing metrics, scenarios, benchmarks, datasets, and simulators, and shared principles to make social navigation evaluations comparable across the community. 
Wang et al.~\cite{wang2022metrics} proposes new \textit{metrics} evaluating the principles defined in \cite{kruse2013human}, such as {comfort}, {naturalness}, and {sociability}.
% We expand the principles in \cite{kruse2013human}, and propose a lifecycle of social navigation with recommendations for metrics, scenarios, benchmarks, datasets, and simulators, along with guidelines for metric usage.
And finally, datasets~\cite{scanddataset, martin2019jrdb} and benchmarks~\cite{kastner2022arena, sprague2023socialgym} have enabled researchers to compare algorithms, revealing limitations of existing solutions, and illuminating promising new directions.  However, all of the surveys discussed above on various aspects of social robot navigation focus on robots in sparse or dense crowds, but not social mini-games. We showcase the difference in Figure~\ref{fig:smg_mrn}.

Perhaps the closest survey that mirrors the same objective as with this survey (that is, attempting to unify a growing, but fractured, research area) is the seminal survey on MAPF (``Multi-Agent Pathfinding: Definitions, Variants, and Benchmarks'') by Stern et al.~\cite{stern2019multi}. Here, the authors present a definition for MAPF, describe the different variants of it, and introduce two benchmarks for proper evaluation. This paper was influential in anchoring and streamlining the future research in MAPF. We envision a similar success via our survey. 
The overall structure of this survey is given in Fig.~\ref{fig:overarching}. We first define the SMG problem in Sec.~\ref{sec:scenarios}, highlight its various scenarios where it appears, and how to evaluate navigation approaches in SMGs. Then in Sec.~\ref{sec:Taxonomy}, we introduce the taxonomy, properties, and algorithmic categories  for research on SMG solvers. Then in Sec.~\ref{sec: Evaluation}, we benchmark and compare some recent state of the art SMG solvers on the scenarios introduced earlier. Finally in Sec.~\ref{sec:trends}, we conclude by discussing general trends, open problems, and future directions.
\begin{figure}[htbp]
    \centering
    \adjustbox{width=\columnwidth,center}{%
        \begin{tikzpicture}[
            font=\rmfamily,
        font=\rmfamily,
        box/.style={
            draw, rounded corners, minimum width=1.5cm, minimum height=1.0cm,
            align=left, fill=none, inner sep=3pt,
            font = \small 
        },
        circ/.style={
            draw, circle, minimum size=1.8cm, align=center,
            fill=darkteal, text=white
        },
        line/.style={
            thick, color=darkteal
        },
        every node/.style={scale=1.0}
    ]
    % Center node
    \node[circ] (mrn) at (0, 0) {\textbf{MRN}\\in\\SMGs};
    % Radius
    \def\r{3.0}
    % --- UPDATED POSITIONS FOR SYMMETRY ---
    % There are now 6 boxes, so they are placed 60 degrees apart.
    % Taxonomy (top)
    \node[box] (taxonomy) at ({\r*cos(90)},  {\r*sin(90)}) {
    \textbf{Taxonomy~\ref{subsec: taxonomy}}\\[-1pt]
        $\bullet$ Coordination\\
        $\bullet$ Communication\\
        $\bullet$ Deadlock Handling\\
        $\bullet$ Invasiveness\\
        $\bullet$ Cooperation\\
        $\bullet$ Observability
    };
    % --- NEW NODE: ALGORITHMS ---
    \node[box] (algorithms) at ({1.1*\r*cos(160)}, {1.1*\r*sin(160)}) {
        \textbf{Algorithms~\ref{sec:algorithms}}\\[-1pt]
        $\bullet$ MARL\\
        $\bullet$ MAPF\\
        $\bullet$ Optimization\\
        $\bullet$ Others
    };
    \node[box] (types) at ({\r*cos(200)}, {\r*sin(200)}) {
        \textbf{SMGs~\ref{sec:scenarios}}\\[-1pt]
        $\bullet$ MRN Preliminaries \\
        $\bullet$ MRN to SMGs \\
        $\bullet$ Scenarios\\
        $\bullet$ Evaluation Metrics
    };
    \node[box] (trends) at ({\r*cos(270)}, {\r*sin(270)}) {
        \textbf{General Trends~\ref{sec:trends}}\\[-1pt]
        $\bullet$ Visual Inputs\\
        $\bullet$ With Humans\\
        $\bullet$ Using Digital Twins
    };
    % Evaluation (lower right)
    \node[box] (eval) at ({1.05*\r*cos(335)}, {1.05*\r*sin(335)}) {
        \textbf{Evaluation~\ref{sec: Evaluation}}\\[-1pt]
        $\bullet$ Methods
        % $\bullet$ SMGLib\\
        % $\bullet$ Comparative Analysis
    };
    % Types (upper right)
\node[box] (props) at ({1.25*\r*cos(15)}, {1.25*\r*sin(15)}) {        \textbf{Properties~\ref{subsec: properties}}\\[-1pt]
        $\bullet$ Safety\\
        $\bullet$ Liveness\\
        $\bullet$ Welfare Maximization\\
        $\bullet$ Social Compliance\\
        $\bullet$ Scalability
    };
    % Connections
    \draw[line] (mrn) -- (taxonomy);
    \draw[line] (mrn) -- (algorithms); % Connection for the new node
    \draw[line] (mrn) -- (props);
    \draw[line] (mrn) -- (trends);
    \draw[line] (mrn) -- (eval);
    \draw[line] (mrn) -- (types);
    \end{tikzpicture}
    }
    \caption{MRN in SMGs Survey Structure}
    \label{fig:overarching}
\end{figure}

\section{Social Mini-Games}
\label{sec:scenarios}

\subsection{MRN Preliminaries}
\label{subsec: MRN_prelim}
{{We first begin by defining a general MRN as a partially observable stochastic game: $\left \langle k, \mathcal{X}, \{\Omega^i\}, \{\mathcal{O}^i\},\{\mathcal{U}^i\}, \mathcal{T},  \{\widetilde\Gamma^i\}, \{\mathcal{J}^i\}\right\rangle$ where $k$ denotes the number of robots. Hereafter, $i$ will refer to the index of a robot and appear as a superscript whereas $t$ will refer to the current time-step and appear as a subscript. The general state space $\mathcal{X}$ (\textit{e.g.} SE(2), SE(3), etc.) is continuous; the $i^\textrm{th}$ robot at time $t$ has a state $x^i_t\in \mathcal{X}$. A state $x^i_t$ consists of both visible parameters (\textit{e.g.} current position, linear and angular velocity and hidden (to other agents) parameters which could refer to the internal state of the robot such as preferred speed, preferred heading, etc. We denote the set of observable parameters as $\overline{x}^i_t$. On arriving at a current state $x^i_t$, each robot generates a local observation,  $o^i_t \in \Omega^i$, via $\mathcal{O}^i: \mathcal{X}\longrightarrow \Omega^i$, where $ \mathcal{O}^i\left(x^i_t\right) = \left\{x^i_t\right\} \cup \left\{\overline{x}^j_t: j\in \mathcal{N}^i\left( x^i_t\right)\right\}$, the set of robots detected by $i$'s sensors. Over a finite horizon $T$, each robot is initialized with a start state $x^i_{0} \in \mathcal{X}_I$, a goal state $x^i_N\in \mathcal{X}_g$ where $\mathcal{X}_I$ and $\mathcal{X}_g$ denote subsets of $\mathcal{X}$ containing the initial and final states. The transition function is given by $\mathcal{T}:\mathcal{X}\times \mathcal{U}^i \longrightarrow \mathcal{X}$, where $\mathcal{U}^i$ is the continuous control space for robot $i$ representing the set of admissible inputs for $i$. A discrete trajectory is specified as $\Gamma^i = \left( x^i_{0}, x^i_1, \ldots, x^i_T  \right)$ and its corresponding input sequence is denoted by $\Psi^i = \left( u^i_{0}, u^i_1, \ldots, u^i_{T-1}  \right)$. Robots follow the discrete-time control-affine system, $x^i_{t+1} = f\left(x^i_t\right) + g\left(x^i_t\right)u^i_t$. Let $\pi^i$ be a policy that maps the observation or state history of agent $i$ to an action}}

% , which is obtained after applying a Runge-Kutta discretization scheme to a continuous-time system describing their motion (derived from first principles) which takes the following form:
% \begin{equation}
% % x^i(t) = f_c\left(x^i(t)\right) + g_c\left(x^i(t)\right)u^i(t),
% x^i(t) = f_c\left(x^i(t)\right) + g_c\left(x^i(t)\right)u^i(t),
%     \label{eq:CTcontrol_affine_dynamics}
% \end{equation}
% where $f_c$ and $g_c$ are locally Lipschitz continuous functions (note that in general, the state and input vectors of the continuous-time and the discrete-time systems that correspond to the same time instant are not  the same due to discretization induced errors). 
% In this paper, we will mainly utilize the discrete-time model~\eqref{eq: control_affine_dynamics}, especially for control design purposes (these methods will rely on optimization techniques), but we will also refer to the continuous-time model \eqref{eq:CTcontrol_affine_dynamics} for analysis purposes. We will also assume that the continuous-time system~\eqref{eq:CTcontrol_affine_dynamics} is small-time controllable, which means that set of points reachable from $x^i(t)$ within the time interval $[t,t^\prime]$ will contain a neighborhood of $x^i(t)$ for any $t^\prime>t$~\cite{laumond2005guidelines}. Small-time controllability allows the following result to hold true:

{We denote by $\widetilde \Gamma^i$ as the set of \textit{preferred} trajectories for robot $i$ that solve the two-point boundary value problem. A preferred trajectory, as defined by existing methods~\mbox{\cite{impc, orca, nh-orca}}, refers to a collision-free path a robot would follow in the absence of dynamic or static obstacles and is generated via a default planner according to some predefined criteria such as shortest path, minimum time, etc. A collision is defined as follows. Let $\mathcal{C}^i\left( x^i_t \right) \subseteq \mathcal{X}$ represent the space occupied by robot $i$ (as a subset of the state space $\mathcal{X}$) at any time $t$ which can be approximated by the convex hull of a finite number of points that determine the boundary of the robot (e.g., vertices of a polytopic set). Then, robots $i$ and $j$ are said to collide at time $t$ if $\mathcal{C}^i\left(x^i_t \right) \cap \mathcal{C}^j\left(x^j_t \right) \neq \emptyset$. Each robot has a running cost $\mathcal{J}^i:\mathcal{X} \times  \mathcal{U}^i \longrightarrow \mathbb{R}$ that assigns a cost to a state-input pair $\left(x^i_t, u^i_t\right)$ at each time step based on user-defined variables \textit{e.g.} distance of the robots current position from the goal, change in the control across subsequent time steps, and distance between the robots preferred and actual paths.} 

\subsection{From General MRN to SMGs}
\label{subsec: mrn_to_smg}
SMGs are a special class of MRN scenarios. Every SMG is an instance of an MRN, but the other way around is not true. Informally, SMGs arise when all the optimal or preferred trajectories of two or more agents collide within a time interval. We make this formal via the definition below.

\begin{tcolorbox}[mydefinition]
\begin{definition}
{A \textbf{social mini-game}~\cite{chandra2025deadlock} occurs if for some $\delta > 0$ and integers $a,b \in (0,T)$ with $ b-a > \delta$, there exists at least one pair $i,j, i\neq j$ such that for all $\Gamma^i \in \widetilde \Gamma^i, \Gamma^j \in \widetilde \Gamma^j$, we have $\mathcal{C}^i\left( x^i_t \right) \cap \mathcal{C}^j\left( x^j_t \right)\neq \emptyset \ \forall \ t \in [a,b]$, where $x^i_t, x^j_t$ are elements of $\Gamma^i$ and $\Gamma^j$.
}    
\end{definition}
\end{tcolorbox}

\begin{figure}[t]
\centering
\includegraphics[width=\columnwidth]{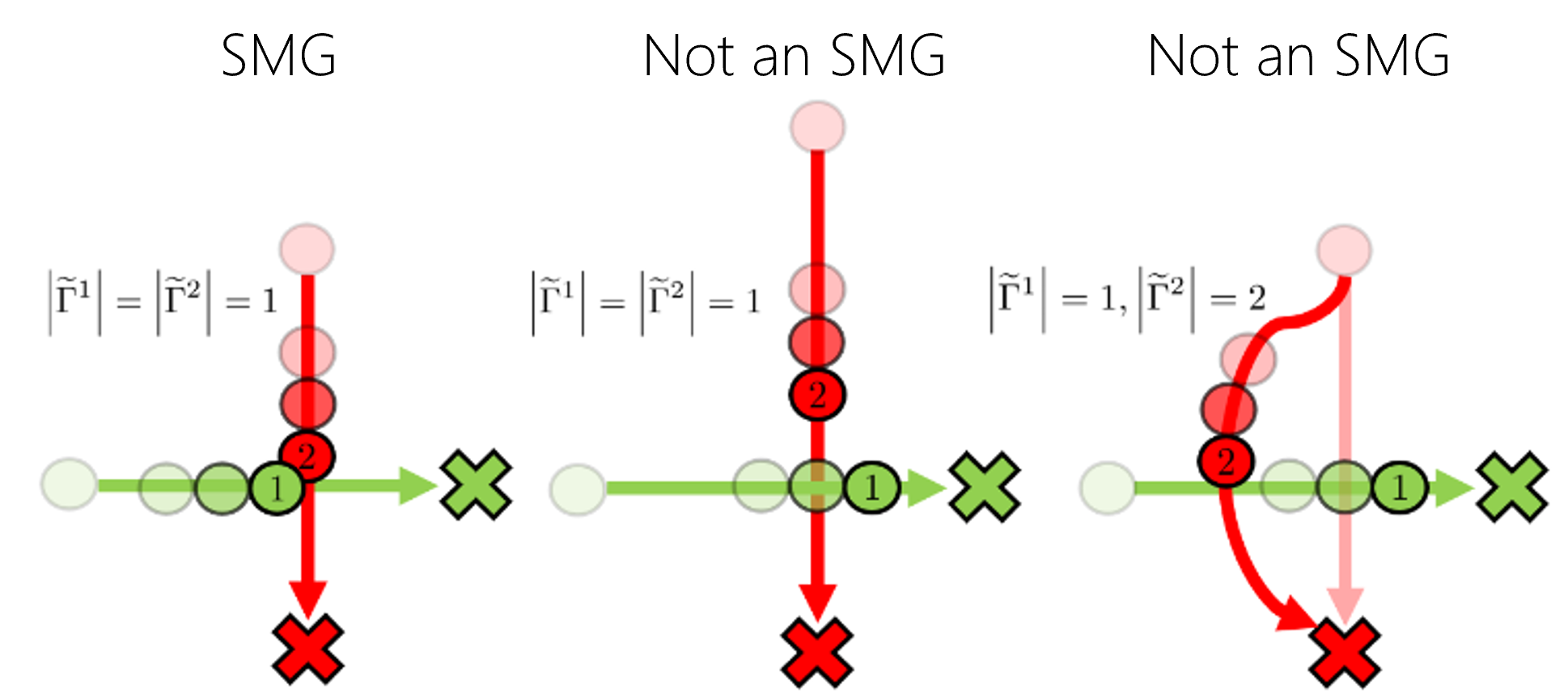}
\caption{\textbf{Examples/Counter-examples of social mini-games:} Arrows indicate the direction of motion for two agents $1$ and $2$ toward their goals marked by the corresponding cross. The first scenario is a social mini-game since both the preferred trajectories of agents $1$ and $2$ are in conflict from some $t=a$ to $t=b$ where $b-a \geq \delta$. The second and third scenarios are \textit{not} social mini-games as there are no conflicts. In the second scenario, there is no duration where agents intersect one another. In the third scenario, agent $2$ has an alternate conflict-free preferred trajectory to fall back on.}
  \label{fig: social_minigame}
  % \vspace{-10pt}
\end{figure}

In practice, we approximate collision by overlap of convex hulls inflated by a safety margin. We depict several examples and non-examples of social mini-games in Figure~\ref{fig: social_minigame}. The first scenario can be characterized as a social mini-game due to the conflicting preferred trajectories of agents $1$ and $2$ within a specific time interval $[a, b]$, where the duration $b-a \geq \delta$. In this case, the agents' trajectories intersect, generating a conflict. However, the second and third scenarios, in contrast, do not qualify as social mini-games since no conflicts arise between the agents. In the second scenario, there is no common time duration where agents intersect each other, and their trajectories remain independent in time. In the third scenario, agent $2$ possesses an alternative preferred trajectory that avoids conflicts during any time duration, allowing for a seamless transition to a conflict-free path.

SMGs are high-agency MRN scenarios. Here, agents are not only moving in close proximity to other agents, but the key distinguishing characteristic of SMGs is that, unless some agent changes its course or takes a different action than its optimally decided one, there will be an undesirable outcome. That is what is meant by high agency--each agent is significantly impacting other agents such that they need to coordinate with each other to survive. Examples of SMGs are shown in Figure~\ref{fig:social_benchmarks} and discussed in detail in the following Section. But from a glance, it can be observed that each of these scenarios is an SMG. Consider for example the doorway scenario: unless either of the green or orange agent alters its course, they will collide. {We next give a mathematical definition of SMGs:}
% It says that if we assume each agent has one or more individual optimal trajectory (for example, in the doorway scenarios, the optimal trajectory for both the green and the orange agent would be a straight line path towards the red cross.), then if all the individual optimal trajectories intersect for all agents within some time interval, then that scenario is an SMG.
Beyond saying that SMGs are a subset of general MRN, it is important to highlight the exact characteristics that distinguish the two types of scenarios. We give such a list below.
% where we define
% where we informally define `\textit{capacity-one resource}' to refer to a region of space-time (e.g., a doorway segment) that at most one agent can occupy at a time. 

\begin{tcolorbox}[mydefinition]
An SMG possesses the following characteristics, at least for some nontrivial interval, that distinguish them from generic MRN scenarios:
\begin{enumerate}
\item \textbf{Mutual occupancy contention:} persistent intersection of agent convex hulls across individual optimal trajectories.
\item \textbf{Capacity-one resource/bottleneck contention:} two or more agents contend for a shared, capacity-one resource (doorway, merge lane), which is a subset $R \subset \mathcal{X}$ where contention occurs when the optimal trajectories of at least two agents intersect $R$ in overlapping time intervals.
\item \textbf{Crossing flows:} orthogonal or oblique approach trajectories induce right-of-way or turn-taking.
\item \textbf{Occlusion/limited observability:} beliefs about others’ motion/goals materially shift best responses (e.g., blind corners).
\item \textbf{Conflicting objectives:} heterogeneous costs (e.g., speed vs. smoothness, priority agent vs. regular agent) create trade-offs.
\end{enumerate}
\end{tcolorbox}

SMGs are embedded within MRN: most trajectories are decoupled, but at specific times (bottlenecks, crossings, occlusions) the interaction becomes a game with coupled best responses. MAPF also resolves conflicts but (classically) in discrete, fully observed, centralized settings that compute joint plans offline; SMGs arise online, continuous-time, often decentralized, with heterogeneous objectives and social norms. Furthermore, it is typical for SMGs to include around 2-4 agents in the most commonly occurring SMGs. Of course, an environment may consist of multiple SMGs occurring simultaneously. Our scenarios (Sec.~\ref{sec:benchmark_scenarios}) are chosen because they activate specific SMG characteristics, and our metrics (Sec.~\ref{sec:evaluation_metrics}) include SMG-aware outcomes that are immaterial in decoupled MRN segments but decisive inside SMGs.

\begin{figure*}[t]
    \centering
    \begin{subfigure}[t]{0.49\linewidth}
        \centering
        \includegraphics[width=\linewidth]{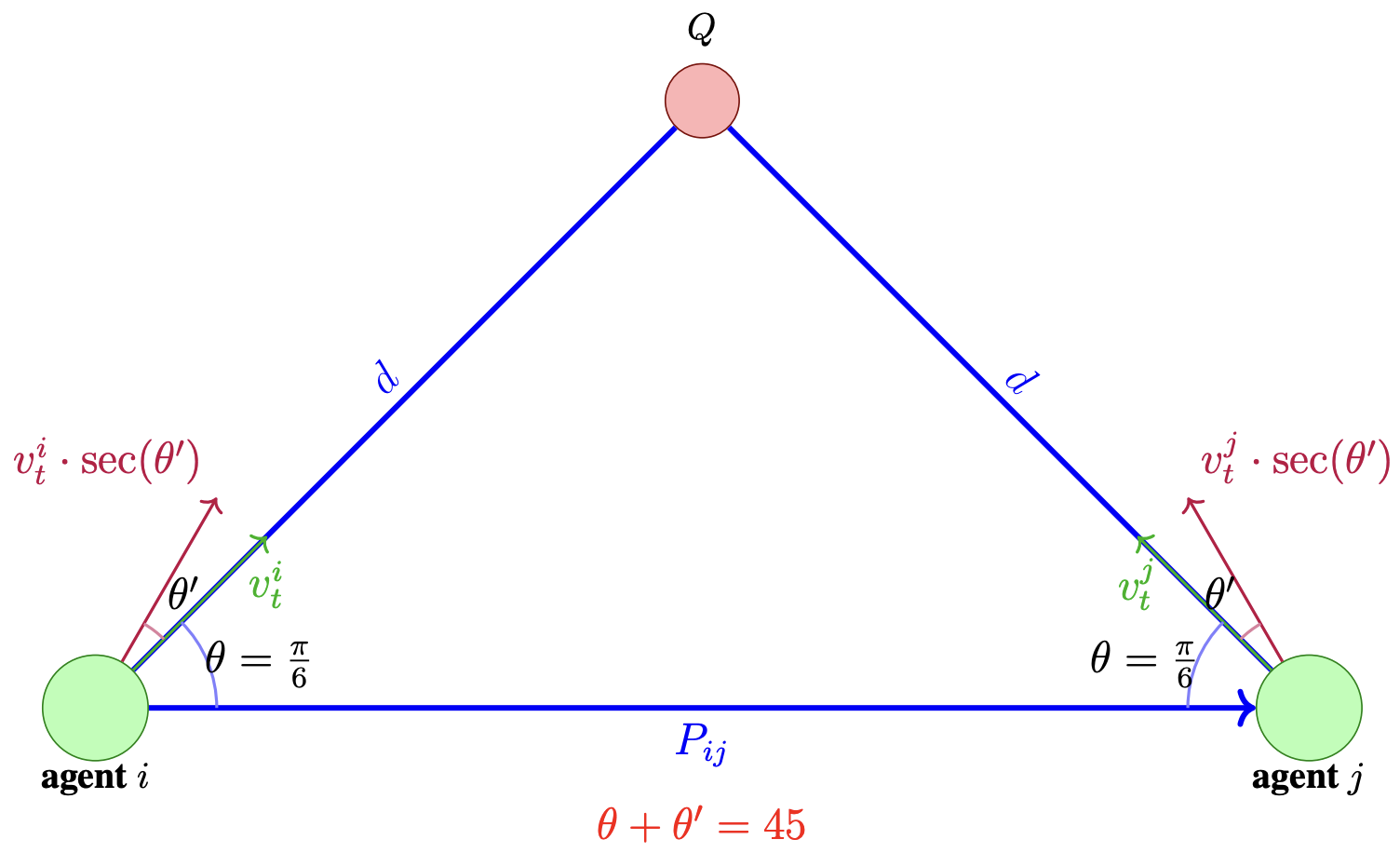}
        \caption{Velocity projection for $\theta = \frac{\pi}{6}$.}
        \label{fig: smg30}
    \end{subfigure}
    \begin{subfigure}[t]{0.49\linewidth}
        \centering
        \includegraphics[width=\linewidth]{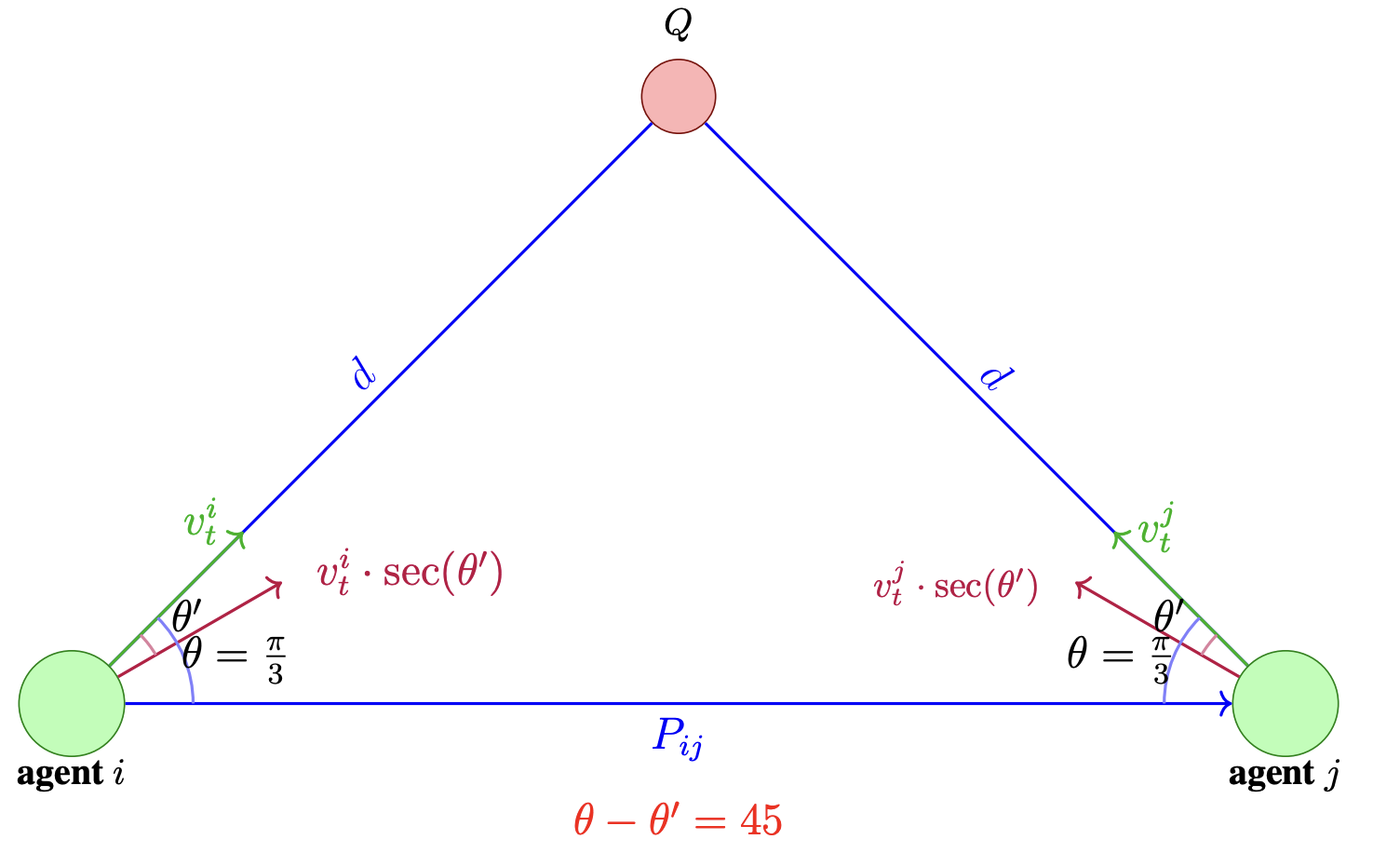}
        \caption{Velocity projection for $\theta = \frac{\pi}{3}$.}
        \label{fig: smg60}
    \end{subfigure}
    \caption{Velocity projection for two symmetrical SMGs with arbitrary angles $\theta$ with respect to the relative position vector. In these two examples, we examine when $\theta = \frac{\pi}{6}$ and $\theta = \frac{\pi}{3}$ (although any arbitrary $\left \lvert \theta \right\rvert < \frac{\pi}{2}$ may be chosen).}
    \label{fig: velocity_projection}
\end{figure*}

\subsection{When Does General MRN Become An SMG?}
\label{subsec: smgness}
Symmetry arises in MRN when multiple agents occupy equivalent roles, possess similar capabilities, and face identical interaction structures. In symmetric SMGs, such as two robots approaching a narrow doorway, agents have identical incentives and information, resulting in multiple equivalent collision-free solutions \cite{chandra2025deadlock}. Without an explicit symmetry-breaking mechanism, such interactions can converge to undesirable equilibria, including deadlocks or oscillatory behavior.

Geometrically, we quantify symmetry through a liveness function $\ell_j\in[0,\pi]$,

\begin{tcolorbox}[mydefinition]
\begin{definition}
\textbf{Liveness Function}~\cite{chandra2025deadlock}: A liveness function $\ell_j:\mathbb{R}^2\times\mathbb{R}^2\rightarrow\mathbb{R}$ between agents $i$ and $j$ ($i\neq j$) as follows:
\begin{align}
\ell_j(p_t^i,v_t^i)=\cos^{-1}\left(\frac{\left \langle \overrightarrow{p_t^i-p^j_t},\overrightarrow{{ v^{i, \prime}_t}- {v^{j, \prime}_t}}\right \rangle}{\left \lVert \overrightarrow{p^i_t-p^j_t}\right \rVert \left \lVert \overrightarrow{{ v^{i, \prime}_t}- { v^{j, \prime}_t}}\right \rVert+\epsilon}\right)
\label{eqn:liveness_formula}
\end{align}
\textit{where ${v^{i, \prime}_t} = v^i_t\cdot \sec(\theta'), {v^{j, \prime}_t} = v^j_t\cdot \sec(\theta')$ correspond to the velocity projection component of each robot's velocity such that it subtends an angle $\theta^\prime = \pm\left(\frac{\pi}{4} - \theta\right)$ with the original velocity vectors, $v^i_t$ and $v^j_t$,
% , as defined by the velocity transformation procedure,
where $\left\lvert\theta\right \rvert < \frac{\pi}{2}$ is the arbitrary SMG angle. 
% Further, $\left \langle a, b \right \rangle$ denotes dot product between vectors $a$ and $b$ 
% and $\|.\|$ denotes the Euclidean norm, 
$\epsilon>0$ ensures that the denominator is positive. Lastly, note that $\ell_j(p_t^i,v_t^i)=\ell_i(p_t^j,v_t^j)$ for $i\neq j$.}
\label{def: liveness_function}
\end{definition}
\end{tcolorbox}

% \noindent\textbf{Quantifying an SMG using the Liveness Fucnction:} 
The liveness function is meant to geometrically capture the symmetry of an SMG. The lower the liveness function value, the closer the scenario is to an SMG, and conversely, the higher the liveness function value, the closer the scenario is towards a general MRN. Formally, the liveness function measures the angle between the relative displacement and the relative linear velocity vectors. In the case of perfect symmetry, $\ell(p_t^i,v_t^i)$ will be close to $0$ as the relative position and velocity vectors are nearly perfectly aligned (parallel) implying the dot product will be nearly $1$. This situation is representative of an SMG where there is perfect contention or conflict over a capacity-one resource such as a doorway or intersection scenario. 

The dot product decreases ($\ell(p_t^i,v_t^i)$ increases) as the symmetry decreases. Recall that if there is no symmetry, then there is low chance of a contention or bottleneck. The $\epsilon$ in the denominator represents the practicality of the fact that there always exists a floating point difference among agents' velocities, no matter how close they may be.

% An important observation is that any symmetrical SMG with arbitrary angle $\theta$ can be reduced to a symmetrical SMG with angle $\frac{\pi}{4}$. In any symmetrical SMG with arbitrary angle $\theta, \left \lvert \theta \right \rvert < \frac{\pi}{2}$ (agents diverge away from each other for $\left \lvert \theta \right \rvert \geq \frac{\pi}{2}$), with respect to the relative position vector, suppose two robots $i$ and $j$ have nearly identical velocities $v^i_t \approx v^j_t$ (otherwise, it would not be a symmetrical SMG) or $\left \lvert v^i_t - v^j_t \right \rvert \leq \epsilon$. Then, without loss of generality, we can project a component of each robots velocity such that it subtends an $\theta^\prime$ with the original corresponding velocity vector. The advantage of such a projection is that it eliminates the dependency on the angle, with the only dependency remaining on the scaling factor $\zeta$. We set $\theta'$ as follows:

% \begin{enumerate}
%     \item \textbf{Case 1:} When the arbitrary angle $\theta < \frac{\pi}{4}$, then we set $\theta' = \frac{\pi}{4} - \theta$. 
%     % \input{images/smg30_tikz}

%     \item \textbf{Case 2:} When the arbitrary angle $\frac{\pi}{2}>\theta \geq \frac{\pi}{4}$, then we set $\theta' = \theta - \frac{\pi}{4}$. 

% \end{enumerate}

% The new transformed velocities are given by $v^{i, \prime}_t = v^i_t \sec\theta^\prime$ and $v^{j, \prime}_t = v^j_t \sec \theta^\prime$. 

Figure~\ref{fig: velocity_projection} demonstrates the velocity projection for two symmetrical SMGs with arbitrary angles $\theta$ with respect to the relative position vector. In Figures~\ref{fig: smg30} and~\ref{fig: smg60}, we examine when $\theta = \frac{\pi}{6}$ and $\theta = \frac{\pi}{3}$, respectively, although any arbitrary $\left \lvert \theta \right\rvert < \frac{\pi}{2}$ may be chosen.

\subsection{Common SMG Scenarios}
\label{sec:benchmark_scenarios}

In this Section, we briefly list a non-exhaustive set of SMG scenarios that occur in everyday life. These scenarios are compiled from a combination of prior social navigation literature~\cite{francis2025principles, francis2023socialhri} and robotics literature~\cite{long2018towards, chen2017socially}. Each canonical scenario activates a distinct subset of SMG characteristics. While not exhaustive, they are chosen to cover the most frequently encountered topologies in real life such as doorways, intersections, and blind corners that expose different deadlock mechanisms, visibility challenges, and coordination demands. Additional scenarios, as and when they are discovered and analyzed, will be added to our accompanying software library.

\begin{itemize}

\item {{Doorway:}} In this scenario, navigating through tight spaces is the main challenge. The doorway creates a bottleneck, testing how well the planners manage congestion and ensure smooth movement. This scenario highlights the importance of prioritizing and making quick decisions to prevent potential deadlocks. 

\item {{Intersection:}} This scenario aims to replicate a common social challenge where agents from different directions meet at an intersection. This scenario tests the planner's ability to handle complex interactions between multiple agents. Anticipating the actions of other agents and making quick decisions are crucial to avoid deadlocks. 

\item {{Hallway:}} This scenario features a corridor with two-way traffic and tests how well planners manage overtaking, yielding and keeping traffic moving smoothly. It evaluates how efficiently space is used, head-on deadlocks are prevented, and smoothness is maintained. This mirrors the real-world hallway or corridor navigation challenges that we often find in a social setting, highlighting the importance of efficient spatial coordination. 

\item {{L-Corner:}} The L corner scenario highlights the challenge of navigating at a  sharp turn in a L-shaped environment. As agents approach the corner, limited visibility and the chance of colliding with other agents requires a strong deadlock avoidance or resolution strategy. In this scenario both agent start from same position in same direction and move towards same goal position. 

\item {{Blind Corner:}} This scenario, with its limited visibility and sudden encounters, tests the planner's ability to anticipate and react. It focuses on how quickly agents can adjust to unexpected situations, make real-time decisions to prevent deadlocks, and navigate smoothly around corners. This is different from L-corner because it involves the agent moving towards each other having same goal position with limited visibility, thus increasing the risk of collision.

\item {{Crowded Traffic:}} This scenario mimics busy environments, testing how well planners handle many dynamic agents moving randomly at once. The main goal is to navigate through the crowd efficiently, avoid collisions, deadlocks, collision with moving obstacles and keep a good pace while ensuring smooth traffic flow. 

\item {{Parallel Traffic:}} This scenario evaluates agents moving in parallel paths for their ability to keep traffic flowing smoothly, avoid side-by-side deadlocks, and handle differences in speed.  

\item {{Perpendicular Traffic:}} This scenario involves agents crossing paths perpendicularly, creating complex traffic dynamics. The challenge is to manage right-of-way, prevent collisions, deadlocks at the intersection, and ensure smooth traffic flow amidst intersecting movements. 

\item {{Circular Traffic:}} This scenario aims to mimic circular traffic patterns and tests how well planners handle circular flows. The focus is on the ability to merge into, navigate through, and exit from circular flows, without causing collisions and deadlocks. 
\end{itemize}

This curated set is representative rather than exhaustive; it spans the minimal SMG characteristics we listed repeatedly across buildings, warehouses, and campuses, while keeping benchmarks compact and diagnostic.

\begin{figure}[!t]
    \centering
    % First Row
     \begin{subfigure}{0.29\columnwidth}
        \includegraphics[width=\linewidth]{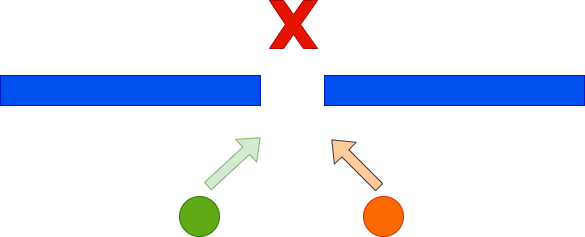}
        \caption{Doorway}
    \end{subfigure}\hfill
     \begin{subfigure}{0.29\columnwidth}
        \includegraphics[width=\linewidth]{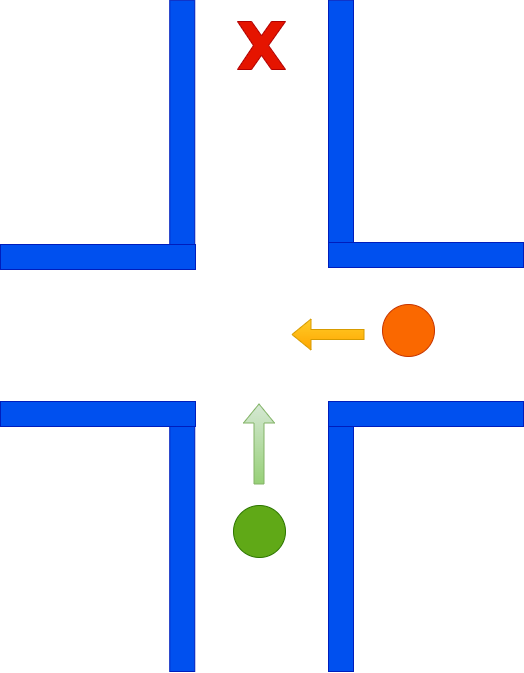}
        \caption{Intersection}
    \end{subfigure}\hfill
    \begin{subfigure}{0.29\columnwidth}
        \includegraphics[width=\linewidth]{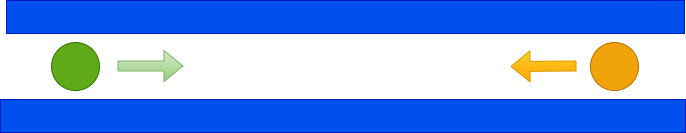}
        \caption{Hallway}
    \end{subfigure}
    \vspace{1ex} % Space between row
    % Second Row
    \begin{subfigure}{0.29\columnwidth}
        \includegraphics[width=\linewidth]{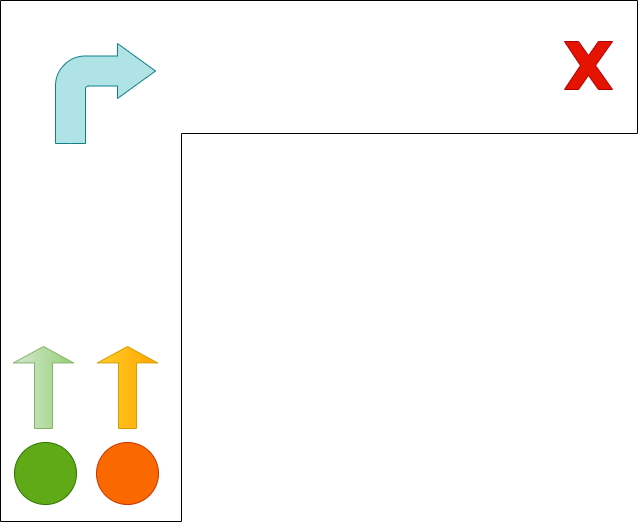}
        \caption{L-Corner}
    \end{subfigure}\hfill
    \begin{subfigure}{0.29\columnwidth}
        \includegraphics[width=\linewidth]{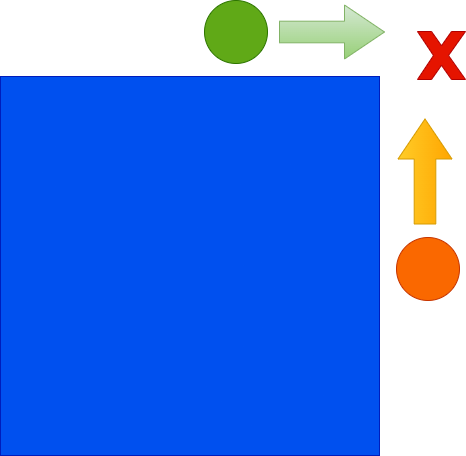}
        \caption{Blind Corner}
    \end{subfigure}\hfill
    \begin{subfigure}{0.29\columnwidth}
        \includegraphics[width=\linewidth]{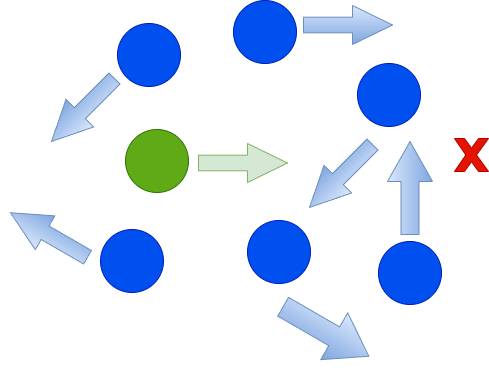}
        \caption{Crowded Traffic}
    \end{subfigure}
    \vspace{1ex}
    % Third Row
    \begin{subfigure}{0.29\columnwidth}
        \includegraphics[width=\linewidth]{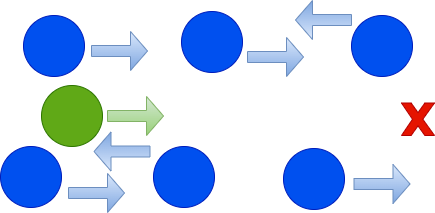}
        \caption{Parallel Traffic}
    \end{subfigure}\hfill
    \begin{subfigure}{0.29\columnwidth}
        \includegraphics[width=\linewidth]{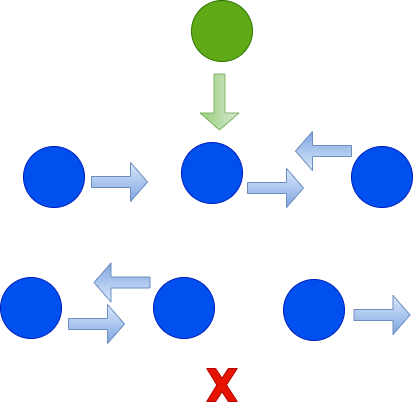}
        \caption{Perpendicular Traffic}
    \end{subfigure}
    \begin{subfigure}{0.29\columnwidth}
        \includegraphics[width=\linewidth]{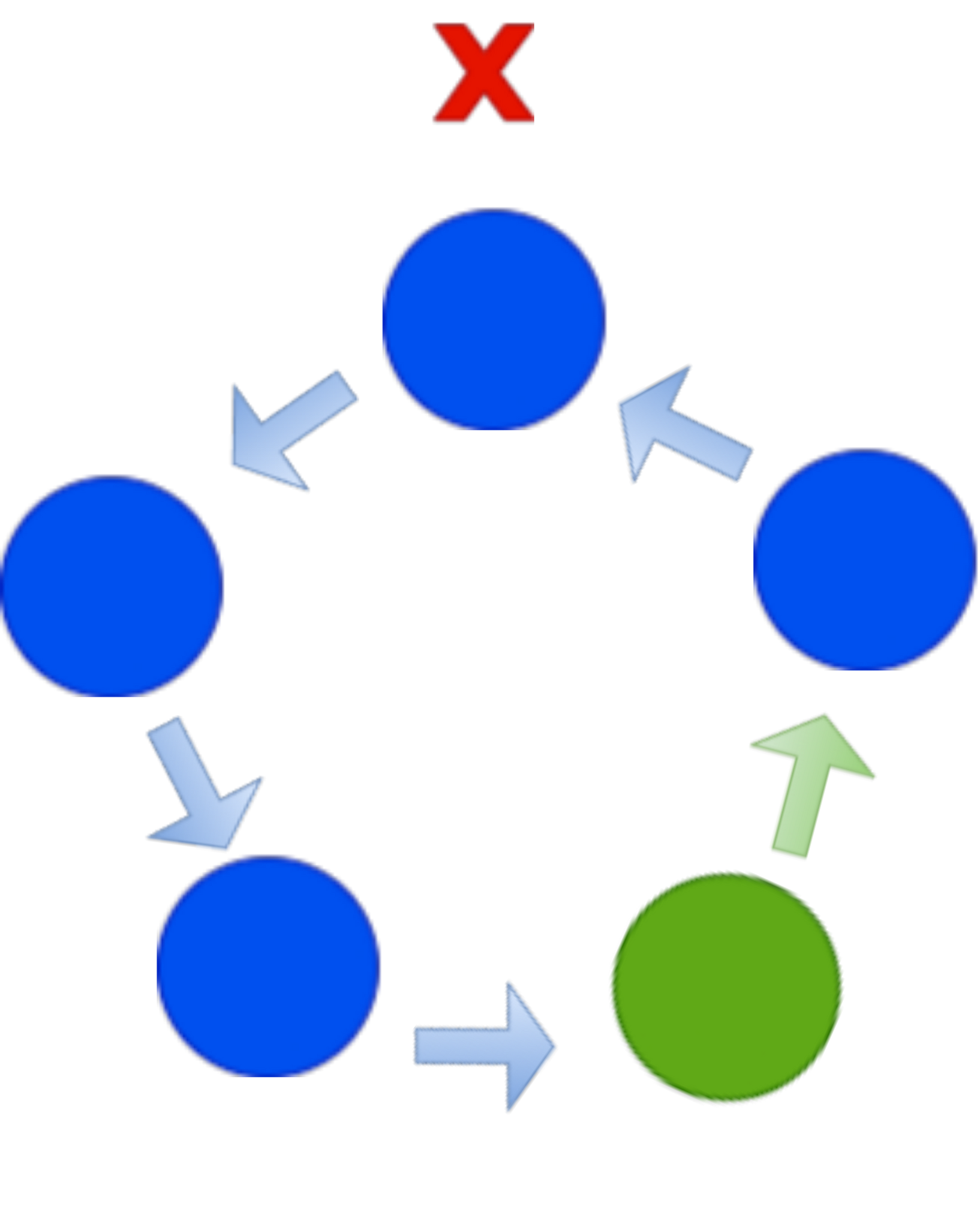}
        \caption{Circular Traffic}
    \end{subfigure}
    \caption{Common examples of SMGs. Here the green agent denotes the agent of interest, with other agents being denoted by orange and blue solid circles. Green agent's goal is marked by red X.}
    \label{fig:social_benchmarks}
\end{figure}

\subsection{Evaluation Metrics}
\label{sec:evaluation_metrics}

The typical metrics for evaluating general MRN are the following: average $\Delta V$, average delay, and path deviation. We chose to include these metrics as they enable benchmarking with general MRN methods. The average $\Delta V$ metric helps measure movement smoothness, the average delay compares time efficiency between agents, and path deviation assesses navigation accuracy. Together, these metrics ensure that the path planners can handle real-world challenges effectively, navigating efficiently, smoothly, and reliably while avoiding deadlocks. This thorough evaluation is essential for developing robust multi-agent systems in complex environments. Although existing metrics for general MRN remain essential, in an SMG, agents have high agency. We therefore also report the following metrics: flow rate, fairness (game-theoretic perspective), and influence (externalities imposed on others). The following sections discuss each of them in more detail:

\begin{itemize}

\item {{Avg $\triangle V$ (Average Change in Velocity)}}: This metric measures how much an agent’s speed changes on average at each time step during its movement. It shows how often and how much the agent adjusts its speed in response to its surroundings and other agents. Since most real world systems use a discrete-time controller, it is more tractable to use a discrete version of the metric so that the metric clock can be synchronized with the robot clock. The metric is computed as: 
\[\Delta V_{avg} = \frac{\sum (\Delta V)}{\sum t}\]

where $\Delta V$ is summed over the entire episode over which the metric is evaluated. A lower Avg $\Delta V$ means smoother and more consistent movement, with fewer disruptions or sudden changes. A higher Avg $\Delta V$ means the agent is often changing its speed, which could indicate navigation challenges or the need for constant adjustment in a changing environment. All experiments use the same fixed time-step, so comparisons across methods are well-defined even if the absolute number changes with $\Delta T$.

\item {{Average Delay}}: 
To quantify the temporal inefficiency introduced by multi-agent interactions, we adopt a delay metric inspired by classical traffic flow theory. 
% Rather than comparing completion times across different agents, 
We measure the additional time incurred by each agent relative to its own nominal travel time in the absence of interactions. For each agent $i$, let $T^i_{\text{SMG}}$ denote the time required for the agent to reach its goal in the SMG. We define a baseline travel time $T^i_{\text{solo}}$ as the time required for the same agent to reach the same goal when navigating alone in the same environment under the same planning or control method. The delay experienced by agent $i$ is then defined as $D^i = T^i_{\text{SMG}} - T^i_{\text{solo}}$. Using this definition, we compute the {Average Delay} across all agents as
\[
\text{AverageDelay} = \frac{1}{k} \sum_{i=1}^{k} D^i.
\]
This metric captures the average additional time agents experience due to congestion, coordination requirements, or bottleneck contention. A value of zero indicates that agents experience no performance degradation relative to solo navigation. Because nominal travel times may vary across agents and scenarios, we additionally report a normalized version of the metric:
\[
\text{NormalizedAverageDelay} =
\frac{1}{k} \sum_{i=1}^{k}
\frac{T^i_{\text{SMG}} - T^i_{\text{solo}}}{T^i_{\text{solo}}}.
\]
The normalized delay enables fair comparison across heterogeneous agents and environments by measuring relative rather than absolute performance degradation. Larger values indicate greater interaction-induced inefficiency and stronger coupling among agents.

\item {{Path Deviation}}: Path Deviation measures how much an agent’s path differs from a reference or ideal path. It uses the Hausdorff distance, which calculates the greatest distance from a point in one path to the closest point in the other path. Further note that there may exist multiple optimal trajectories. Let $\Gamma^*_\textrm{opt}$ denote the set of nominal optimal paths (e.g., straight lines or solutions from a reference planner). We define Path Deviation as

\[\min_{\Gamma\in \Gamma^*_\textrm{opt}} d_\textrm{Haus} \left( \Gamma, \Gamma^* \right)\]

Note that this metric is purposefully kept broad so that it can be adapted to other deviation metrics by the user. In this case, the deviation is measured against the ideal scenario where only one agent moves towards the goal without any interactions or obstacles, defined as the nominal agent scenario. A low Path Deviation means the agent’s actual path is very close to the optimal path, indicating efficient navigation. A high Path Deviation means the agent took significant detours or deviations, possibly due to interactions with other agents, obstacles, avoiding or resolving deadlocks in the way or irregular control generation leading to poor navigation.

\item {{Flow Rate}}: We define the {flow rate} to quantify how efficiently multiple agents traverse a spatial bottleneck such as a doorway, corridor opening, or narrow passage. Let $G \subset \mathcal{X}$ represent a \emph{gap region}, defined as a line segment or spatial region (e.g., a doorway cross-section) with physical width $z$ (measured in meters). We assume each agent $i$ has a convex hull footprint $C_i(x^i_t)$ determined by its state $x^i_t$. An agent is considered to have \emph{passed through the gap} when the center of its convex hull transitions from one side of $G$ to the other. Let $S$ denote the set of agents assigned to traverse the gap $G$. To ensure consistency in evaluation, we define a measurement time interval using a standardized convention. The start time $t_{\text{start}}$ is defined as the earliest time at which any agent in $S$ enters a predefined entry region adjacent to $G$. The end time $t_{\text{end}}$ is defined as the time at which the last agent in $S$ exits a predefined post-gap region on the opposite side of $G$. Alternative implementations may use a fixed time horizon, but throughout this work we adopt the entry-exit convention described above. The flow rate is then defined as
\begin{equation}
\text{FlowRate} = \frac{|S|}{z \left( t_{\text{end}} - t_{\text{start}} \right)},
\end{equation}
where $|S|$ denotes the number of agents required to traverse the gap. This metric captures the density-normalized throughput of agents across the bottleneck, and is analogous to classical traffic-flow formulations that measure volume of traffic per unit time normalized by available capacity.

\item Fairness: We adopt a game-theoretic formulation of fairness that captures how individual agents balance their own objectives with the collective performance of the multi-agent system. Let $\pi = \left(\pi^1,\dots,\pi^k\right)$ denote a \emph{strategy profile}. Let $\Gamma^i(\pi)$ be the trajectory realized by agent $i$ when all agents execute the strategy profile $\pi$. Thus, strategies represent decision-making rules, while trajectories represent the resulting state evolution induced by those rules. Let $R^i_{\text{local}}(\pi)$ denote the cumulative reward obtained by agent $i$ when all agents execute $\pi$. We define the global system reward as
\begin{equation}
R_{\text{global}}(\pi) = \sum_{i=1}^k \phi^i R^i_{\text{local}}(\pi),
\end{equation}
where $\phi^i \geq 0$ represents the priority or importance weight of agent $i$ (e.g., emergency vehicles may be assigned larger $\phi^i$ values).

While the underlying planning or learning algorithm may seek to optimize either local rewards or the global reward, fairness is evaluated using the realized outcomes induced by the selected strategy profile. Specifically, we measure fairness through the disparity between each agent's achieved local reward and the reward it would obtain under a globally optimal cooperative strategy:
\begin{equation}
F = 1 - \frac{1}{k} \sum_{i=1}^k 
\frac{
\left| R^i_{\text{local}}(\pi^\star) - R^i_{\text{local}}(\pi) \right|
}{
\max \left( |R^i_{\text{local}}(\pi^\star)|, \epsilon \right)
},
\end{equation}
where $\pi^\star = \arg\max_\pi R_{\text{global}}(\pi)$ denotes a fully cooperative strategy profile and $\epsilon > 0$ is a small constant used to avoid division by zero.

This metric quantifies the degree to which the executed strategy profile $\pi$ aligns with a socially optimal allocation of rewards. Higher fairness values indicate that agents collectively achieve reward distributions closer to those under the cooperative optimum. Importantly, fairness is evaluated on the realized trajectories induced by the strategy profile rather than on the optimization procedure itself.

\item Influence Score (IS): The influence score measures the extent to which the presence of an agent alters the behavior of other agents in the system. We define influence using a game-theoretic counterfactual formulation. As before, let $\pi = (\pi^1,\dots,\pi^k)$ denote a strategy profile. For a given agent $i$, we define the counterfactual strategy profile $\pi^{-i}$ as the strategy profile obtained by removing agent $i$ from the environment and re-solving the same SMG using the same planning or learning algorithm for the remaining agents. The resulting control input for agent $j \neq i$ is denoted $u^{j,-i}_t$.

The influence score of agent $i$ is defined as
\begin{equation}
IS^i = \sum_{j \neq i} \int_0^T \left \lVert u^j_t - u^{j,-i}_t \right \rVert_2 \, dt.
\end{equation}

This metric quantifies how strongly agent $i$ perturbs the control behavior of other agents. Higher values of $IS^i$ indicate that the presence of agent $i$ forces larger control deviations in other agents, suggesting more aggressive or disruptive interaction behavior. 

In practice, for deterministic planners or controllers, $u^j_t$ and $u^{j,-i}_t$ are computed using two simulations with identical initial conditions except for removing agent $i$. For stochastic planners or learning-based policies, the control signals may be replaced by expected controls or averaged across multiple simulation rollouts. We emphasize that the influence score is an evaluation metric applied to the realized outcomes induced by a selected strategy profile. It does not modify or redefine the underlying planning or learning objective.

\end{itemize}

\section{Algorithms, Taxonomy and Properties for SMGs}
\label{sec:Taxonomy}

% \noindent
% \textbf{Scope.}
We analyze algorithms and system design choices {conditioned on an active
SMG}, i.e., when at least one SMG property from Sec.~\ref{sec:benchmark_scenarios} is present (bottleneck contention, crossing flows, persistent mutual occupancy, occlusion,
or conflicting objectives). Let $\mathcal{K}_m(t)\!\subseteq\!\{1,\dots,k\}$ denote
the {active coupling set}, that is, the minimal subset of agents whose best responses
are mutually coupled during $[a,b]$ (Sec.~\ref{sec:benchmark_scenarios}). At any time $t$, agent interactions may induce multiple, possibly overlapping regions of strategic coupling. To formalize this, we define an interaction graph $\mathcal{G}(t) = (\mathcal{V}, \mathcal{E}(t))$, where the node set $\mathcal{V} = \{1,\dots,k\}$ corresponds to agents, and an undirected edge $(i,j) \in \mathcal{E}(t)$ exists if the best-response policy of agent $i$ depends nontrivially on the actions or state of agent $j$ over a nonzero time interval containing $t$. We define the \emph{coupling sets} $\{\mathcal{K}_m(t)\}_m$ as the connected components of the interaction graph $\mathcal{G}(t)$. Each $\mathcal{K}_m(t) \subseteq \mathcal{V}$ represents a maximal subset of agents whose decision-making processes are mutually coupled at time $t$. When the context is clear, we use $\mathcal{K}(t)$ to denote a generic coupling set.

This formulation naturally allows for multiple coupling sets to exist simultaneously. Moreover, coupling sets may overlap across time as interaction contexts evolve. For example, if agents $\{1,2\}$ are coupled while traversing doorway A and agents $\{2,3\}$ are later coupled while traversing doorway B, then distinct coupling sets $\{1,2\}$ and $\{2,3\}$ may exist at different times. If the planning or learning algorithm reasons jointly over all interacting agents, the union $\{1,2,3\}$ may be treated as a single coupled group over the corresponding time interval. Throughout this paper, our taxonomy, metrics, and SMG analyses are applied independently to each active coupling set $\mathcal{K}_m(t)$. This enables disjoint social mini-games to be identified, analyzed, and evaluated without requiring global coupling across all agents. Throughout, we tie design choices to {game-aware outcomes} introduced in Sec.~\ref{sec:evaluation_metrics}: {Social Welfare Gap (SWG)}, {Turn-Taking Fairness (TTF)}, and {Invasiveness Score (IS)}, reported alongside standard MRN metrics. This section thus specifies {how MRN methods become SMG solvers} once a
trigger activates and a coupling set emerges.

\subsection{Categorization of SMG Solvers by Paradigm}
\label{sec:algorithms}

While classical MRN algorithms span multiple paradigms such as MARL, MAPF, and optimization-based methods, in SMGs these paradigms play a specific role: they govern how agents in the active coupling set $\mathcal{K}_m(t)$ generate strategies once an SMG characteristic is observed. Because SMGs are local high-agency scenarios embedded within broader MRN, algorithmic choices must balance two concerns: $(i)$ scalability and safety across the full swarm, and $(ii)$ responsiveness and fairness within $\mathcal{K}_m(t)$. In this subsection, we categorize existing methods into four broad paradigms: Multi-Agent Reinforcement Learning (MARL), Multi-Agent Path Finding (MAPF), Optimization (including MPC/CBFs), and heuristic or hybrid approaches, and discuss how each paradigm addresses SMG-specific challenges such as improving fairness and flow rate, and reducing the invasiveness score.

\subsubsection{Multi-Agent Reinforcement Learning (MARL)}

MARL extends single-agent reinforcement learning to environments with multiple learning agents, operating within the stochastic games framework where the key challenge is that each agent's learning process is affected by other learning agents (also known as the well-known non-stationarity problem). A MARL formulation models the system using the POSG tuple defined in Section~\ref{subsec: MRN_prelim}. In MARL, the primary objective of each agent $i$ is to minimize an expected cost function: $\pi^{i,*} = \arg\max_{\pi^i} \mathbb{E}_{\pi^i, \pi^{-i}}[\mathcal{J}^i(u^i, u^{-i})] $ using standard RL techniques, where $\pi^{-i}, u^{-i}$ denote the joint policies and actions of all agents except $i$.

% stochastic game defined by the tuple
% \[
% \left\{ S, \{A_i\}_{i=1}^N, P, \{R_i\}_{i=1}^N \right\},
% \]
% where:
% \begin{itemize}
%     \item $S$ is the joint state space,
%     \item $A_i$ is the action space of agent $i$,
%     \item $P(s' \mid s, a_1,\dots,a_N)$ is the transition probability,
%     \item $R_i(s,a_1,\dots,a_N)$ is the reward function of agent $i$.
% \end{itemize}

% A strategy (policy) for agent $i$ is a mapping
% \[
% \pi_i : S \rightarrow A_i
% \]
% (or from observation histories to actions in partially observable settings). The joint policy $\pi = (\pi_1,\dots,\pi_N)$ induces a trajectory distribution over state sequences.? 

An SMG arises when there exists non-stationarity in a MARL approach. Non-stationarity refers to the scenario where the RL policy of an agent fails to converge because the agent's environment keeps changing due to varying actions of other agents. That is, when the reward or transition function induces strong interdependence among a subset of agents such that it leads to non-stationarity, then it can be shown that those agents are the ones that typically belong to $\mathcal{K}_m(t)$, and therefore institute a SMG. In particular, the local state relevant to the SMG consists of $\{x^i : i \in \mathcal{K}_m(t)\}$ and environmental features defining the bottleneck (e.g., doorway geometry), the actions are velocity or acceleration commands, and the rewards couple agents through collision penalties, shared resource contention, or delay penalties. Thus, in MARL, an SMG corresponds to a localized stochastic subgame restricted to agents in $\mathcal{K}_m(t)$. In active SMGs, MARL naturally models strategic coupling and heterogeneous costs,
but requires $(i)$ {safety layers} (e.g., CBF filters) to avoid
catastrophic off-equilibrium play, $(ii)$ {symmetry-breaking} priors or norms
to reduce inefficient equilibria, and $(iii)$ {partial observability handling}
(occlusions) via belief or prediction modules. With respect to SMGs, MARL can improve fairness and reduce IS.

MARL algorithms are categorized as- a) Value-based b) Multi-Agent Deep Q-Networks c) Policy-based. A popular Value-based method is Q-Learning~\cite{tampuu2017multiagent}, where each agent's Q-value update rule takes in account of the optimal policy for all agents at equilibrium, governed by an evaluation operator. Multi-Agent Deep Q-Networks (MADQN)~\cite{orr2023multi} extend Deep Q-Networks to multi-agent settings. Value-based methods suffer from scalability issues due to joint action space. On the other hand, Policy-based methods can learn a parametrized policy for each agent $i$. This is the essence of MADDPG~\cite{lowe2017multi}, where each agent learns its own deterministic or stochastic policy. A fundamental challenge in MARL is solving stochastic games, which can be interpreted as a sequence of normal-form (matrix) games. Often the agents seek Nash equilibrium (NE) solutions where no agent can unilaterally improve their performance by deviating from their policy. The NE can also be directly fed into the Q-learning update by replacing the evaluation operator with a Nash-based operator~\cite{hu2003nash}. However finding and converging to such equilibrium remains computationally challenging. Consequently, one can replace explicit equilibrium computation with learned policies guarded by a reachability certified safety module. Layered Safe MARL \cite{choi2025resolving} trains a policy to minimize multi-agent conflicts and then applies a CBVF \cite{choi2021robust} reachability based safety filter to resolve the hardest pairwise interactions, yielding safe, efficient navigation in dense traffic and drone swarm hardware tests. 

% \noindent \textbf{SMG implications (MARL).} In active SMGs, MARL naturally models strategic coupling and heterogeneous costs,
% but requires $(i)$ {safety layers} (e.g., CBF filters) to avoid
% catastrophic off-equilibrium play, $(ii)$ {symmetry-breaking} priors or norms
% to reduce inefficient equilibria, and $(iii)$ {partial observability handling}
% (occlusions) via belief or prediction modules. With respect to SMGs, MARL can improve fairness and reduce IS.

\subsubsection{Multi-Agent Path Finding (MAPF)}
MAPF algorithms solve the problem of computing conflict-free paths for multiple agents navigating a graph. The environment is modeled as a graph $G = (V,E)$.
Each agent $i$ has an initial vertex $v^i_{\text{start}} \in V$ and a goal vertex $v^i_{\text{goal}} \in V$. At each discrete time step, an agent may move to an adjacent vertex or wait. A solution consists of collision-free paths $\Gamma^i = \left(v^i_0, v^i_1, \dots, v^i_T \right)$
such that $v^i_0 = v^i_{\text{start}}$ and $v^i_T = v^i_{\text{goal}}$. Typical objectives include sum of costs (total path length) and Makespan (maximum arrival time). 

An SMG appears when multiple agents' shortest paths traverse a shared low-capacity region (e.g., a single-width corridor node or edge). In this case, the state is the discrete occupancy configuration, actions are graph transitions, and resource contention manifests as vertex or edge conflicts. The coupling set $\mathcal{K}_m(t)$ consists of agents whose shortest paths intersect at shared vertices or edges within overlapping time intervals. The SMG is the local discrete coordination problem required to resolve those conflicts. Classical MAPF is discrete, centralized, fully cooperative and observable. In SMGs, it is best used as a {local joint-planning subroutine} on $\mathcal{K}_m(t)$: to approximate a socially optimal joint action at bottlenecks/crossings, or to resolve deadlock by injecting priorities. Continuous-time and non-holonomic feasibility is enforced by a short-horizon motion layer. We can achieve strong fairness guarantees when $\mathcal{K}_m(t)$ is small and communication is available.

MAPF problems are discrete, centralized, fully cooperative and observable, and entails a global cost function. Two primary optimization criteria dominate MAPF research: sum-of-cost, which minimizes the total steps taken by all agents, and makespan, which minimizes the time for the last agent to reach its goal. MAPF algorithms must balance three fundamental properties that create inherent trade-offs. Optimality ensures minimum global cost, but can come with exponential computational complexity. An example of this is the K-agent A* which scales exponentially with the number of agents. Efficiency prioritizes fast execution, exemplified by Prioritized Planning~\cite{silver2005cooperative} that assigns sequential movement order to agents, though this approach sacrifices optimality as later agents spend excessive time waiting. Completeness guarantees finding a solution when one exists or correctly reporting failure. Operator Decomposition is an algorithmic approach within the context of MAPF that addresses scalability by breaking the complex multi-agent planning into manageable components called operators. This decomposition allows us to run the A* independently on each operator, thus reducing branching factor. Conflict-based Search~\cite{sharon2015conflict} employs a two-level approach with a high-level conflict tree construction and a low-level constraint-based path re-computation, enabling optimal solutions while managing computational complexity. The Increasing Tree Cost Search~\cite{sharon2013icts} treats MAPG as a lookup problem, maintaining a dictionary of optimal costs and systematically increasing individual agent costs until collision-free solutions emerge.  Optimal Reciprocal Collision Avoidance (ORCA-MAPF)~\cite{dergachev2021distributed} bridges discrete MAPF with continuous collision avoidance by combining A*-generated initial paths with velocity obstacles to select a speed for an agent to enable real-time deadlock resolution. Social-MAPF uses auctions to enable the agents to bid dynamically on paths/resources promoting a more balanced distribution, and thus promotes fairness.

% \noindent \textbf{SMG implications (MAPF).}
% Classical MAPF is discrete, centralized, fully cooperative and observable. In SMGs, it is best used as a {local joint-planning subroutine} on $\mathcal{K}_m(t)$: to approximate a socially optimal joint action at bottlenecks/crossings, or to resolve deadlock by injecting priorities. Continuous-time and non-holonomic feasibility is enforced by a short-horizon motion layer. We can achieve strong fairness guarantees when $\mathcal{K}_m(t)$ is small and communication is available.

\subsubsection{Optimization}
% Optimization-based methods form a central paradigm in multi-agent planning, offering a principled way to generate coordinated, efficient, and safe behaviors for teams of robots.
Optimization-based approaches formulate navigation as a constrained optimal control problem. For each agent $i$, one solves:
\begin{tcolorbox}[mydefinition]
\begin{definition}
\textbf{SMG Optimization:} For the $i^\textrm{th}$ robot, given its initial state $x^i_0$, an optimal trajectory $\Gamma^{i,*}$ and corresponding optimal input sequence $\Psi^{i,*}$ are given by,
\begin{subequations}
\begin{align}
\left( \Gamma^{i,*}, \Psi^{i,*}\right) =& \arg\min_{(\Gamma^i,  \Psi^{i})} \sum_{t={0}}^{T-1} \mathcal{J}^i\left(x^i_t, u^i_t\right) + \mathcal{J}^i_T\left(x^i_N\right) \label{eq: solving_3}\\
\text{s.t}\;\;  x^i_{t+1}=& f\left(x^i_t\right) + g\left(x^i_t\right)u^i_t,\quad\forall t\in[1;T-1]\label{eq: discrete_dynamics}\\
\mathcal{C}^i\left( x^i_t \right) &\cap \mathcal{C}^j\left( x^j_t \right)= \emptyset \ \forall j \in \mathcal{N}^i\left(x^i_t \right) \ \forall t \label{eq: sub_coll}\\
% &x^i_0\in \mathcal{X}_i\\ 
&x^i_T\in \mathcal{X}_g\label{eq: libe_constraint}
\end{align}
\label{eq: best_response_example}
\end{subequations}
\label{def: best_response}
\end{definition}
\end{tcolorbox}

where $\mathcal{J}^i_T$ is the terminal cost.  A solution to optimal navigation in social mini-games is a (finite) sequence of state-input pairs $\left(\left(\Gamma^{1,*}, \Psi^{1,*}\right),\left(\Gamma^{2,*}, \Psi^{2,*}\right),\ldots, \left(\Gamma^{k,*}, \Psi^{k,*}\right)\right )$. In centralized formulations, the optimization is performed jointly over $(u_1,\dots,u_N)$.
In decentralized formulations, collision avoidance appears as constraints coupling agents.

An SMG corresponds to a localized constrained subproblem over agents in $\mathcal{K}_m(t)$, where the decision variables are trajectories or control sequences over a horizon. Constraints encode collision avoidance and possibly resource capacity (e.g., at most one agent inside a doorway region at a time). The coupling set $\mathcal{K}_m(t)$ is determined by agents whose constraints are mutually active. Thus, in optimization-based methods, an SMG is characterized by the activation of shared inequality constraints among a subset of agents over a finite time interval. Optimization-based planners expose SMG structure via explicit constraints
(safety, rules-of-the-road) and multi-objective costs (ego progress versus social
costs). On $\mathcal{K}_m(t)$, adding priority or turn-taking constraints yields
predictable equilibria with low IS and strong fairness; horizons and solver latency
limit $|\mathcal{K}|$ scalability. Communication tightens coupling,
while no-comm variants rely on shared norms encoded as constraints.

Optimization-based approaches frame the planning problem as the minimization (or maximization) of a cost (or reward) function subject to system dynamics and various constraints, such as collision avoidance, kinematic limits, and task requirements. Optimization-based methods underpin key multi-agent collision avoidance strategies. A key metric fundamentally used for these algorithms is of Velocity Obstacle (VO). VO is the set of velocities to causes an agent to collide with other agents, assuming that other agents maintain their current velocities. Reciprocal Velocity Obstacle (RVO)~\cite{berg2008reciprocal} presents a solution by sharing the responsibility of collision avoidance, through adding a correction velocity to both the agents. However, more the correction velocity increases, more an agent is distracted to the side the more conservative the system becomes. Optimal Reciprocal Collision Avoidance (ORCA)~\cite{nh-orca} extends the RVO to general n-body collision avoidance problem. ORCA can also be extended to non-holonomic constraints on linear and the angular velocities of the agents too. While VO-based schemes operate in velocity space and can become conservative in cluttered scenes, an alternative is to reason directly in configuration space by decomposing free space into convex sets that integrate naturally with MPC. Free Space Ellipsoid Graphs \cite{ray2022free} couple a convex decomposition of the workspace with model-predictive control by reusing the same ellipsoids as graph nodes and state constraints, aligning high-level coordination with low-level safety. To pair global convex partitions with fast local reactions, the control update itself can be formulated as a tractable per-agent convex optimization for reciprocal avoidance under uncertainty. CARP \cite{shah2021reciprocal} formulates reciprocal multi-robot collision avoidance under asymmetric state uncertainty as a per-agent convex program and extends it to smooth polynomial trajectories amenable to high-rate onboard execution. Beyond per step convex avoidance, coupling intent inference with a runtime safety layer enables exploratory interaction without sacrificing formal guarantees. Shielding-aware dual-control planner couples implicit stochastic MPC for active intent inference with a runtime safety filter, balancing exploration with guaranteed safety during interaction planning \cite{hu2024active}.

Game theory is the field of mathematical study of strategic interactions and decision-making here each player's action influences the outcome of all the players in the game. Games are categorized by different criteria. Static games are where all players make a single decision without knowing what decisions other players are making. An example of static game is the prisoner's dilemma. On the other hand, dynamic games are sequential in nature. Poker is an example of a dynamic game. A zero-sum game refers to one where one player doing better implies that other players will do worse, while a general sum game refer to those where this does not hold. Wang et al.~\cite{wang2020game} proposed a game-theoretic planning framework where each agent is risk-aware. The approach formulates the multi-agent planning problem as a risk-sensitive dynamic game, where each agent seeks to minimize a risk embedded cost function. The proposed algorithm iteratively approximates the feedback Nash Equilibrium using linearizations of the system dynamics and quadratic approximations of the cost.  Iterative linear quadratic games approximate general-sum dynamic games by repeatedly linearizing the dynamics and quadratizing the costs to compute feedback Nash strategies, yielding interactive policies suitable for multi-robot collision avoidance and merging scenarios \cite{fridovich2020efficient}. A Stackelberg formulation addresses action ordering where branch and play searches over leader follower orders with a mixed-integer planner coupled to trajectory optimization, producing socially efficient, collision free multi-robot plans in crowded interactions \cite{hu2024plays}. To scale stochastic dynamic games, a partial-belief iLQG variant selectively propagates only the most informative beliefs while preserving equilibrium quality interactions, enabling real time multi-agent planning under uncertainty \cite{vakil2024partial}. 

% \noindent \textbf{SMG implications (Optimization/MPC).}
% Optimization-based planners expose SMG structure via explicit constraints
% (safety, rules-of-the-road) and multi-objective costs (ego progress versus social
% costs). On $\mathcal{K}_m(t)$, adding priority or turn-taking constraints yields
% predictable equilibria with low IS and strong fairness; horizons and solver latency
% limit $|\mathcal{K}|$ scalability. Communication tightens coupling,
% while no-comm variants rely on shared norms encoded as constraints.

\subsubsection{Others}

There are several algorithms that do not fall under the category of the three algorithmic categories discussed so far. These contain algorithms which are based on heuristics, evolutionary algorithms, etc. For example, Zhu et al.~\cite{das2023autonomous} introduce a Genetic Algorithm-based Topological Optimisation (GATO) framework for multi-robot logistics in agricultural settings, where the operational environment is represented as a discretized topological map. \cite{turhanlar2021deadlock} proposed a Flexible System Design approach for automated warehouses, where a multi-agent simulation modeling based approach is used for motion planning. The approach proposed in \cite{coskun2021deadlock} is based on coordination space framework, where each robot's trajectory is represented in a high-dimensional space, and collisions are encoded as obstacles in this space. \cite{pratissoli2023hierarchical} proposes a hierarchical method with a three-player control system designed for real-world factories. They employ the use of time-expanded graphs for coordination and deadlock prevention. This class of methods are characterized by their adaptability, scalability and are well-suited for domains like warehousing, logistics and large-scale robotic systems. To complement graph-based coordination, we consider data driven controllers that reason directly over continuous state–action spaces under uncertainty. A GP-based decentralized planner learns other agent's actions and safety envelopes online, enforcing individual and joint safety in continuous spaces without shared policies or centralized control~\cite{zhu2020multi}. Coupling this with stable neighbor selection in prediction improves both learning stability and trajectory quality. A smooth attention prior for multi-agent trajectory prediction stabilizes which neighbors an agent attends to over time, improving sample efficiency and accuracy on naturalistic driving datasets~\cite{cao2022leveraging}.

Local reactive methods (e.g., ORCA variants) are effective for pairwise
interactions but can stall in symmetric SMGs (head-on hallway, doorway) without
{symmetry-breaking} or {priority} rules. Hybrids (e.g., ORCA+local-MAPF)
treat SMG activations as events that trigger a brief centralized or negotiated
joint solve on $\mathcal{K}_m(t)$ to restore liveness and fairness.
% 
% \noindent \textbf{SMG implications (Heuristics/Hybrids).}

\subsection{Taxonomy of SMG Solvers}
\label{subsec: taxonomy}

Beyond high-level paradigms, SMG solvers can also be understood through a taxonomy of system design dimensions: coordination, communication, deadlock handling, invasiveness, cooperation, and observability. Each of these choices shapes how agents in $\mathcal{K}_m(t)$ resolve the strategic coupling that defines an SMG. For example, decentralized no-communication policies may perform adequately in generic MRN but can yield inefficient equilibria in SMGs unless augmented with norms or symmetry-breaking rules. Conversely, explicit communication or centralized coordination within $\mathcal{K}_m(t)$ can lower IS and raise fairness, though at higher computational or infrastructure costs. In the subsections below, we connect each taxonomy axis to SMG outcomes, highlighting how design choices change when evaluated under the framework of SMGs rather than generic MRN.

% Centralized
\begin{figure*}[!t]
    \renewcommand{\arraystretch}{1.4}
    \setlength{\tabcolsep}{10pt}
    \begin{tabular}{>{\centering\arraybackslash}m{0.25\textwidth} | >{\centering\arraybackslash}m{0.75\textwidth}}
        \includegraphics[width=0.90\linewidth]{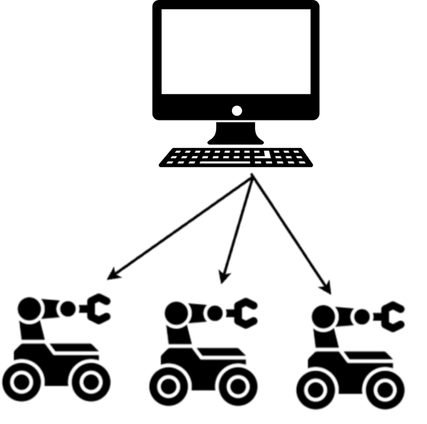}
        &
\begin{minipage}[t]{\linewidth}
  \begin{tabular}{M M M M M M M}  % 8 fixed-width columns
    \cellcolor{green!30}\cite{chung2024deadlock} &
    \cellcolor{blue!30}\cite{chandra2025deadlock} &
    \cellcolor{green!30}\cite{gao2024pce} &
    \cellcolor{cyan!60}\cite{ye2022multi} &
    \cellcolor{blue!30}\cite{arul2023ds} &
    \cellcolor{blue!30}\cite{he2022deadlock} &
    \cellcolor{cyan!60}\cite{fujitani2022deadlock} \\
    \cellcolor{orange!30}\cite{sauer2022decentralized} &
    \cellcolor{blue!30}\cite{chandra2022gameplan} &
    \cellcolor{cyan!60}\cite{wagner2022minimizing} &
    \cellcolor{green!30}\cite{yamauchi2022standby} &
    \cellcolor{green!30}\cite{alonso2018reactive} &
    \cellcolor{green!30}\cite{decastro2018collision} &
    \cellcolor{blue!30}\cite{yong2017cooperative} \\
    \cellcolor{blue!30}\cite{wang2017safety} &
    \cellcolor{green!30}\cite{cirillo2014lattice} &
    \cellcolor{blue!30}\cite{csenbacslar2023mrnav} 
  \end{tabular}
\end{minipage}

    \end{tabular}
\caption{Centralized Methods. \hspace{0.1em}  
\legendbox{cyan!60}\enspace Multi-Agent RL \quad
\legendbox{blue!30}\enspace Optimization \quad
\legendbox{green!30}\enspace MAPF \quad
\legendbox{orange!30}\enspace Others (e.g., Heuristic).}
    \label{fig:overview_centralized}
\end{figure*}

% Decentralized
\begin{figure*}[!t]
    \renewcommand{\arraystretch}{1.4}
    \setlength{\tabcolsep}{10pt}
    \begin{tabular}{>{\centering\arraybackslash}m{0.25\textwidth} | >{\centering\arraybackslash}m{0.75\textwidth}}
        \includegraphics[width=0.90\linewidth]{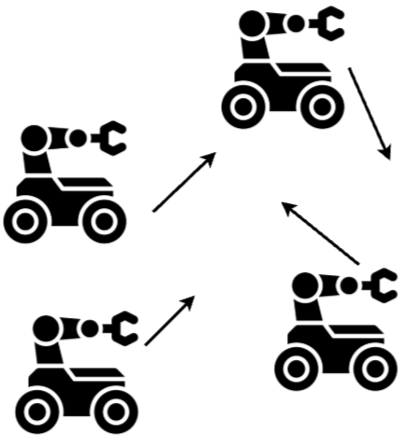}
        &

\begin{minipage}[t]{\linewidth}
  \begin{tabular}{M M M M M M M}  % 8 columns with fixed width
    \cellcolor{blue!30}\cite{garg2024deadlock} &
    \cellcolor{cyan!60}\cite{garg2024large} &
    \cellcolor{cyan!60}\cite{ma2021distributed} &
    \cellcolor{orange!30}\cite{wang2023polynomial} &
    \cellcolor{blue!30}\cite{park2023decentralized} &
    \cellcolor{blue!30}\cite{park2023dlsc} &
    \cellcolor{cyan!60}\cite{muller2023multi} \\
    \cellcolor{blue!30}\cite{chen2023multi} &
    \cellcolor{orange!30}\cite{das2023autonomous} &
    \cellcolor{blue!30}\cite{das2023autonomous} &
    \cellcolor{blue!30}\cite{toumieh2022decentralized} &
    \cellcolor{green!30}\cite{chan2022multi} &
    \cellcolor{green!30}\cite{ye2022multi} &
    \cellcolor{orange!30}\cite{chan2022multi} \\
    \cellcolor{green!30}\cite{zakir2022robot} &
    \cellcolor{cyan!60}\cite{goto2022solving} &
    \cellcolor{blue!30}\cite{grover2023before} &
    \cellcolor{orange!30}\cite{turhanlar2021deadlock} &
    \cellcolor{green!30}\cite{heselden2021crh} &
    \cellcolor{blue!30}\cite{abdullhak2021deadlock} &
    \cellcolor{orange!30}\cite{coskun2021deadlock} \\
    \cellcolor{cyan!60}\cite{ma2021distributed} &
    \cellcolor{cyan!60}\cite{dergachev2021distributed} &
    \cellcolor{blue!30}\cite{kim2021multi} &
    \cellcolor{blue!30}\cite{zhang2021natural} &
    \cellcolor{blue!30}\cite{mannucci2021provably} &
    \cellcolor{orange!30}\cite{pratissoli2023hierarchical} &
    \cellcolor{orange!30}\cite{turhanlar2024autonomous} \\
    \cellcolor{green!30}\cite{chan2022multi} &
    \cellcolor{green!30}\cite{zhou2020distributed} &
    \cellcolor{blue!30}\cite{wu2019collision} &
    \cellcolor{green!30}\cite{das2019sparcas} &
    \cellcolor{green!30}\cite{zhou2018distributed} &
    \cellcolor{green!30}\cite{rao2018trust} &
    \cellcolor{blue!30}\cite{gregoire2018locally} \\
    \cellcolor{green!30}\cite{wang2018trust} &
    \cellcolor{blue!30}\cite{chen2018verifiable} &
    \cellcolor{green!30}\cite{zhou2017collision} &
    \cellcolor{orange!30}\cite{draganjac2016decentralized} &
    \cellcolor{blue!30}\cite{kimmel2016decentralized} &
    \cellcolor{green!30}\cite{vcap2016provably} &
    \cellcolor{green!30}\cite{iftikhar2016resource} \\
    \cellcolor{green!30}\cite{sun2014behavior} &
    \cellcolor{green!30}\cite{wei2014multi} &
    \cellcolor{blue!30}\cite{csenbacslar2021rlss} &
    \cellcolor{blue!30}\cite{csenbacslar2024dream}    
  \end{tabular}
\end{minipage}

    \end{tabular}
    \caption{Decentralized Methods. \hspace{0.1em}  
\legendbox{cyan!60}\enspace Multi-Agent RL \quad
\legendbox{blue!30}\enspace Optimization \quad
\legendbox{green!30}\enspace MAPF \quad
\legendbox{orange!30}\enspace Others (e.g., Heuristic).}
    \label{fig:overview_decentralized}
\end{figure*}

\begin{figure*}[!t]
    \renewcommand{\arraystretch}{1.4}
    \setlength{\tabcolsep}{10pt}
    \begin{tabular}{>{\centering\arraybackslash}m{0.25\textwidth} | >{\centering\arraybackslash}m{0.75\textwidth}}
        \includegraphics[width=0.90\linewidth]{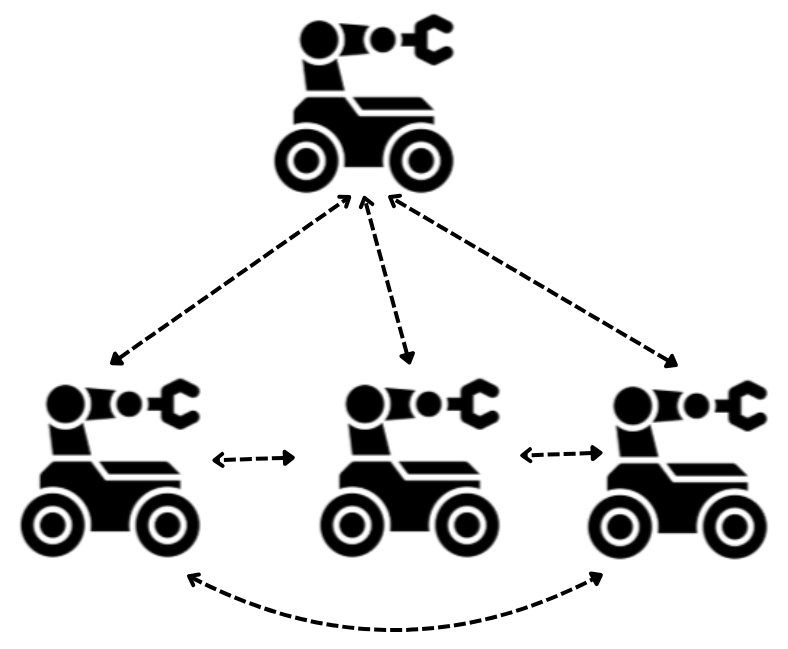}
        &
\begin{minipage}[t]{\linewidth}
  \begin{tabular}{M M M M M M M}  % 8 fixed-width columns
    \cellcolor{cyan!60}\cite{das2019tarmac} &
    \cellcolor{orange!30}\cite{sukhbaatar2016commnet} &
    \cellcolor{blue!30}\cite{wang2020multi} &
    \cellcolor{green!30}\cite{cao2021multi} &
    \cellcolor{blue!30}\cite{csenbacslar2023mrnav} 
  \end{tabular}
\end{minipage}

    \end{tabular}
\caption{Distributed Methods. \hspace{0.1em}  
\legendbox{cyan!60}\enspace Multi-Agent RL \quad
\legendbox{blue!30}\enspace Optimization \quad
\legendbox{green!30}\enspace MAPF \quad
\legendbox{orange!30}\enspace Others (e.g., Heuristic).}
    \label{fig:overview_distributed}
\end{figure*}

\subsubsection{Coordination}
In SMGs, coordination among agents is crucial. Coordination refers to how the agents make decisions and take actions within their environment. It includes the mechanisms through which the motion-planner decisions are made, shared, and implemented. The first type of coordination is \textit{Centralized} which involves a single central authority making decisions for all agents \cite{saha2006multi,li2017centralized}. In this setup, individual agents follow the commands from the central entity, without generating their own controls. This class of planners ensure that all agents work together smoothly and efficiently, as they all follow the same strategy. This kind of coordination is used in methods like Right Hand Rule and Auctions \cite{wang2017safety}, where the central entity generates the command for all the other agents for the safe navigation making sure each agent reaches their goal.

Typical centralized multi-robot pathfinding (MAPF) methods also follow this coordination style. Algorithms like Conflict-Based Search (CBS) \cite{sharon2015conflict} and its suboptimal variant Enhanced CBS (ECBS) \cite{barer2014suboptimal} are widely used, where a central planner coordinates all agents’ paths by resolving conflicts globally. Similarly, methods like M* \cite{wagner2011mstar} and ICTS (Increasing Cost Tree Search) \cite{sharon2013icts} compute joint plans that guarantee collision-free paths for all agents by centrally searching the combined state space. These approaches ensure global consistency and optimality (or bounded suboptimality), but their complexity grows rapidly with the number of agents. Therefore, centralized MAPF methods are highly effective in structured environments with moderate agent counts, offering strong coordination guarantees but requiring powerful centralized computation.

Beyond classical centralized planning approaches, recent research has explored \textit{Distributed} methods that consist of centralized training of agents using reinforcement learning (RL) with communication protocols learned or designed to support coordination \cite{das2019tarmac, sukhbaatar2016commnet}. In this setup, agents are trained together using a shared reward signal and centralized critic, but may execute actions independently with limited or explicit communication. Methods such as those developed by Prorok et al. \cite{cao2021multi,wang2020multi} focus on learning decentralized policies that embed communication for distributed coordination. These approaches differ from traditional MAPF methods by allowing agents to dynamically adapt to new scenarios without explicit replanning. By combining centralized optimization during training with communication-aware decentralized execution, such RL-based methods achieve flexible, robust, and scalable multi-robot coordination in complex environments where centralized control at runtime may not be feasible.

The last type of coordination is \textit{Decentralized} which allows each agent to make its own decisions independently. Instead of relying on a central authority, each agent acts based on its own observations, rewards, and goals. This approach promotes independence and flexibility, enabling agents to adapt to their surroundings and other agents dynamically \cite{azarm1997conflict, demesure2017decentralized}. Although it may seem less organized, with the right algorithms and planners, decentralized coordination can generate effective and efficient controls to develop a safe, collision-free, and deadlock-free trajectory \cite{das2019sparcas, long2018towards, desaraju2012decentralized, maoudj2023improved, toumieh2022decentralized}. Examples of this approach include IMPC-DR \cite{chen2023multi} and ORCA-MAPF \cite{dergachev2021distributed}, where agents independently navigate in their environment while avoiding collisions and deadlocks.
Figures~\ref{fig:overview_centralized} and \ref{fig:overview_decentralized} list the key papers for the Centralized and Decentralized communication categories. Note that the color legend signifies the categorization based on \textit{Algorithmic/Methodological paradigm} discussed in Sec.~\ref{sec:algorithms}, where we use different colors for different approaches- multi-robot Reinforcement Learning (cyan), Optimization (violet), multi-robot Path Finding (green), and Others (Orange).

% legend
% \begin{figure}[!t]
%     \centering
%     \renewcommand{\arraystretch}{1.0}
%     \setlength{\tabcolsep}{2pt}
%     \begin{tabular}{>{\centering\arraybackslash}m{0.8cm} >{\centering\arraybackslash}m{2.2cm} >{\centering\arraybackslash}m{0.8cm} >{\centering\arraybackslash}m{2.2cm}}
%         \cellcolor{cyan!60}\hspace{0.5em} & Multi-Agent RL &
%         \cellcolor{blue!30}\hspace{0.5em} & Optimization \\
%         \cellcolor{green!30}\hspace{0.5em} & MAPF &
%         \cellcolor{orange!30}\hspace{0.5em} & Others (e.g. Heuristic) \\
%     \end{tabular}
%     \caption{\textbf{Color legend across figures:} Categories of MRN Algorithms based on Algorithmic/Methodological Paradigm.}
%     \label{fig:method_legend}
% \end{figure}

%------------------------------------%
% Communicated
%------------------------------------%

\begin{figure*}[!t]
    \renewcommand{\arraystretch}{1.4}
    \setlength{\tabcolsep}{10pt}
    \begin{tabular}{>{\centering\arraybackslash}m{0.25\textwidth} | >{\centering\arraybackslash}m{0.75\textwidth}}
        \includegraphics[width=0.75\linewidth]{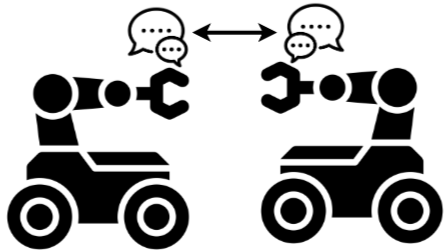}
        &
\begin{minipage}[t]{\linewidth}
  \begin{tabular}{M M M M M M M}  % 8 fixed-width columns
    \cellcolor{green!30}\cite{gao2024pce} &
    \cellcolor{green!30}\cite{maoudj2023improved} &
    \cellcolor{cyan!60}\cite{ye2022multi} &
    \cellcolor{green!30}\cite{ma2021distributed} &
    \cellcolor{blue!30}\cite{garg2024deadlock} &
    \cellcolor{blue!30}\cite{garg2024large} &
    \cellcolor{blue!30}\cite{park2023decentralized} \\
    \cellcolor{orange!30}\cite{pratissoli2023hierarchical} &
    \cellcolor{orange!30}\cite{das2023autonomous} &
    \cellcolor{orange!30}\cite{turhanlar2024autonomous} &
    \cellcolor{blue!30}\cite{he2022deadlock} &
    \cellcolor{cyan!60}\cite{fujitani2022deadlock} &
    \cellcolor{green!30}\cite{chan2022multi} &
    \cellcolor{orange!30}\cite{arif2022robot} \\
    \cellcolor{green!30}\cite{zakir2022robot} &
    \cellcolor{green!30}\cite{yamauchi2022standby} &
    \cellcolor{orange!30}\cite{turhanlar2021deadlock} &
    \cellcolor{cyan!60}\cite{heselden2021crh} &
    \cellcolor{orange!30}\cite{coskun2021deadlock} &
    \cellcolor{blue!30}\cite{zhang2021natural} &
    \cellcolor{blue!30}\cite{mannucci2021provably} \\
    \cellcolor{green!30}\cite{zhou2020distributed} &
    \cellcolor{blue!30}\cite{wu2019collision} &
    \cellcolor{green!30}\cite{das2019sparcas} &
    \cellcolor{orange!30}\cite{alonso2018reactive} &
    \cellcolor{green!30}\cite{wang2018trust} &
    \cellcolor{blue!30}\cite{chen2018verifiable} &
    \cellcolor{green!30}\cite{decastro2018collision} \\
    \cellcolor{blue!30}\cite{yong2017cooperative} &
    \cellcolor{blue!30}\cite{wang2017safety} &
    \cellcolor{orange!30}\cite{draganjac2016decentralized} &
    \cellcolor{green!30}\cite{iftikhar2016resource} &
    \cellcolor{green!30}\cite{cirillo2014lattice} &
    \cellcolor{green!30}\cite{sun2014behavior} &
    \cellcolor{cyan!60}\cite{ma2021distributed} \\
    \cellcolor{blue!30}\cite{csenbacslar2023mrnav} &
    \cellcolor{cyan!60}\cite{das2019tarmac} &
  \end{tabular}
\end{minipage}

    \end{tabular}
    \caption{Communicated Methods. \hspace{0.1em}  
\legendbox{cyan!60}\enspace Multi-Agent RL \quad
\legendbox{blue!30}\enspace Optimization \quad
\legendbox{green!30}\enspace MAPF \quad
\legendbox{orange!30}\enspace Others (e.g., Heuristic).}
    \label{fig:overview_comm}
\end{figure*}

%------------------------------------%
% Un-Communicated
%------------------------------------%

\begin{figure*}[!t]
    \renewcommand{\arraystretch}{1.4}
    \setlength{\tabcolsep}{10pt}
    \begin{tabular}{>{\centering\arraybackslash}m{0.25\textwidth} | >{\centering\arraybackslash}m{0.75\textwidth}}
        \includegraphics[width=0.75\linewidth]{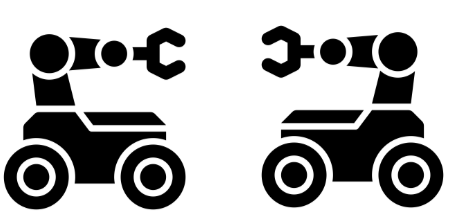}
        &

\begin{minipage}[t]{\linewidth}
  \begin{tabular}{M M M M M M M}
    \cellcolor{green!30}\cite{chung2024deadlock} &
    \cellcolor{blue!30}\cite{chandra2025deadlock} &
    \cellcolor{green!30}\cite{wang2023polynomial} &
    \cellcolor{blue!30}\cite{park2023dlsc} &
    \cellcolor{blue!30}\cite{arul2023ds} &
    \cellcolor{cyan!60}\cite{muller2023multi} &
    \cellcolor{blue!30}\cite{chen2023multi} \\
    \cellcolor{blue!30}\cite{schafer2023rmtruck}  &
    \cellcolor{orange!30}\cite{sauer2022decentralized} &
    \cellcolor{blue!30}\cite{toumieh2022decentralized} &
    \cellcolor{blue!30}\cite{chandra2022gameplan} &
    \cellcolor{green!30}\cite{wagner2022minimizing} &
    \cellcolor{green!30}\cite{chan2022multi} &
    \cellcolor{blue!30}\cite{csenbacslar2021rlss} \\
    \cellcolor{cyan!60}\cite{goto2022solving} &
    \cellcolor{blue!30}\cite{grover2023before} &
    \cellcolor{blue!30}\cite{abdullhak2021deadlock} &
    \cellcolor{green!30}\cite{dergachev2021distributed} &
    \cellcolor{blue!30}\cite{kim2021multi} &
    \cellcolor{green!30}\cite{zhou2018distributed} &
    \cellcolor{green!30}\cite{rao2018trust} \\
    \cellcolor{green!30}\cite{gregoire2018locally} &
    \cellcolor{green!30}\cite{zhou2017collision} &
    \cellcolor{orange!30}\cite{kimmel2016decentralized} &
    \cellcolor{green!30}\cite{vcap2016provably} &
    \cellcolor{green!30}\cite{wei2014multi} &
    \cellcolor{blue!30}\cite{csenbacslar2024dream}
  \end{tabular}
\end{minipage}

    \end{tabular}
    \caption{ Un-Communicated. \hspace{0.1em}  
\legendbox{cyan!60}\enspace Multi-Agent RL \quad
\legendbox{blue!30}\enspace Optimization \quad
\legendbox{green!30}\enspace MAPF \quad
\legendbox{orange!30}\enspace Others (e.g., Heuristic).}
    \label{fig:overview_uncomm}
\end{figure*}

\subsubsection{Communication}
Communication among agents plays a key role in understanding the path planner for deadlock avoidance. Agents can either be \textit{Communicated} or \textit{Uncommunicated}, depending on whether they share information with each other during decision-making. When agents are communicated, they share information and coordinate their actions.  By sharing their plans, intentions, and observations, Communicated agents can work together more effectively. This helps in avoiding conflicts and ensuring smoother navigation, as all agents are aware of each other’s actions and can adjust accordingly. The centralized coordinated methods are generally communicated because the agents have to communicate with the central coordinator/agent for decision-making. In contrast, Uncommunicated agents do not share information with each other. Each agent makes decisions independently, based only on its own observations and goals. While this approach simplifies the communication requirements, it can lead to more conflicts and less coordinated movements, and raise potential deadlocks among agents. However, with a well-designed control strategy, uncommunicated agents can still navigate effectively, relying on their ability to adapt to changing conditions and other agents’ actions. The decentralized methods generally behave in an uncommunicated manner, where each agent makes its own independent decision to generate the controls and methods like  IMPC-DR \cite{chen2023multi} and ORCA-MAPF \cite{dergachev2021distributed} come under this category. Fig~\ref{fig:overview_comm} and \ref{fig:overview_uncomm} categorize the various papers under Communicated and Uncommunicated, respectively.

% Deadlock handling- resolution
\begin{figure*}[!t]
    \renewcommand{\arraystretch}{1.4}
    \setlength{\tabcolsep}{10pt}
    \begin{tabular}{>{\centering\arraybackslash}m{0.25\textwidth} | >{\centering\arraybackslash}m{0.75\textwidth}}
        \includegraphics[width=0.75\linewidth]{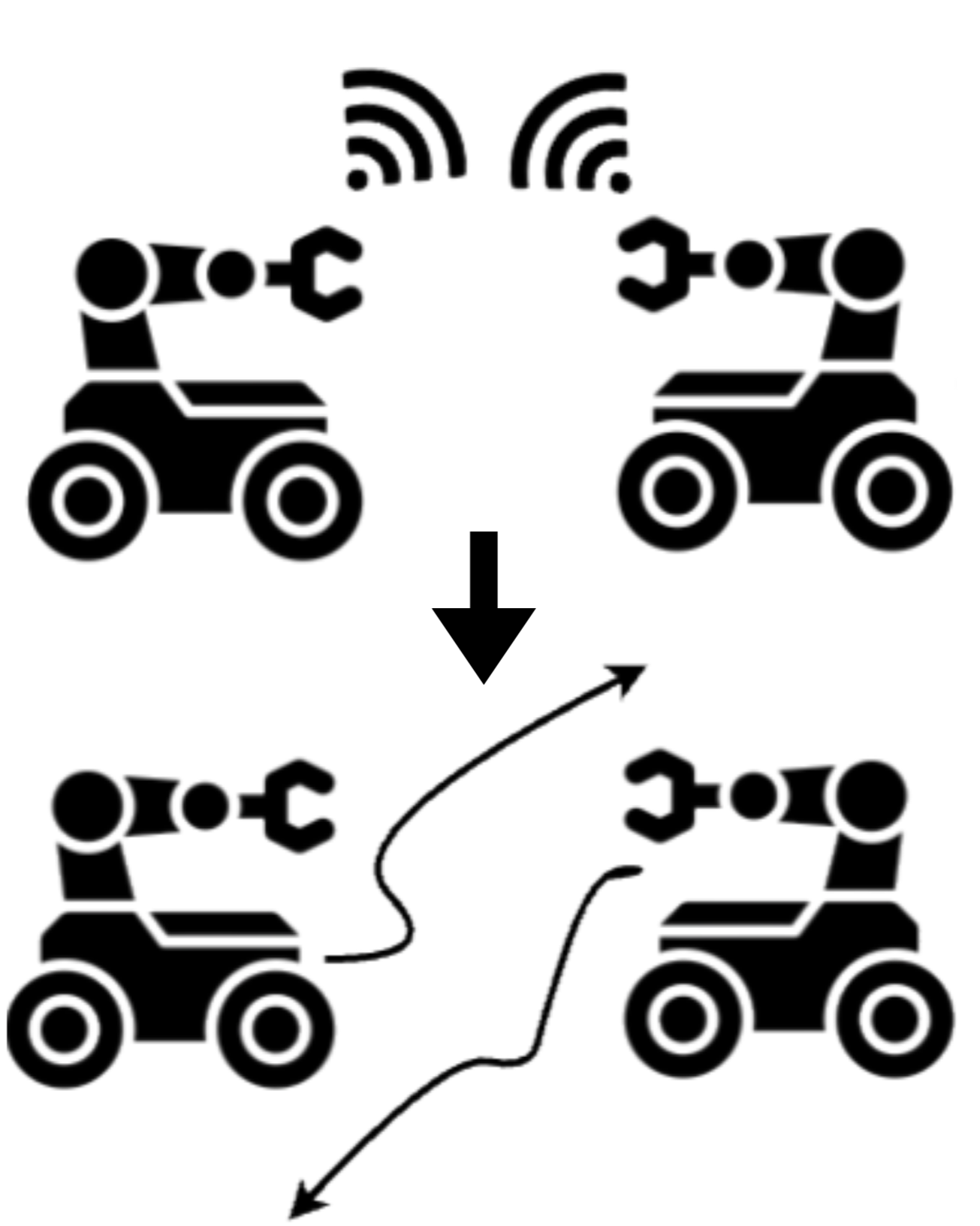}
        &
\begin{minipage}[t]{\linewidth}
  \begin{tabular}{M M M M M M M}
    \cellcolor{blue!30}\cite{garg2024deadlock} &
    \cellcolor{blue!30}\cite{garg2024large} &
    \cellcolor{green!30}\cite{maoudj2023improved} &
    \cellcolor{green!30}\cite{wang2023polynomial} &
    \cellcolor{blue!30}\cite{park2023decentralized} &
    \cellcolor{blue!30}\cite{park2023dlsc} &
    \cellcolor{green!30}\cite{muller2023multi} \\
    \cellcolor{blue!30}\cite{chen2023multi} &
    \cellcolor{orange!30}\cite{das2023autonomous} &
    \cellcolor{blue!30}\cite{schafer2023rmtruck} &
    \cellcolor{blue!30}\cite{toumieh2022decentralized} &
    \cellcolor{green!30}\cite{ye2022multi} &
    \cellcolor{orange!30}\cite{arif2022robot} &
    \cellcolor{green!30}\cite{zakir2022robot} \\
    \cellcolor{blue!30}\cite{grover2023before} &
    \cellcolor{blue!30}\cite{abdullhak2021deadlock} &
    \cellcolor{orange!30}\cite{coskun2021deadlock} &
    \cellcolor{cyan!60}\cite{ma2021distributed} &
    \cellcolor{green!30}\cite{dergachev2021distributed} &
    \cellcolor{blue!30}\cite{kim2021multi} &
    \cellcolor{blue!30}\cite{zhang2021natural} \\
    \cellcolor{orange!30}\cite{alonso2018reactive} &
    \cellcolor{blue!30}\cite{chen2018verifiable} &
    \cellcolor{green!30}\cite{decastro2018collision} &
    \cellcolor{blue!30}\cite{yong2017cooperative} &
    \cellcolor{blue!30}\cite{wang2017safety} &
    \cellcolor{green!30}\cite{iftikhar2016resource} &
    \cellcolor{green!30}\cite{sun2014behavior} \\
    \cellcolor{green!30}\cite{wei2014multi} &
    \cellcolor{blue!30}\cite{csenbacslar2024dream}&
    \cellcolor{blue!30}\cite{csenbacslar2023mrnav} 
    &
    
    \cellcolor{blue!30}\cite{zhang2025adaptive}
    &
    
    \cellcolor{blue!30}\cite{zhang2025deadlock}
  \end{tabular}
\end{minipage}

    \end{tabular}
    \caption{ Deadlock Handling: Resolution. \hspace{0.1em}  
\legendbox{cyan!60}\enspace Multi-Agent RL \quad
\legendbox{blue!30}\enspace Optimization \quad
\legendbox{green!30}\enspace MAPF \quad
\legendbox{orange!30}\enspace Others (e.g., Heuristic).}
    \label{fig:overview_deadlock_resolution}
\end{figure*}

% Deadlock handling- prevention
\begin{figure*}[!t]
    \renewcommand{\arraystretch}{1.4}
    \setlength{\tabcolsep}{10pt}
    \begin{tabular}{>{\centering\arraybackslash}m{0.25\textwidth} | >{\centering\arraybackslash}m{0.75\textwidth}}
        \includegraphics[width=1.0\linewidth]{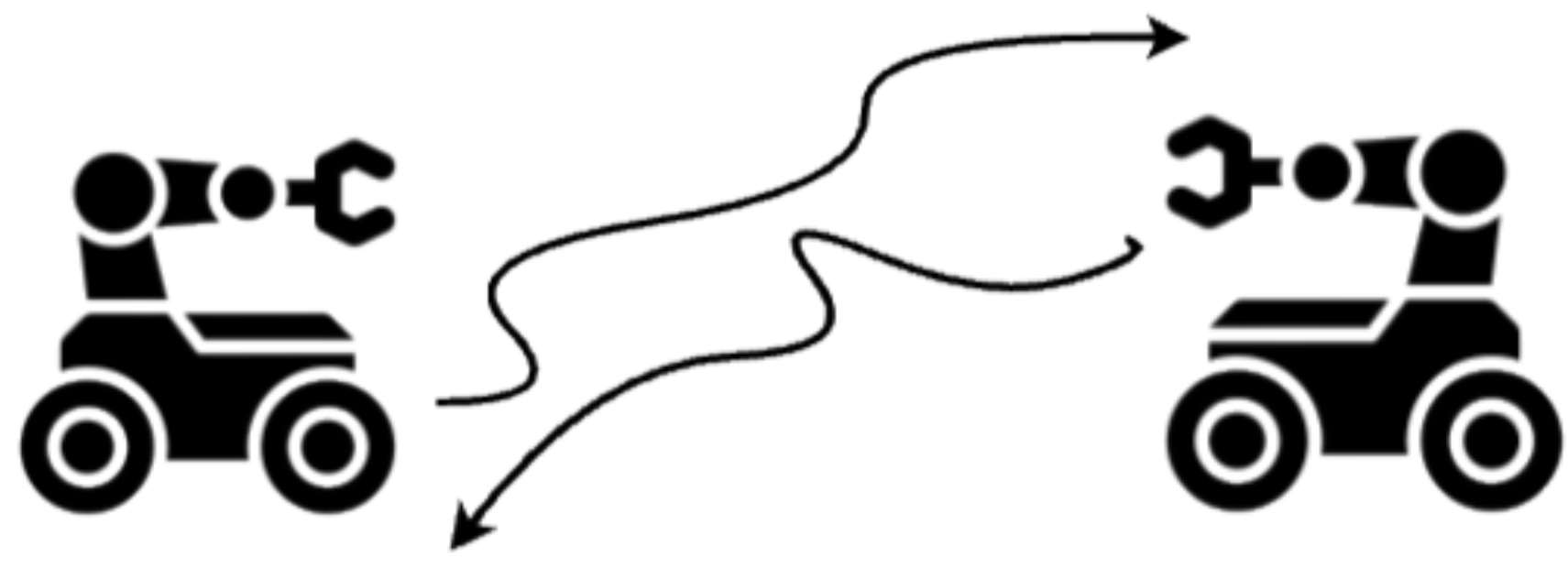}
        &
\begin{minipage}[t]{\linewidth}
  \begin{tabular}{M M M M M M M}
    \cellcolor{green!30}\cite{chung2024deadlock} &
    \cellcolor{blue!30}\cite{chandra2025deadlock} &
    \cellcolor{green!30}\cite{gao2024pce} &
    \cellcolor{cyan!60}\cite{ye2022multi} &
    \cellcolor{blue!30}\cite{arul2023ds} &
    \cellcolor{blue!30}\cite{he2022deadlock} &
    \cellcolor{green!30}\cite{fujitani2022deadlock} \\
    \cellcolor{orange!30}\cite{sauer2022decentralized} &
    \cellcolor{blue!30}\cite{chandra2022gameplan} &
    \cellcolor{green!30}\cite{wagner2022minimizing} &
    \cellcolor{blue!30}\cite{csenbacslar2021rlss} &
    \cellcolor{green!30}\cite{yamauchi2022standby} &
    \cellcolor{orange!30}\cite{turhanlar2024autonomous} &
    \cellcolor{orange!30}\cite{pratissoli2023hierarchical} \\
    \cellcolor{orange!30}\cite{turhanlar2021deadlock} &
    \cellcolor{green!30}\cite{heselden2021crh} &
    \cellcolor{cyan!60}\cite{goto2022solving} &
    \cellcolor{green!30}\cite{chan2022multi} &
    \cellcolor{green!30}\cite{zhou2020distributed} &
    \cellcolor{blue!30}\cite{wu2019collision} &
    \cellcolor{green!30}\cite{das2019sparcas} \\
    \cellcolor{green!30}\cite{zhou2018distributed} &
    \cellcolor{green!30}\cite{rao2018trust} &
    \cellcolor{blue!30}\cite{gregoire2018locally} &
    \cellcolor{green!30}\cite{wang2018trust} &
    \cellcolor{green!30}\cite{zhou2017collision} &
    \cellcolor{orange!30}\cite{draganjac2016decentralized} &
    \cellcolor{orange!30}\cite{kimmel2016decentralized} \\
    \cellcolor{green!30}\cite{vcap2016provably} &
    \cellcolor{green!30}\cite{cirillo2014lattice} 
    &
    \cellcolor{orange!30}\cite{lee2025merry} 
  \end{tabular}
\end{minipage}
    \end{tabular}
    \caption{ Deadlock Handling: Prevention. \hspace{0.1em}  
\legendbox{cyan!60}\enspace Multi-Agent RL \quad
\legendbox{blue!30}\enspace Optimization \quad
\legendbox{green!30}\enspace MAPF \quad
\legendbox{orange!30}\enspace Others (e.g., Heuristic).}
    \label{fig:overview_deadlock_prevention}
\end{figure*}

\subsubsection{Deadlock Handling}
\label{sec:deadlock-handling}

In multi-robot social navigation, how an agent deals with deadlocks is very important. Deadlock often refers to situations where robot states converge to an undesirable equilibrium that causes the robots to stall \cite{grover2023before, zhang2025adaptive, zhang2025deadlock}. Deadlock handling can be divided into two main types: \textit{Prevention} and \textit{Resolution}. Generally, there are two main approaches: \textit{Proactive} (Deadlock Prevention) and \textit{Reactive} (Deadlock Resolution), each with its own style and advantages. The Proactive Strategy
% , on the other hand, 
involves planning ahead. Agents using this approach constantly analyze their environment to predict and prevent potential deadlocks before they happen. This is like a chess player who thinks several moves ahead. 
% \red{
Examples of this approach include IMPC-DR~\cite{chen2023multi} and MERRY-GO-Round \cite{lee2025merry} that combine deadlock resolution strategies with trajectory planning and navigation to ensure smoother movements, thus avoiding potential deadlocks.
% }. 
% Proactive agents aim to avoid problems, ensuring smoother movement. 
However, this requires a good understanding and prediction of the environment, as wrong predictions can lead to unnecessary detours or inefficient paths. The Reactive Strategy,
% \red{
on the other hand
% ,}
, focuses on immediate response. Agents using this method stay alert for potential deadlocks as they move. They don’t act or react until a deadlock is about to happen. Once they detect it, the robots quickly respond to resolve or avoid the deadlock by taking temporarily adjusted trajectories. This strategy deals with problems as they come up. Examples of this approach are methods that include the right-hand rule~\cite{wang2017safety}, auctions~\cite{chandra2023socialmapf}, ORCA-MAPF~\cite{dergachev2021distributed}, and adaptive rotational strategies  \cite{zhang2025adaptive, zhang2025deadlock}. The Reactive Strategy is flexible and can adapt to changes, but it might struggle if many problems occur at once. Fig.~\ref{fig:overview_deadlock_resolution} and \ref{fig:overview_deadlock_prevention} categorize the MRNs for Deadlock Resolution and Prevention, respectively.

% Invasive
\begin{figure*}[!t]
    \renewcommand{\arraystretch}{1.4}
    \setlength{\tabcolsep}{10pt}
    \begin{tabular}{>{\centering\arraybackslash}m{0.25\textwidth} | >{\centering\arraybackslash}m{0.75\textwidth}}
        \hspace*{-0.2cm}\includegraphics[width=1.1\linewidth]{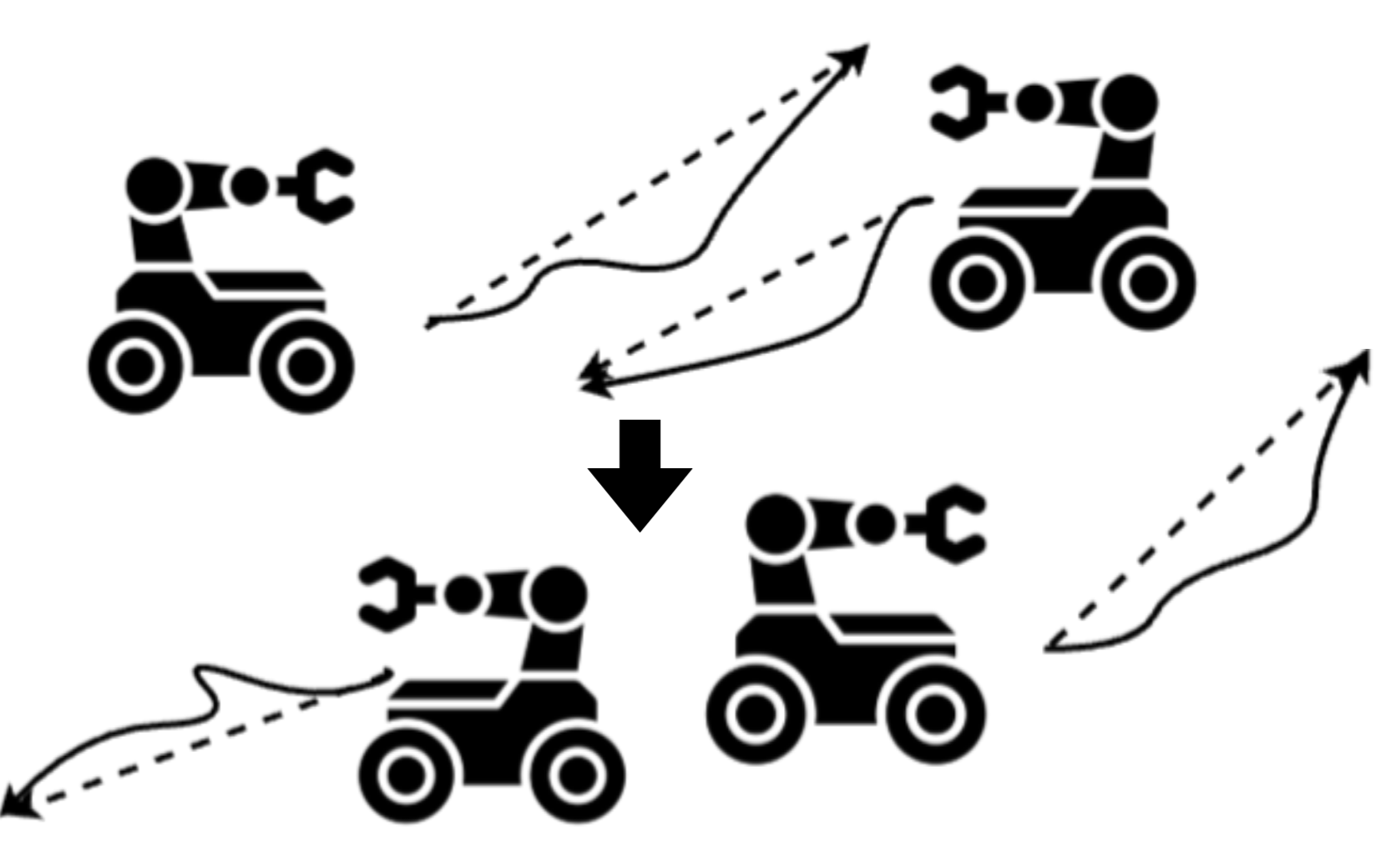}
        &
\begin{minipage}[t]{\linewidth}
  \begin{tabular}{M M M M M M M}
    \cellcolor{blue!30}\cite{garg2024deadlock} &
    \cellcolor{cyan!60}\cite{garg2024large} &
    \cellcolor{cyan!60}\cite{ma2021distributed} &
    \cellcolor{orange!30}\cite{wang2023polynomial} &
    \cellcolor{blue!30}\cite{park2023decentralized} &
    \cellcolor{blue!30}\cite{park2023dlsc} &
    \cellcolor{cyan!60}\cite{muller2023multi} \\
    \cellcolor{blue!30}\cite{chen2023multi} &
    \cellcolor{orange!30}\cite{das2023autonomous} &
    \cellcolor{blue!30}\cite{das2023autonomous} &
    \cellcolor{blue!30}\cite{toumieh2022decentralized} &
    \cellcolor{green!30}\cite{ye2022multi} &
    \cellcolor{orange!30}\cite{chan2022multi} &
    \cellcolor{green!30}\cite{zakir2022robot} \\
    \cellcolor{cyan!60}\cite{goto2022solving} &
    \cellcolor{blue!30}\cite{grover2023before} &
    \cellcolor{blue!30}\cite{abdullhak2021deadlock} &
    \cellcolor{orange!30}\cite{coskun2021deadlock} &
    \cellcolor{cyan!60}\cite{ma2021distributed} &
    \cellcolor{cyan!60}\cite{dergachev2021distributed} &
    \cellcolor{blue!30}\cite{kim2021multi} \\
    \cellcolor{blue!30}\cite{zhang2021natural} &
    \cellcolor{blue!30}\cite{mannucci2021provably} &
    \cellcolor{blue!30}\cite{alonso2018reactive} &
    \cellcolor{blue!30}\cite{chen2018verifiable} &
    \cellcolor{green!30}\cite{decastro2018collision} &
    \cellcolor{blue!30}\cite{yong2017cooperative} &
    \cellcolor{blue!30}\cite{wang2017safety} \\
    \cellcolor{orange!30}\cite{kimmel2016decentralized} &
    \cellcolor{green!30}\cite{iftikhar2016resource} &
    \cellcolor{green!30}\cite{sun2014behavior} &
    \cellcolor{green!30}\cite{wei2014multi} &
    \cellcolor{blue!30}\cite{csenbacslar2021rlss}&
    \cellcolor{blue!30}\cite{csenbacslar2024dream} &
    \cellcolor{blue!30}\cite{csenbacslar2023mrnav} 
  \end{tabular}
\end{minipage}
    \end{tabular}
    \caption{ Invasive Methods. \hspace{0.1em}  
\legendbox{cyan!60}\enspace Multi-Agent RL \quad
\legendbox{blue!30}\enspace Optimization \quad
\legendbox{green!30}\enspace MAPF \quad
\legendbox{orange!30}\enspace Others (e.g., Heuristic).}
    \label{fig:overview_invasive}
\end{figure*}
% Non-invasive
\begin{figure*}[!t]
    \renewcommand{\arraystretch}{1.4}
    \setlength{\tabcolsep}{10pt}
    \begin{tabular}{>{\centering\arraybackslash}m{0.25\textwidth} | >{\centering\arraybackslash}m{0.75\textwidth}}
        \includegraphics[width=1.1\linewidth]{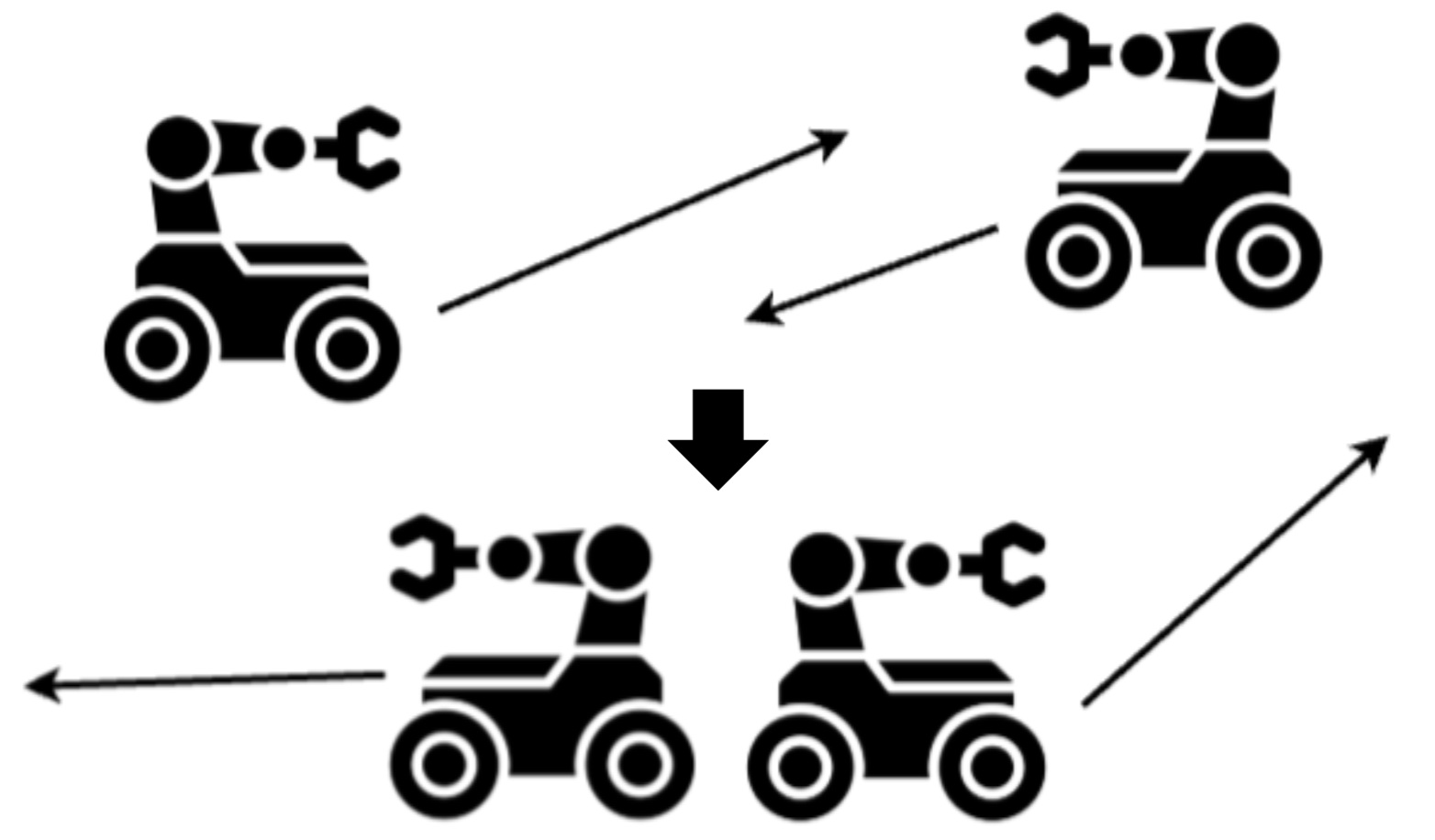}
        &
\begin{minipage}[t]{\linewidth}
  \begin{tabular}{M M M M M M M}
    \cellcolor{green!30}\cite{chung2024deadlock} &
    \cellcolor{blue!30}\cite{chandra2025deadlock} &
    \cellcolor{green!30}\cite{gao2024pce} &
    \cellcolor{cyan!60}\cite{ye2022multi} &
    \cellcolor{blue!30}\cite{arul2023ds} &
    \cellcolor{blue!30}\cite{he2022deadlock} &
    \cellcolor{cyan!60}\cite{fujitani2022deadlock} \\
    \cellcolor{orange!30}\cite{sauer2022decentralized} &
    \cellcolor{blue!30}\cite{chandra2022gameplan} &
    \cellcolor{cyan!60}\cite{wagner2022minimizing} &
    \cellcolor{green!30}\cite{yamauchi2022standby} &
    \cellcolor{orange!30}\cite{turhanlar2024autonomous} &
    \cellcolor{orange!30}\cite{pratissoli2023hierarchical} &
    \cellcolor{orange!30}\cite{turhanlar2021deadlock} \\
        \cellcolor{green!30}\cite{cirillo2014lattice} &
    \cellcolor{green!30}\cite{heselden2021crh} &
    \cellcolor{green!30}\cite{zhou2020distributed} &
    \cellcolor{blue!30}\cite{wu2019collision} &
    \cellcolor{green!30}\cite{das2019sparcas} &
    \cellcolor{green!30}\cite{zhou2018distributed} &
    \cellcolor{green!30}\cite{rao2018trust} \\
    \cellcolor{blue!30}\cite{gregoire2018locally} &
    \cellcolor{green!30}\cite{wang2018trust} &
    \cellcolor{green!30}\cite{zhou2017collision} &
    \cellcolor{orange!30}\cite{draganjac2016decentralized} &
    \cellcolor{green!30}\cite{vcap2016provably} &
    \cellcolor{cyan!60}\cite{das2019tarmac} 
  \end{tabular}
\end{minipage}
    \end{tabular}
    \caption{Non-invasive Methods. \hspace{0.1em}  
\legendbox{cyan!60}\enspace Multi-Agent RL \quad
\legendbox{blue!30}\enspace Optimization \quad
\legendbox{green!30}\enspace MAPF \quad
\legendbox{orange!30}\enspace Others (e.g., Heuristic).}
    \label{fig:overview_noninvasive}
\end{figure*}

\subsubsection{Invasiveness}

Invasiveness is a property of an SMG solver that refers to the degree of {change} to an agent's optimal or preferred trajectory induced by a control input during an SMG. For example, in the doorway example, in order to avoid a deadlock or collision, the agents must react. On the one extreme, a controller may just have an agent come to a complete stop or ``go off to the side and wait'', yielding to the other agent. This is an example of an invasive solver. And if the controller simply tweaks the speed of the agent, without requiring it to deviate from its current trajectory, then the solver is said to be \textit{minimally} invasive. We formalize this as follows:

\begin{tcolorbox}[mydefinition]

\begin{definition}
    \textbf{Minimal Invasiveness.} A control strategy $u^i_t = \left[ v^i_t, \omega^i_t\right]^\top$, where $v^i_t$ is the linear velocity or speed and $\omega^i_t$ is the angular velocity prescribed for robot $i$ at time $t$ with current heading angle $\theta^i_t$, is said to be minimally invasive if:
    \begin{enumerate}
        \item $\Delta \theta^i_t = \theta^i_{t+1} - \theta^i_t = 0$ (does not deviate from the preferred trajectory).
        \item $v_{t+1}^i = v_t^i +\delta_{\text{opt}}(t)$ where $\delta_{\text{opt}}(t) = \arg\min\left\Vert v_t^i + \delta \right\rVert, \delta \in \mathbb{R}$.
    \end{enumerate}
    where $\delta_{\text{opt}}(t)$ is a prescribed velocity perturbation. 

    \label{def: min_invasive}
\end{definition}
\end{tcolorbox}
        % \begin{subequations}
        %         \begin{align}
        % \delta_{\text{opt}}(t)=&\underset{\delta\in\mathbb{R}}{\arg\min}\;\|v^i_t+\delta\|,\\
        % &v^i_{t+1}=v_t^i+\delta\\
        % &u^i_t\in\mathscr{U}^i,\;\;u^i_{t+1}\in\mathscr{U}^i\\
        % &\left(x^i_{t+1},u^i_{t+1}\right)\notin\mathcal{D}^i(t+1)
        % \label{eqn:minimally_invasive_optimization}
        % \end{align}
        %         \end{subequations}
% Note that the optimization problem \eqref{eqn:minimally_invasive_optimization} is a non-trivial problem to solve as the set $\mathcal{D}^i(t)$ is not known priori. Furthermore, even if the set $\mathcal{D}^i(t)$ is known apriori, finding $\delta_{\text{opt}}(t)$ could be computationally expensive especially when the set $\mathcal{D}^i(t)$ is non-convex.

Any solver that is not minimally invasive is generally termed as invasive to varying degrees. A minimally invasive perturbation does not cause a robot to deviate from its preferred trajectory (condition ($1$)) only allowing it to speed up or slow down (condition ($2$)). Invasive SMG solvers significantly change the agent's environment or their paths to avoid deadlocks or collisions. They may also inadvertently alter the paths or influence the movements of other agents. This approach ensures that the agent can navigate smoothly, but it can disrupt the environment and other agents’ plans.

IMPC-DR~\cite{chen2023multi} and Right Hand Rule ~\cite{wang2017safety} are the primary example for this category where a fixed rule-based approach necessitates agents to follow a specific order, which may cause agents who are moving fast but may have lower priority to slow down suddenly to prevent/resolve the deadlock. Examples of minimally invasive solvers include Auctions and ORCA-MAPF~\cite{dergachev2021distributed} where the agents adjust their speeds by the smallest amount needed to avoid potential deadlocks and collisions without deviating from their path. Fig.~\ref{fig:overview_invasive} and \ref{fig:overview_noninvasive} categorizes the different MRN algorithms in the Invasive and Non-invasive categories, respectively.

\subsubsection{Cooperation}
\begin{figure*}[!t]
    \renewcommand{\arraystretch}{1.4}
    \setlength{\tabcolsep}{10pt}
    \begin{tabular}{>{\centering\arraybackslash}m{0.25\textwidth} | >{\centering\arraybackslash}m{0.75\textwidth}}
        \includegraphics[width=1.0\linewidth]{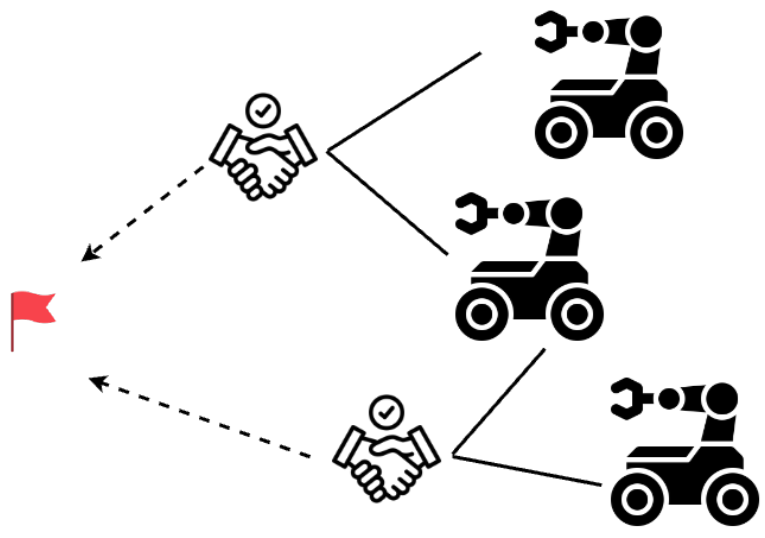}
        &
\begin{minipage}[t]{\linewidth}
  \begin{tabular}{M M M M M M M}
    \cellcolor{blue!30}\cite{everett2019primal} &
    \cellcolor{blue!30}\cite{ma2016optimal} &
    \cellcolor{green!30}\cite{chung2024deadlock} &
    \cellcolor{green!30}\cite{gao2024pce} &
    \cellcolor{blue!30}\cite{das2023autonomous} &
    \cellcolor{cyan!60}\cite{ye2022multi} &
    \cellcolor{orange!30}\cite{patwardhan2022distributing} \\
    \cellcolor{blue!30}\cite{chen2018verifiable} &
    \cellcolor{green!30}\cite{amador2014dynamic} &
    \cellcolor{green!30}\cite{kim2021multi}&
    \cellcolor{orange!30}\cite{freda20193d} &
    \cellcolor{blue!30}\cite{yong2017cooperative} &
    \cellcolor{cyan!60}\cite{sadhu2017improving} &
    \cellcolor{green!30}\cite{ma2017multi} \\
    \cellcolor{orange!30}\cite{turhanlar2021deadlock} &
    \cellcolor{green!30}\cite{cirillo2014lattice} &
    \cellcolor{green!30}\cite{arif2022robot}&
    \cellcolor{orange!30}\cite{zakir2022robot}
  \end{tabular}
\end{minipage}

    \end{tabular}
\caption{Fully Cooperative. Agents share identical algorithms and cost functions. \hspace{0.1em}  
\legendbox{cyan!60}\enspace Multi-Agent RL \quad
\legendbox{blue!30}\enspace Optimization \quad
\legendbox{green!30}\enspace MAPF \quad
\legendbox{orange!30}\enspace Others (e.g., Heuristic).}
\label{fig:overview_fullycooperative}

\end{figure*}

\begin{figure*}[!t]
    \renewcommand{\arraystretch}{1.4}
    \setlength{\tabcolsep}{10pt}
    \begin{tabular}{>{\centering\arraybackslash}m{0.25\textwidth} | >{\centering\arraybackslash}m{0.75\textwidth}}
        \includegraphics[width=1.0\linewidth]{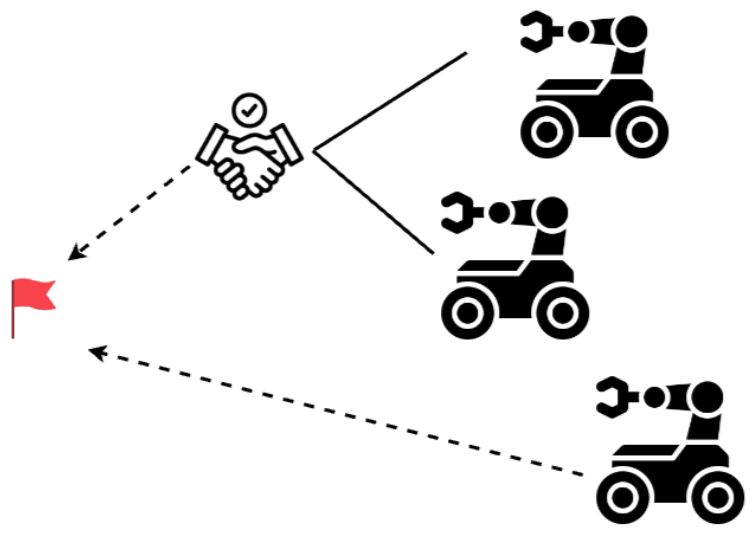}
        &
\begin{minipage}[t]{\linewidth}
  \begin{tabular}{M M M M M M M}
    \cellcolor{blue!30}\cite{chandra2025deadlock} &
    \cellcolor{cyan!60}\cite{garg2024large} &
    \cellcolor{blue!30}\cite{zheng2024mcca} &
    \cellcolor{green!30}\cite{maoudj2023improved} &
    \cellcolor{orange!30}\cite{wang2023polynomial} &
    \cellcolor{cyan!60}\cite{park2022decentralized} &
    \cellcolor{blue!30}\cite{chen2024deadlock} \\
    \cellcolor{blue!30}\cite{park2023dlsc} &
    \cellcolor{cyan!60}\cite{xie2023drl} &
    \cellcolor{blue!30}\cite{arul2023ds} &
    \cellcolor{cyan!60}\cite{muller2023multi} &
    \cellcolor{blue!30}\cite{schafer2023rmtruck} &
    \cellcolor{blue!30}\cite{chen2023multi} &
    \cellcolor{orange!30}\cite{turhanlar2024autonomous} \\
    \cellcolor{blue!30}\cite{he2022deadlock} &
    \cellcolor{cyan!60}\cite{fujitani2022deadlock} &
    \cellcolor{blue!30}\cite{park2023decentralized} &
    \cellcolor{blue!30}\cite{toumieh2022decentralized} &
    \cellcolor{blue!30}\cite{chandra2022gameplan} &
    \cellcolor{cyan!60}\cite{wagner2022minimizing} &
    \cellcolor{blue!30}\cite{zhang2021natural} \\
    \cellcolor{blue!30}\cite{mannucci2021provably} &
    \cellcolor{blue!30}\cite{cappello2020hybrid} &
    \cellcolor{orange!30}\cite{mavrogiannis2020decentralized} &
    \cellcolor{green!30}\cite{zhou2020distributed} &
    \cellcolor{green!30}\cite{pianpak2019distributed} &
    \cellcolor{blue!30}\cite{grover2022semantically} &
    \cellcolor{green!30}\cite{das2019sparcas} \\
    \cellcolor{green!30}\cite{zhou2018distributed} &
    \cellcolor{green!30}\cite{rao2018trust} &
    \cellcolor{green!30}\cite{wang2018trust} &
    \cellcolor{blue!30}\cite{alonso2018reactive} &
    \cellcolor{blue!30}\cite{ gregoire2018locally} &
    \cellcolor{green!30}\cite{zhou2017collision} &
    \cellcolor{green!30}\cite{decastro2018collision} \\
    \cellcolor{blue!30}\cite{wang2017safety} &
    \cellcolor{orange!30}\cite{draganjac2016decentralized} &
    \cellcolor{orange!30}\cite{kimmel2016decentralized} &
    \cellcolor{green!30}\cite{vcap2016provably} &
    \cellcolor{green!30}\cite{iftikhar2016resource} &
    \cellcolor{blue!30}\cite{wang2016safety} &
    \cellcolor{green!30}\cite{sun2014behavior} \\
    \cellcolor{green!30}\cite{wei2014multi} &
    \cellcolor{orange!30}\cite{desaraju2012decentralized} &
    \cellcolor{orange!30}\cite{matsuura1998deadlock} &
    \cellcolor{blue!30}\cite{csenbacslar2021rlss} &
    \cellcolor{blue!30}\cite{csenbacslar2024dream} & 
    \cellcolor{blue!30}\cite{csenbacslar2023mrnav} &
    \cellcolor{cyan!60}\cite{das2019tarmac} 
  \end{tabular}
\end{minipage}

    \end{tabular}
\caption{Semi-Cooperative. Agents share identical algorithms but non-identical cost functions. \hspace{0.1em}  
\legendbox{cyan!60}\enspace Multi-Agent RL \quad
\legendbox{blue!30}\enspace Optimization \quad
\legendbox{green!30}\enspace MAPF \quad
\legendbox{orange!30}\enspace Others (e.g., Heuristic).}
    \label{fig:overview_semicooperative}
\end{figure*}

Cooperation refers to how agents work together to achieve smooth and efficient movement. Cooperation can be characterized based on whether agents share the same/different objectives (cost functions) and whether they use the same or different planning strategies. In fully \textit{Cooperative} settings, agents optimize a common cost function, working toward shared goals through coordinated planning. Methods like PRIMAL \cite{everett2019primal} and CBM \cite{ma2016optimal} follow this style, using explicit communication or negotiation to achieve joint plans. In \textit{Non-Cooperative} settings, agents may have distinct cost functions and prioritize their own objectives, requiring implicit cooperation through local observations. Implicit (Semi) cooperation methods, such as ORCA-MAPF and Reciprocal Velocity Obstacles (RVO)~\cite{van2011reciprocal}, adaptively adjust trajectories based on predicted collisions without explicit communication. This relationship between cost and planning strategies (algorithms) are organized as shown in Fig~\ref{fig:overview_fullycooperative} and Fig~\ref{fig:overview_semicooperative}. Understanding these structures is key to designing socially compliant and deadlock-free multi-robot systems. \\
%---------------------------------------------------
% fully cooperative
%---------------------------------------------------
%---------------------------------------------------
% Semi-cooperative
%---------------------------------------------------
\subsubsection{Observability}
\begin{figure*}[!t]
    \renewcommand{\arraystretch}{1.4}
    \setlength{\tabcolsep}{10pt}
    \begin{tabular}{>{\centering\arraybackslash}m{0.25\textwidth} | >{\centering\arraybackslash}m{0.75\textwidth}}
    \includegraphics[width=1.0\linewidth]{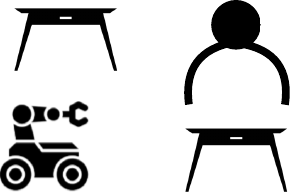}
        &
\begin{minipage}[t]{\linewidth}
  \begin{tabular}{M M M M M M M}
\cellcolor{green!30}\cite{chung2024deadlock} &
\cellcolor{cyan!60}\cite{garg2024large} &
\cellcolor{blue!30}\cite{das2023autonomous} &
\cellcolor{blue!30}\cite{chen2023multi} &
\cellcolor{blue!30}\cite{schafer2023rmtruck} &
\cellcolor{blue!30}\cite{he2022deadlock} &
\cellcolor{cyan!60}\cite{muller2023multi} \\
\cellcolor{orange!30}\cite{turhanlar2021deadlock} &
\cellcolor{green!30}\cite{pianpak2019distributed} &
\cellcolor{green!30}\cite{das2019sparcas} &
\cellcolor{green!30}\cite{wang2018trust} &
\cellcolor{blue!30}\cite{chen2018verifiable} &
\cellcolor{blue!30}\cite{yong2017cooperative} &
\cellcolor{green!30}\cite{ma2017multi} \\
\cellcolor{green!30}\cite{vcap2016provably} &
\cellcolor{green!30}\cite{iftikhar2016resource} &
\cellcolor{green!30}\cite{cirillo2014lattice} &
\cellcolor{orange!30}\cite{wang2023polynomial} &
\cellcolor{green!30}\cite{arif2022robot} &
\cellcolor{orange!30}\cite{freda20193d} &
\cellcolor{cyan!60}\cite{wagner2022minimizing} \\
\cellcolor{blue!30}\cite{everett2019primal} &
\cellcolor{blue!30}\cite{ma2016optimal} &
\cellcolor{green!30}\cite{chung2024deadlock} &
\cellcolor{blue!30}\cite{das2023autonomous} &
\cellcolor{green!30}\cite{arif2022robot} &
\cellcolor{orange!30}\cite{turhanlar2021deadlock} &
\cellcolor{green!30}\cite{amador2014dynamic} \\
\cellcolor{green!30}\cite{kim2021multi} &
\cellcolor{orange!30}\cite{freda20193d} &
\cellcolor{blue!30}\cite{chen2018verifiable} &
\cellcolor{green!30}\cite{cirillo2014lattice} &
\cellcolor{blue!30}\cite{yong2017cooperative} &
\cellcolor{cyan!60}\cite{sadhu2017improving} &
\cellcolor{blue!30}\cite{csenbacslar2023mrnav} \\
\cellcolor{blue!30}\cite{csenbacslar2024dream} &
\cellcolor{blue!30}\cite{csenbacslar2023mrnav} &
  \end{tabular}
\end{minipage}

    \end{tabular}
    \caption{Full Observability. The image on the left shows that the agent (robot) is fully aware of the other agents (humans) and the environment (tables).\hspace{0.1em}  
\legendbox{cyan!60}\enspace Multi-Agent RL \quad
\legendbox{blue!30}\enspace Optimization \quad
\legendbox{green!30}\enspace MAPF \quad
\legendbox{orange!30}\enspace Others (e.g., Heuristic).}
    \label{fig:overview_fullobservability}
\end{figure*}

\begin{figure*}[!t]
    \renewcommand{\arraystretch}{1.4}
    \setlength{\tabcolsep}{10pt}
    \begin{tabular}{>{\centering\arraybackslash}m{0.25\textwidth} | >{\centering\arraybackslash}m{0.75\textwidth}}
        \includegraphics[width=1.0\linewidth]{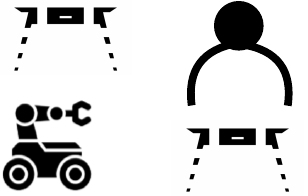}
        &
\begin{minipage}[t]{\linewidth}
  \begin{tabular}{M M M M M M M}

\cellcolor{blue!30}\cite{chandra2025deadlock} &
\cellcolor{blue!30}\cite{zheng2024mcca} &
\cellcolor{green!30}\cite{gao2024pce} &
\cellcolor{green!30}\cite{maoudj2023improved} &
\cellcolor{orange!30}\cite{mavrogiannis2020decentralized} &
\cellcolor{blue!30}\cite{chen2023multi} &
\cellcolor{blue!30}\cite{chen2024deadlock} \\
\cellcolor{blue!30}\cite{park2023decentralized} &
\cellcolor{blue!30}\cite{park2023dlsc} &
\cellcolor{cyan!60}\cite{xie2023drl} &
\cellcolor{orange!30}\cite{patwardhan2022distributing} &
\cellcolor{green!30}\cite{yamauchi2022standby} &
\cellcolor{blue!30}\cite{grover2023before} &
\cellcolor{cyan!60}\cite{everett2021collision} \\
\cellcolor{orange!30}\cite{matsuura1998deadlock} &
\cellcolor{cyan!60}\cite{ma2021distributed} &
\cellcolor{cyan!60}\cite{dergachev2021distributed} &
\cellcolor{green!30}\cite{zhou2020distributed} &
\cellcolor{green!30}\cite{zhou2018distributed} &
\cellcolor{green!30}\cite{rao2018trust} &
\cellcolor{blue!30}\cite{gregoire2018locally} \\
\cellcolor{blue!30}\cite{alonso2018reactive} &
\cellcolor{blue!30}\cite{wang2016safety} &
\cellcolor{blue!30}\cite{wang2017safety} &
\cellcolor{green!30}\cite{sun2014behavior} &
\cellcolor{orange!30}\cite{desaraju2012decentralized} &
\cellcolor{blue!30}\cite{chandra2022gameplan} &
\cellcolor{orange!30}\cite{zakir2022robot} \\
\cellcolor{cyan!60}\cite{ye2022multi} &
\cellcolor{blue!30}\cite{zhang2021natural} &
\cellcolor{blue!30}\cite{mannucci2021provably} &
\cellcolor{blue!30}\cite{cappello2020hybrid} &
\cellcolor{cyan!60}\cite{park2022decentralized} &
\cellcolor{blue!30}\cite{arul2023ds} &
\cellcolor{orange!30}\cite{turhanlar2024autonomous} \\
\cellcolor{cyan!60}\cite{fujitani2022deadlock} &
\cellcolor{blue!30}\cite{toumieh2022decentralized} &
\cellcolor{blue!30}\cite{grover2022semantically} &
\cellcolor{green!30}\cite{zhou2017collision} &
\cellcolor{green!30}\cite{decastro2018collision} &
\cellcolor{orange!30}\cite{draganjac2016decentralized} &
\cellcolor{orange!30}\cite{kimmel2016decentralized} \\
\cellcolor{green!30}\cite{wei2014multi} &
\cellcolor{blue!30}\cite{csenbacslar2021rlss} &
\cellcolor{cyan!60}\cite{das2019tarmac} &

  \end{tabular}
\end{minipage}

    \end{tabular}
    \caption{ Local Observability. The image shows that the agent (robot) is aware of the other agents, but not the environment (table).\hspace{0.1em}  \\
\legendbox{cyan!60}\enspace Multi-Agent RL \quad
\legendbox{blue!30}\enspace Optimization \quad
\legendbox{green!30}\enspace MAPF \quad
\legendbox{orange!30}\enspace Others (e.g., Heuristic).} 
    \label{fig:overview_localobservability}
\end{figure*}

Observability describes how much information an agent has about other agents in the environment. Different levels of observability influence how agents plan and coordinate their actions. Under \textit{Full Observability}, agents have complete knowledge of all other agents' states, including positions, velocities, and goals, enabling highly coordinated planning, as seen in centralized methods like CBS \cite{sharon2015conflict}, ECBS \cite{barer2014suboptimal} and other MAPF solvers (Categorization in Fig.~\ref{fig:overview_fullobservability}). In \textit{Local Observability}, agents rely only on nearby sensing to detect others, as used in decentralized methods such as ORCA, Reciprocal Velocity Obstacles (RVO) \cite{berg2008reciprocal}, and SIPP \cite{phillips2011sipp} (see Fig.~\ref{fig:overview_localobservability}). \textit{Predicted Observability} refers to agents forecasting others' future movements using internal models, often applied in motion prediction frameworks like Social-LSTM \cite{alahi2016social}. This is where the agents are fully aware of the environment, but partially aware of other agents. Belief-space observability involves maintaining probabilistic beliefs over others’ states or intents, which is commonly addressed through POMDP-based planners \cite{somani2013despot} or uncertainty-aware imitation learning. Finally, under \textit{Minimal Observability}, agents do not explicitly model others and treat them as static or reactive obstacles, simplifying planning but limiting social awareness. In this case, the agents are neither aware of other agents nor the environment. Choosing the appropriate observability level is crucial for balancing system complexity, coordination ability, and robustness in dynamic social environments.

%---------------------------------------------------
% fully Observable
%---------------------------------------------------

%---------------------------------------------------
% locally Observable
%---------------------------------------------------

\subsection{Properties of MRN in SMGs}
\label{subsec: properties}

MRN in SMGs is particularly challenging, since more agents are interacting more intricately in less space. Consequently, to evaluate and compare successful MRN solutions, it is important to highlight properties that are desirable. First, it is imperative to avoid collisions and maintain \textit{Safety}, which becomes a primary concern when robots attempt to pass through a narrow corridor or attempt to negotiate an intersection simultaneously. Next, roboticists frequently observe in practice that in the interest of being safe, most robots choose not to move at all, making no progress towards the goal, thereby leading to a deadlock. So we also need to consider \textit{Liveness}. Third, robots can be heterogeneous with different priorities. As such, it is important to ensure that the higher priority robot get the right of way during conflict resolution, so we need to ensure MRN solvers imbibe \textit{Fairness}. Next, robots need to be aware of how they move, for instance, robots can be both safe and deadlock-free, but they can still move in ways that annoys people around them, so robots must exhibit \textit{Socially Compliance}. Finally, we envision a world where teams of tens of hundreds of robots are navigating and co-existing in complex spaces. So we will discuss \textit{Scalability}.

\subsubsection{Safety}
Provable safety can be achieved by single-integrator systems \textit{e.g.} ORCA framework from Van Den Berg et al.~\cite{van2011reciprocal} and its non-holonomic variant~\cite{nh-orca}, which are effective for fast and exact multi-robot navigation. ORCA conservatively imposes collision avoidance constraints on the motion of a robot in terms of half-planes in the space of velocities. The optimal collision-free velocity can then be quickly found by solving a linear program. 
% The original framework limits itself to holonomic systems but has been extended in~\cite{nh-orca} to model non-holonomic constraints with differential drive dynamics. ORCA also generates collision-free velocities that deviate minimally from the robots' preferred velocities. The major limitation of the ORCA framework is that the structure of the half-planes so constructed often results in deadlocks~\cite{dergachev2021distributed}.
Proving safety is harder for systems with double-integrator dynamics, therefore safety in these systems depends on the planning frequency of the system. For example, the NH-TTC algorithm~\cite{davis2019nh} uses gradient descent to minimize a cost function comprising a goal reaching term and a time-to-collision term, which rises to infinity as the agent approaches immediate collision. NH-TTC guarantees safety in the limit as the planning frequency approaches infinity. Other optimization-based approaches use Model Predictive Control(MPC); in such approaches, safety depends not only on the planning frequency but also on the length of the planning horizon. Finally, Control Barrier Functions (CBFs)~\cite{wang2017safety} can be used to design controllers that can guarantee safety via the notion of forward invariance of a set \textit{i.e.} if an agent starts out in a safe set at the initial time step, then it remains safe for all future time steps, that is, it will never leave the safe set. CBFs can also be extended to a probabilistic setting \cite{luo2020multi} to account for realistic uncertainties in system dynamics and observations.

\subsubsection{Liveness} 

Liveness is a property that means robots are continuously making progress towards their goals, and only stop once they have reached their goals. In general MRN, liveness is typically considered a ``soft constraint'' as opposed to safety, which is often a hard constraint. However, MRN in SMGs often result in deadlocks, so they become critical. Methods for ensuring liveness are characterized along the taxonomy can either be deadlock-preventing~\cite{chandra2025deadlock, gouru2024livenet, chen2025livepoint} or deadlock resolving~\cite{wang2017safety, chen2024deadlock, grover2023before, grover2022semantically} (previously discussed in Sec.~\ref{sec:deadlock-handling}). The difference lies in when they solve a deadlock. Deadlock prevention methods are generally smoother and less invasive as they do not wait for a deadlock to happen in the first place. Next, certain liveness methods rely on agents communicating between themselves to resolve conflicts~\cite{garg2024deadlock, mahadevan2025gamechat, garg2024large}. Finally, some liveness methods use global planning methods to plan non-colliding paths~\cite{dergachev2021distributed}.

% Liveness is a fundamental property in multi-robot social navigation, referring to the ability of agents to continually make progress toward their goals. In the context of social mini-games, such as navigation through doorways, hallways, and other interactive environments, liveness ensures that agents do not remain stuck or enter states of indefinite inactivity. Maintaining liveness is critical for the overall efficiency and functionality of the system, as it prevents deadlocks and prolonged stagnation that could disrupt task completion. The importance of liveness lies in its role in enabling agents to adapt to dynamic and potentially congested environments while still achieving their objectives. Strategies to promote liveness include dynamic re-planning \cite{chandra2025deadlock, chandra2022game}, where agents adjust their paths based on the evolving positions of others; the implementation of negotiation and priority mechanisms to resolve movement conflicts; and deadlock detection and resolution methods that allow agents to recognize and recover from non-progress situations. By ensuring liveness, multi-robot systems can operate smoothly, adaptively, and reliably, thereby supporting continuous and effective navigation even in complex, crowded settings.

\subsubsection{Welfare Maximizing}
Social welfare is an important property in SMGs that takes into account the priorities of the agents, along with the objective of the global system designer or society as a whole. Typically, social welfare is defined as a weighted linear combination of agents' individual priorities as well as their local objective functions. For example, consider an ambulance robot and a delivery robot arrive at an intersection. From a safety and liveness perspective, there are two solutions (ambulance goes first followed by the delivery robot, and vice-versa). However, the ambulance clearly has the right of way, and so from a social welfare perspective, the only correct solution is for the ambulance robot to go through first. Welfare maximization is a difficult property to achieve as the individual priority levels are private and are typically unknown to other robots, and must instead be inferred. Additionally, maximizing welfare and individual agent priorities typically are at odds with each other. Welfare maximization can be achieved through mechanisms that balance priorities~\cite{chandra2023socialmapf, suriyarachchi2022gameopt, chandra2022gameplan}, or fixed, agreed-upon rule-based decision-making such as the right-hand-rule~\cite{zhou2017fast, wang2017safety} that ensures all agents have a reasonable opportunity to make progress toward their goals. Welfare maximization has also been studied as a credit-assignment problem in MARL~\cite{nagpal2025leveraging, zhou2020learning, rahmattalabi2016d++, nguyen2018credit, feng2022multi}.

\subsubsection{Social Compliance} 
Social compliance refers to the ability of agents to navigate in a manner that respects social norms, conventions, and expectations within shared environments. Many simulators~\cite{sprague2023socialgym, holtz2022socialgym} and datasets~\cite{scanddataset} have been proposed to develop methods that ensure social compliance. Social robot navigation is a vast area of research and we refer the reader to surveys in this area for a comprehensive treatment~\cite{francis2025principles}. Strategies to promote social compliance include modeling human behavior through imitation learning~\cite{karnan2022voila, xiao2022learning, park2023learning, rosmann2017online}, rule-based systems that encode social conventions~\cite{holtz2021iterative}, and reinforcement learning approaches that reward socially acceptable actions~\cite{chen2017socially, chen2017decentralized}. Recently, research has also started to study hybrid methods that combine imitation learning and classical methods~\cite{raj2024rethinking}.

\subsubsection{Scalability}

% \paragraph{Scalability (SMG view).}
Scalability refers to an MRN solver’s ability to handle a growing number of robots without a loss in performance. This property is especially important in fields like warehouses and logistics, where many robot agents need to work together efficiently. Solvers that analytically prevent or resolve deadlocks are often constrained to a limited number of robots~\cite{wang2016safety, chandra2025deadlock, grover2022semantically, grover2023before}. On the other hand, multi-robot pathfinding algorithms operate in discrete space and discrete time and many efficient and fast solvers have been developed that are able to handle over thousands of robots~\cite{yan2025advancing}. We evaluate scalability in the SMG regime with respect to the {active coupling set size}
$\mathcal{K}_m(t)$, not total $k$. Many solvers scale to large $k$ when SMGs are
rare/small, but performance hinges on how cost/latency grow with $\mathcal{K}_m(t)$
during trigger activations (doorways, crossings). Reporting outcomes (SWG, TTF, IS, DR) stratified by $\mathcal{K}_m(t)$ separates algorithmic scaling from scene density.

% \section{Evaluation in SMGs}
\section{Representative SMG Solvers}
\label{sec: Evaluation}
\begin{figure}[!htbp]
    \centering
    % Row 1
    \begin{subfigure}{0.48\linewidth}
        \centering
        \includegraphics[width=\linewidth]{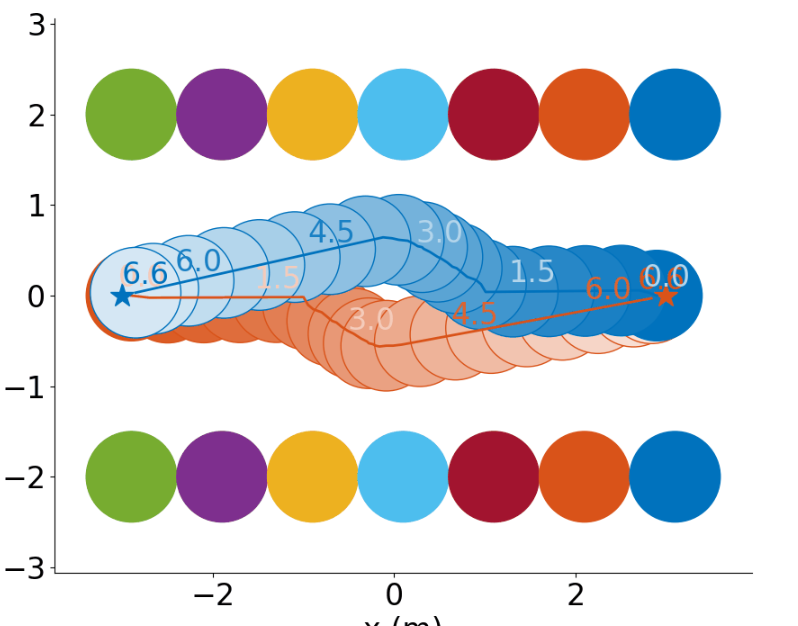}
        \caption{CADRL~\cite{everett2021collision, chen2017socially, chen2017decentralized}}
    \end{subfigure}
    \hfill
    \begin{subfigure}{0.48\linewidth}
        \centering
        \includegraphics[width=\linewidth]{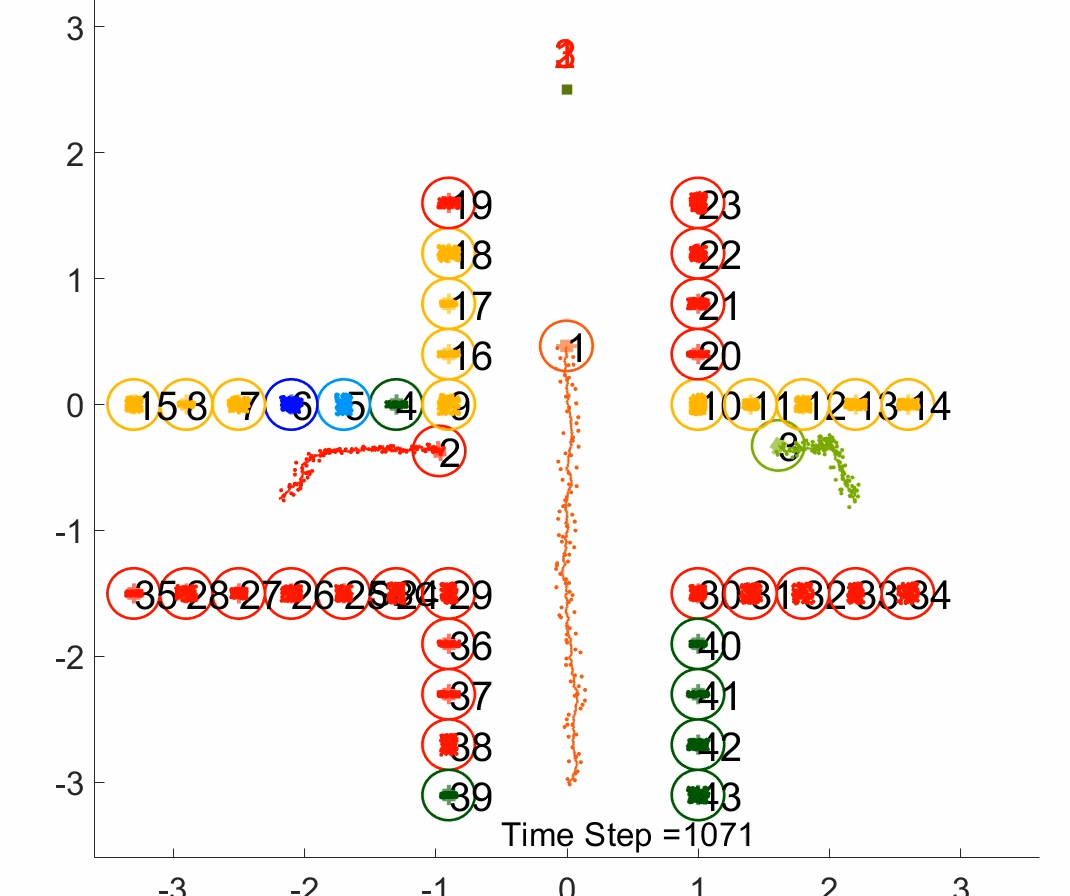}
        \caption{Auction-Based~\cite{chandra2023socialmapf, chandra2022gameplan}}
    \end{subfigure}
    
    \vspace{1ex}
    
    % Row 2
    \begin{subfigure}{0.48\linewidth}
        \centering
        \includegraphics[width=\linewidth]{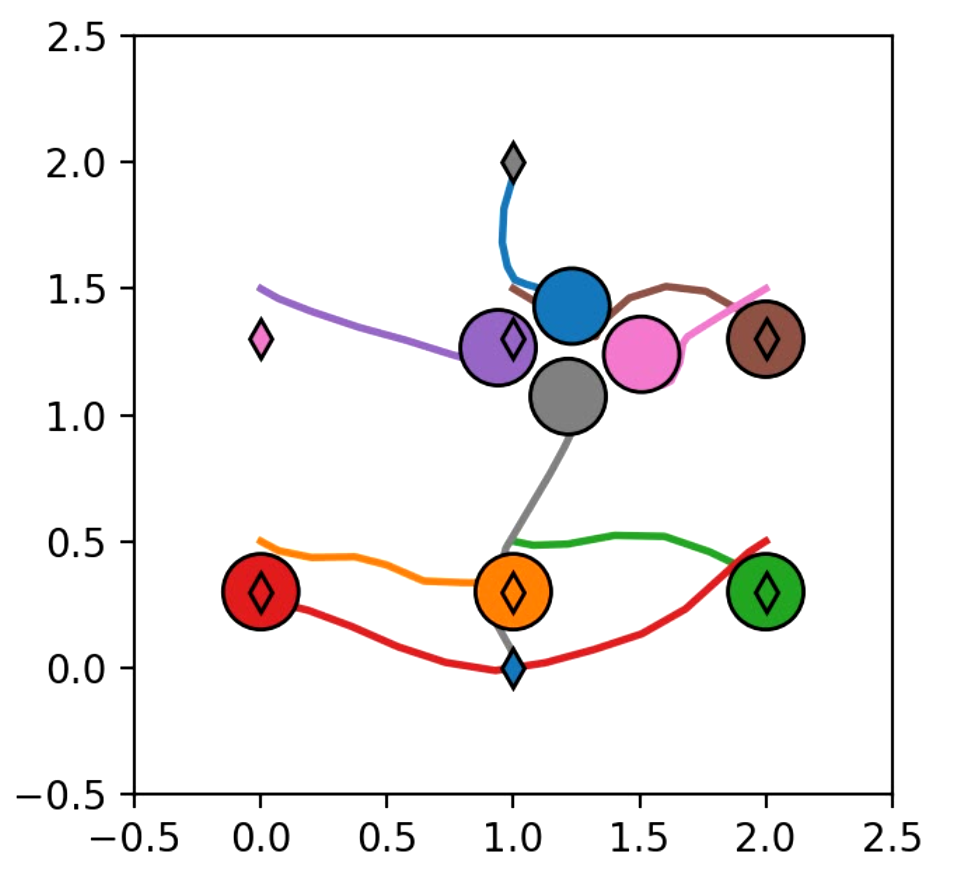}
        \caption{IMPC-DR~\cite{chen2023multi}}
    \end{subfigure}
    \hfill
    \begin{subfigure}{0.48\linewidth}
        \centering
        \includegraphics[width=\linewidth]{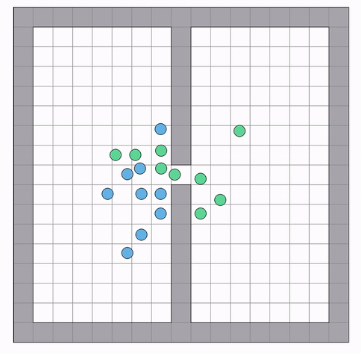}
        \caption{ORCA-MAPF~\cite{dergachev2021distributed}}
    \end{subfigure}
    
    \vspace{1ex}
    
    % Row 3
    \begin{subfigure}{0.48\linewidth}
        \centering
        \includegraphics[width=\linewidth]{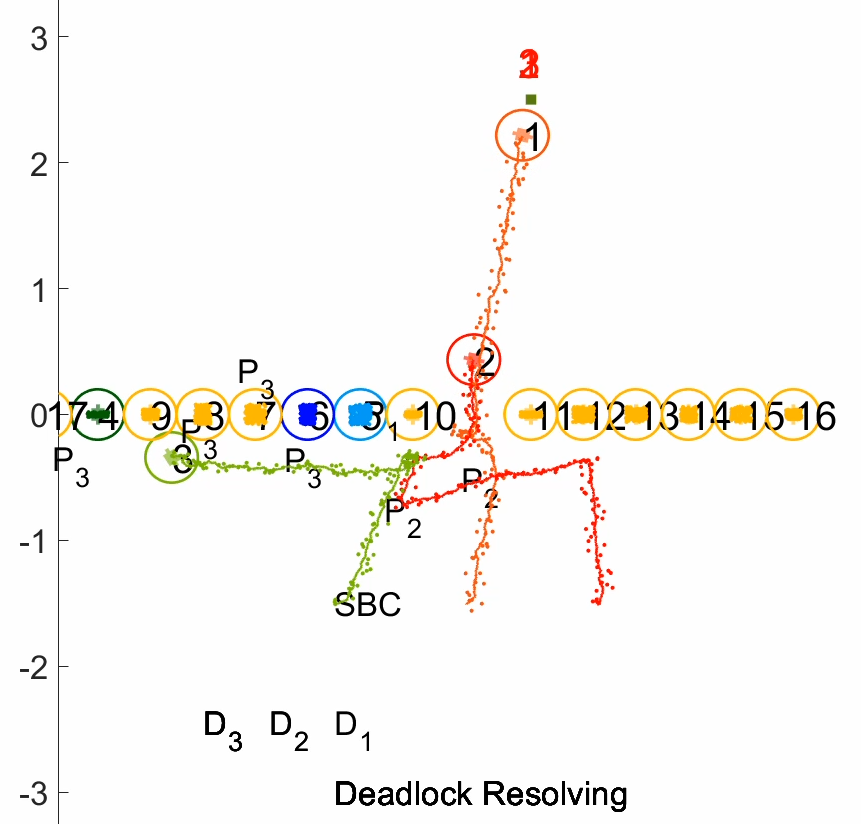}
        \caption{Right Hand Rule~\cite{wang2017safety}}
    \end{subfigure}
    
\caption{Representative SMG solvers in canonical Social Mini-Game scenarios (doorways, hallways, and intersections). 
In all subfigures, agents are shown as colored circles navigating toward their goals (red marker), and the trajectories illustrate how each method resolves deadlock. 
(a) CADRL: a learned reactive collision-avoidance policy that produces smooth local interactions. 
(b) Auction-Based: prioritization via auctioning and velocity scaling, creating a hierarchy where some agents are given priority to improve efficiency and reduce deadlocks in congested scenarios.
(c) IMPC-DR: optimization-based deadlock prevention using constrained MPC. 
(d) ORCA-MAPF: a hybrid approach combining discrete MAPF guidance with continuous ORCA velocity-obstacle avoidance for real-time coordination. 
(e) Right-Hand-Rule: a rule-based clockwise yielding strategy enforced through barrier-function safety constraints, resolving conflicts at the cost of potential fairness violations.}

    \label{fig:methods}
\end{figure}

In this Section, we focus on a few specific approaches that have been been widely tested an adopted as baselines when solving SMGs. We acknowledge that this is an ongoing and emerging research area and this list is by no means exhaustive, but only serves to list the prominent approaches to serve as a starting point for new researches to the field. Figure~\ref{fig:methods} shows these methods graphically with multiple agents in action. In the following sections, we discuss inner-workings, strengths/limitations of some of the techniques, some of which were previously mentioned in Sec.~\ref{sec:Taxonomy}.

\subsection{CADRL \cite{everett2021collision}:}

Collision Avoidance with Deep Reinforcement Learning (CADRL) is a method that combines the fast decision-making of reaction-based approaches with the smooth motion of trajectory-based planning. It uses reinforcement learning to shift heavy computations from real-time execution to an offline training phase. This feature allows the system to learn a collision avoidance policy in advance, making real-time navigation faster and more efficient. Each agent forms an observation from its own pose/velocity + goal and a set of nearby agents’ relative positions, velocities, and sizes/radii. The network outputs a real-time velocity command (often published as a commanded twist / linear+angular speed or planar velocity). A key challenge CADRL addresses is handling a changing number of agents in dynamic environments. To manage this, it uses Long Short-Term Memory (LSTM) networks to encode the varying information about nearby agents into a fixed-size vector. This enables CADRL to make decisions based on any number of surrounding agents without needing a fixed input size. By using LSTM, CADRL not only adapts to dynamic multi-agent scenarios but also retains important past information, leading to better decision-making. However, CADRL-style policies are learned and typically do not provide formal collision-free guarantees. Overall, CADRL provides a scalable and efficient solution for real-time collision avoidance in complex, crowded environments.

\subsection{Right-Hand-Rule (RHS) \cite{wang2017safety, wang2016safety}:}RHS implements perturbation following the right hand rule (clockwise priority) to resolve conflicts among agents in SMGs. The idea behind RHS is to enforce an ordering to agent's movements, avoiding strictly random paths that can lead to deadlocks, especially in crowded or tight spaces. RHS is designed to reduce the chances of deadlocks caused by overly structured and predictable paths. RHS-style “rule-following” can be enforced by safety barrier certificates that compute a control by solving a quadratic program that stays as close as possible to a nominal controller while satisfying pairwise safety constraints. The nature of this method allows agents to respond dynamically to changing scenarios in real time. However, due to the rule enforcement, this method brings some limitations as well. While, on the one hand, it aids in resolving the deadlocks and creating smooth navigation in shared spaces, on the other hand, it can violate fairness constraints resulting in less efficient routes, highlighting a trade-off between goal completion and optimality. 

\subsection{Auction-Based \cite{chandra2022gameplan, chandra2023socialmapf}:}The Auction-Based algorithm combines Control Barrier Functions (CBF) in a auction based manner for prioritizing the agents for motion planning. By creating a competitive environment among agents and carefully adjusting their speeds, the Auction-Based method aims to improve navigation efficiency, even in crowded scenarios. In SocialMAPF~\cite{chandra2023socialmapf}, agents have private incentives and submit bids to an auction program; the allocation/reward/payment rule determines who gets priority during conflict resolution. A key feature of this method is its ability to prioritize agents. Through velocity scaling, it creates a hierarchy where some agents are given priority, leading to more efficient navigation preventing deadlocks, especially in congested areas.  However, while velocity scaling helps in managing priority-based navigation, it can sometimes increase the time it takes for certain agents to reach their destinations. This can potentially lengthen the overall navigation time, reflecting a trade-off in the prioritization scheme used by the Auction-Based method. Despite this trade-off, the method aims for global optimization in path planning, ensuring more orderly and efficient navigation, especially in scenarios with many agents.

\subsection{IMPC-DR \cite{chen2023multi}:}IMPC-DR is a method specifically designed to avoid deadlocks using infinite-horizon Model Predictive Control (MPC). The MPC uses a special structure called modified buffered Voronoi \cite{zhou2017fast} with a warning band. IMPC-DR solves a convex optimization at each replanning step over \textit{modified buffered Voronoi cells} (MBVCs), which partition the workspace among robots such that each robot's feasible region accounts for both inter-robot separation and a configurable warning band of width $\epsilon$.
 This Voronoi structure helps to understand when deadlocks might happen, similar to a state of force equilibrium. When a potential deadlock is detected, the method quickly uses an adaptive resolution scheme to resolve it. This ensures that no stable deadlocks can occur under specific conditions, allowing smooth navigation for the multi-robot system. The planning algorithm in this method ensures that the optimization problem remains feasible at every time step while meeting input and model constraints. It works simultaneously across all robots, promoting coordinated and synchronized navigation. The method relies on local communication, avoiding the need for a central hub or extensive networks, which ultimately reduces the communication burden and improves the scalability and real-time response. Local communication also strengthens the decentralized nature of the method, making it a reliable solution to generate collision-free trajectories in shared spaces. This communication approach systematically prevents and resolves deadlocks in multi-robot navigation.

\subsection{ORCA-MAPF \cite{van2011reciprocal, dergachev2021distributed}:}ORCA-MAPF (Optimal Reciprocal Collision Avoidance - Multi-Agent Path Finding) addresses the challenge of preventing collisions in multi-agent navigation, especially in decentralized settings with limited communication and sensing. In such environments, collision avoidance is usually reactive, based on local observations or sparse communications among agents. While strategies like ORCA are known for their efficiency and scalability, they often struggle in complex scenarios, such as navigating through narrow passages. In these cases, deadlocks are common due to the self-centered behavior of agents, preventing them from reaching their goals. The innovative approach of ORCA-MAPF uses locally confined multi-agent path finding (MAPF) solvers to coordinate sub-groups of agents at risk of deadlock.
In normal operation, each agent uses ORCA to compute collision-free velocities by solving a linear program over half-plane constraints. When a deadlock is detected—via an average-speed buffer dropping below a threshold—agents switch to MAPF mode: they share local states, discretize the conflict zone into a grid-based MAPF instance, and solve it using Push-and-Rotate or ECBS. Because all agents in range construct the same instance from shared information, they reach deterministic agreement without a central controller. After executing their individual MAPF plans, agents revert to ORCA. Fig.~17d illustrates this hybrid behavior: ORCA governs open-space navigation while the MAPF solver resolves bottleneck deadlocks. ORCA-MAPF significantly improves the success rate of navigation in simple or empty shared spaces, increasing the safety rate from 15\% to 99\% in some tests. This improvement highlights ORCA-MAPF's effectiveness in handling deadlocks and enhancing overall navigation success in decentralized multi-agent systems. By integrating MAPF solvers, ORCA-MAPF overcomes the limitations of existing collision avoidance techniques, offering more robust and reliable multi-agent navigation in complex, real-world environments. Through strategic coordination and effective deadlock management, ORCA-MAPF advances the field of multi-agent navigation, especially in decentralized and communication-limited settings.

\section{General Trends \& Open Challenges}
\label{sec:trends}

\subsection{MRN in SMGs with Visual Inputs}

Conventional wisdom~\cite{margolis2021,chandra2024_socialgames,gouru2024,zinage2024} tells us that in order for robots to achieve human-like mobility in cluttered environments, their low-level controllers need exact and accurate state measurements of their surroundings which is difficult in practice to realize, especially if the environment is dynamic. Most roboticists, however, would ideally prefer navigation systems that produce human-like trajectories directly using input from onboard sensors such as lidars and cameras, without relying on expensive mapping and perception for exact state measurements~\cite{desa2024_pointcloud}. For instance, Sa et al.~\cite{desa2024_pointcloud} perform point cloud-based single robot navigation to handle dynamic environments, allowing robots to react to rapidly changing obstacles. There are several challenges to safe and deadlock-free multi-robot navigation in cluttered environments with high-dimensional inputs such as point clouds. The first key challenge is that ensuring both safety and liveness using dense point clouds can be complex \cite{desa2024_pointcloud, guan2022m3detr} and thus, computationally expensive. For instance, some analytical methods use control barrier functions (CBFs)~\cite{desa2024_pointcloud,tong2023_nerf} to guarantee safety, which requires intensive computations to evaluate the barrier function and its derivatives. Additionally, in decentralized systems, we have no central authority that can coordinate agents in a manner that deadlocks will be prevented or resolved, particularly critical when agents are self-interested (each optimizes its own individual objective function) and have conflicting objectives~\cite{chandra2023socialmapf, chandra2022game, chandra2022game, suriyarachchi2022gameopt, suriyarachchi2024gameopt+}. Lastly, learning-based methods~\cite{pointcloudbasedreinforcement,cosner2022_stereovision,dawson2022learningsafegeneralizableperceptionbased,xiao2022differentiablecontrolbarrierfunctions, mavrogiannis2022b} can struggle with generalization and lack formal safety guarantees. 

\subsection{MRN in SMGs with Humans}

Robot navigation in densely populated human environments has garnered significant attention, especially in scenarios involving tightly coupled social interactions among pedestrians, scooters, and vehicles. These scenarios~\cite{chandra2025deadlock} occur in both outdoor settings~\cite{chandra2022game, chandra2022gameplan, suriyarachchi2022gameopt, mavrogiannis2022b} and indoor spaces such as restaurants, grocery stores, hospitals, and university campuses~\cite{scanddataset, raj2024rethinking, sprague2023socialgym, chandra2023socialmapf}. Effective navigation in such dense and cluttered human environments is challenging and requires robots to be socially compliant which involves predicting and responding to human intentions~\cite{poddar2023crowd, francis2025principles}. Many approaches have approached intent prediction \textit{implicitly} via trajectory forecasting~\cite{gupta2018social, sadeghian2019sophie, Kosaraju2019BIGAT, giuliari2020transformer, marchetti2024smemo, sun2021three, ngiam2021scene, yuan2021agentformer, salzmann2023robots, salzmann2020trajectron++, he2021where}. While trajectory forecasting has yielded impressive results on sparse crowd datasets such as ETH, UCY, and JRDB, and structured traffic datasets such as the WAYMO Motion Forecasting Dataset~\cite{waymo}, it is unclear whether they perform similarly well in SMGs that include passing, weaving, yielding to others, etc. Alternatively, some approaches have modeled intent \textit{explicitly} by computing humans' reward functions via inverse reinforcement learning. Recently, Gonon and Billard~\cite{singleirl} used maximum entropy IRL (inverse reinforcement learning)~\cite{maxentirl} for achieving socially compliant navigation in pedestrian crowds. However, inherently a single-agent approach, maximum entropy IRL (MaxEnt IRL) assumes a single objective function for all the agents, which works reasonably well in sparse crowds that lack many tightly coupled interactions, as shown in~\cite{singleirl}, but fails to accommodate humans' objectives in denser and more unstructured pedestrian crowds containing SMGs.

\subsection{MRN in SMGs Using Digital Twins}

To successfully deploy robots in complex human environments such as airports, grocery stores, hospitals, restaurants, and homes, it is essential to ensure social compliance between robots and humans. But in order to effectively train robots to be socially compliant, we require simulation environments that reflect the complexity of the real world--that is, simulators must provide multi-agent learning support, tightly constrained indoor scenarios, realistic robot and human motion models, and lastly, simulators must be configurable and extensible. Current simulators only \textit{partially} satisfy the above requirements~\cite{kastner2022arena, tsoi2022sean, crowdnav,aroor2017mengeros,social_forces,biswas2022socnavbench,crowdbot,holtz2022socialgym}. All of these simulators are currently single-agent navigation in simple open environments. In addition, these simulators model human crowds using reciprocal policies~\cite{van2011reciprocal} or replay stored trajectories from a dataset~\cite{biswas2022socnavbench}, or both. SEAN 2.0~\cite{tsoi2022sean} defines social navigation scenarios via social maneuvers such as crossing, following, and overtaking, but these only apply in open environments, excluding geometrically constrained scenarios. 
Crossing or passing may be impossible and lead to sub-optimal trajectories like colliding with walls when navigating through a narrow doorway, for example. 
Furthermore, while several simulators~\cite{tsoi2022sean,biswas2022socnavbench,crowdbot,sprague2023socialgym} model real-world robot dynamics and kinematics realistically, only two simulators (CrowdBOT and our previous work, SocialGym) allow configurability and extensibility to experiment and benchmark different robot kinodynamic configurations. 

 \section{Conclusion}
\label{sec:conclusion}
We have provided a survey on Multi-robot navigation algorithms in social mini-games for various motion planning methods from different categories. We also provide a benchmark thathelps evaluate how well these methods perform in different scenarios. As a recommendation, in the future, we should focus on developing non-invasive techniques for MRN. Non-invasive methods allow agents to navigate without making significant changes to the environment or other agent's paths. This approach reduces interruptions and helps maintain a smooth movement. Additionally, we should prioritize deadlock prevention over resolution. Preventing deadlocks before they happen is more effective than resolving them once they occur. By focusing on strategies that avoid deadlocks from the start, we can ensure more efficient and reliable navigation for multi-agent systems.

\bibliography{ref}% common bib file
\bibliographystyle{ieeetr}
\clearpage
\appendix
\section{SMGLib}
\label{sec:impl}

SMGLib is a simulation environment designed to connect theoretical analysis with practical applications for MRN in SMGs. SMGLib provides access to run diverse simulations for existing path planners in SMG scenarios, analyses the results, and compare with other motion planners to know the efficacy of these planners on the given benchmark scenarios (sec~\ref{sec:benchmark_scenarios}). The library includes a variety of real-world scenarios imitating the social navigation, including both static and dynamic scenarios. 
% The details of the usage of SMGLib are given in Appendix~\ref{sec:impl}.

While SMGLib is an extensible library with work in progress, in this subsection, we present the comparative analysis for the benchmarking of some MRNs in terms of their safety, smoothness, invasiveness, deadlock handling capability, and efficiency in SMGs. The analysis is done with two-agent setup in a static environment, which includes the scenarios Doorway, Intersection, and Hallway. The comparison is carried out using five evaluation metrics (sec.~\ref{sec:evaluation_metrics}) (Average $\triangle V$, Makespan Ratio, Path Deviation, Success-Rate, and Flow-rate), which gives an idea about the effectiveness of each MRN.
% Note that each scenario is simulated considering each method, but the comparison of metrics in Table~\ref{tab:compare} is only for the cases where the success rate is 100\%.

\textbf{\textit{SMGLib Design}}: The design of SMGLib includes a command-line interface, which makes it easy to test different scenarios for different motion-planners, quickly. The library is also extensible, thus allowing the researchers to adapt the library to their specific needs without the usual complexity. SMGLib offers interactive elements for defining the simulation space and agent characteristics. The user can select the number of agents and their state parameters, the motion-planner, type of SMG to test on, the evaluation capabilities to save the log, evaluation metrics and generate the plots. 
% Table~\ref{tab:method_option} gives details of these options with the supported values. 
\begin{table}[t]
\centering
\caption{SMGLib Options}
\footnotesize
\renewcommand{\arraystretch}{1.1}
\begin{tabular}{|>{\raggedright\arraybackslash}p{1.8cm}|>{\raggedright\arraybackslash}p{4.2cm}|}
\hline
\textbf{Option} & \textbf{Description \& Supported values} \\
\hline
Method & Specifies the navigation method: \\
& \texttt{social-orca} -- ORCA with social context integration \\
& \texttt{social-impc-dr} -- socially-aware iMPC with deadlock resolution \\
& \texttt{CADRL} -- Collision Avoidance with Deep RL \\
\hline
Environment & Scenario type: \texttt{doorway}, \texttt{hallway}, \texttt{intersection} \\
\hline
Number of Robots & Number of agents simulated per run (typically 1--4) \\
\hline
Agent Configuration & User-defined start and goal positions within a 2D grid, \(x \in [0, X_{\text{pos}}]\), \(y \in [0, Y_{\text{pos}}]\) \\
\hline
\end{tabular}
\label{tab:method_option}
\end{table}

\\
\textbf{\textit{Working Example}}: 
To demonstrate SMGLib's working, we present a scenario using the Social-ORCA algorithm. In our example, the user starts by setting up the simulation with a given MRN algorithm (Social-ORCA in Fig.~\ref{fig:terminal_method_selection}). The GUI then asks the user to select the benchmark scenario (e.g. doorway, hallway, intersection), for which the user selected option-1 as shown in Fig.~\ref{fig:terminal_env_selection}. Finally, the terminal asks the user to input the number of agents, followed by initial and target positions for each agent on a 2D grid (Fig.~\ref{fig:doorway_config_three_robots}). As the simulation runs, SMGLib renders the agent's trajectories in real-time.  Figures~\ref{fig:smglib_phases} (a-e) show an example of three agents navigating the doorway, thus demonstrating Social ORCA's collision avoidance capability, efficiency in path-finding, and deadlock avoidance. The agents, shown as dynamic entities that adjust their paths in response to the other agents, reach their respective goals at the end of the simulation run. Frame c shows an example of a close encounter, but due to the efficacy and responsiveness of the algorithm for handling deadlocks, the agents finish reaching their respective goals. 
% SMG main window- method selection
\begin{figure}[t]
    \centering
    \begin{terminalbox}
Welcome to the Multi-Agent Navigation Simulator\\
==============================================\\ \\
Available Methods:\\
1. Social-ORCA\\
2. Social-IMPC-DR\\ \\ 
Enter method number (1-2): 1
    \end{terminalbox}
    \caption{Terminal interaction showing the selection of a MRN.}
    \label{fig:terminal_method_selection}
\end{figure}
%
% SMG main window- method selection
\begin{figure}[t]
    \centering
    \begin{terminalbox}
Available environments: \\
1. doorway \\
2. hallway \\
3. intersection \\
\\
Enter environment type (1-3): 1
    \end{terminalbox}
    \caption{Terminal interaction for selecting the scenario in the simulator.}
    \label{fig:terminal_env_selection}
\end{figure}
%
% Terminal for inputting the robot position
%
\begin{figure}[t]
    \centering
    \begin{adjustbox}{max width=\textwidth}
    \begin{terminalbox}
Enter number of robots (1-4): 3 \\ \\

Doorway Configuration: \\ 
- The doorway has walls at x=30-31 with a gap at y=30-34 \\
- Y coordinates should be between 0 and 63 \\
- X coordinates should be between 0 and 63 \\ \\ 

Robot 1 configuration:\\ 
Enter start X position (0-63) for robot 1: 15 \\
Enter start Y position (0-63) for robot 1: 32 \\ 
Enter goal X position (0-63) for robot 1: 45 \\ 
Enter goal Y position (0-63) for robot 1: 32 \\ 
Robot 1 will move from (15.0, 32.0) to (45.0, 32.0) \\ \\ 

Robot 2 configuration: \\ 
...
    \end{terminalbox}
    \end{adjustbox}
    \caption{Terminal user-interaction for inputting robot configurations}
    \label{fig:doorway_config_three_robots}
\end{figure}
\textbf{\textit{Post-simulation}}: The user can use SMGLib's analytical tools to examine the performance of planner. Metrics of interest include success rate, makespan ratio, path deviation, and average $\Delta V$. After the simulation ends, a detailed visualization of the robot's paths is generated, complete with performance data. This gives a comprehensive idea about the algorithm executed strategies for navigating within the SMG. This integration of statistics shows SMGLib's use as a tool for researchers aiming to experiment and validate navigation algorithms, thus finding potential gaps for improving the algorithms. The library's combination of detailed simulation control, robust analytical tools, visualization capabilities makes it a handy tool for evaluating deadlock capabilities in social mini-games for multi-agent navigation, Figure~\ref{fig:doorway_log_sample_robot0} shows the result of the simulation for the doorway example for the first robot.
%
% Terminal post simulation
\begin{figure}[t]
    \centering
    \scriptsize
    \begin{adjustbox}{max width=\textwidth}
    \begin{terminalbox}
Configuration saved to config\_doorway\_3\_robots.xml \\

Running Social-ORCA Simulation \\
============================= \\

Using configuration file: config\_doorway\_3\_robots.xml \\

Running simulation with 3 robots... \\

Log file generated: logs/config\_doorway\_3\_robots\_log.xml \\
Trajectory CSV files generated in: logs/trajectories \\
Animation saved to: animations/robot\_movement.gif \\

Evaluating trajectories... \\

Evaluating Robot 0 trajectory: \\
************************************************ \\
Robot 0 Path Deviation Metrics: \\
L2 Norm: 7836.3800 \\
Hausdorff distance: 1.2077 \\
************************************************ \\
************************************************ \\
Robot 0 Avg delta velocity: 1.1085 \\
************************************************
    \end{terminalbox}
    \end{adjustbox}
    \caption{Sample log output for Robot-1 for the 3-robot doorway simulation shows the Path deviation and Average $\Delta$ V (Sec.~\ref{sec:evaluation_metrics})}
    \label{fig:doorway_log_sample_robot0}
\end{figure}
\begin{figure}[t]
  \centering
  \setlength{\tabcolsep}{3pt}

  % Row 1
  \begin{tabular}{@{}c c@{}}
    \subcaptionbox{}{\includegraphics[width=0.48\columnwidth]{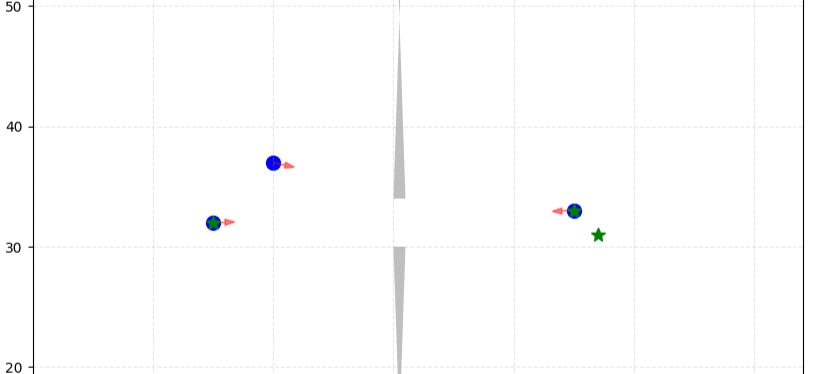}\label{fig:smglib_frame1}} &
    \subcaptionbox{}{\includegraphics[width=0.48\columnwidth]{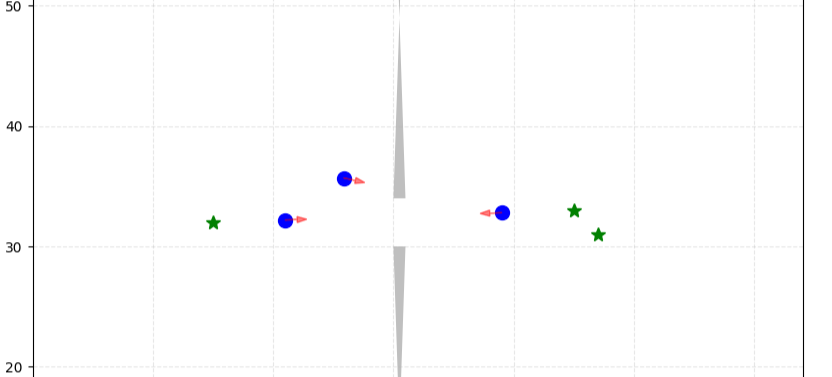}\label{fig:smglib_frame2}}
  \end{tabular}

  % Row 2
  \begin{tabular}{@{}c c@{}}
    \subcaptionbox{}{\includegraphics[width=0.48\columnwidth]{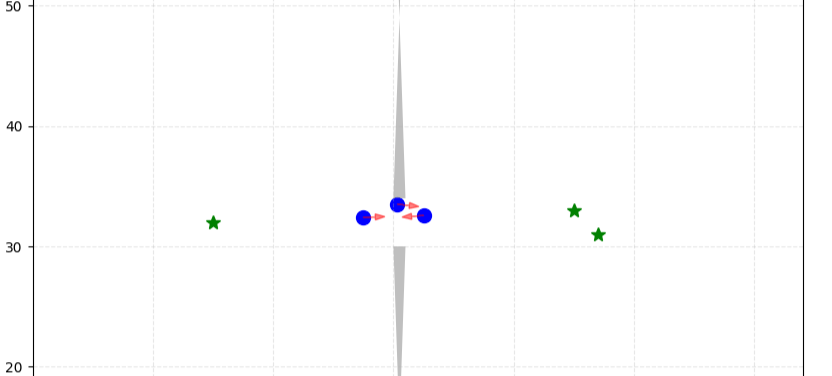}\label{fig:smglib_frame3}} &
    \subcaptionbox{}{\includegraphics[width=0.48\columnwidth]{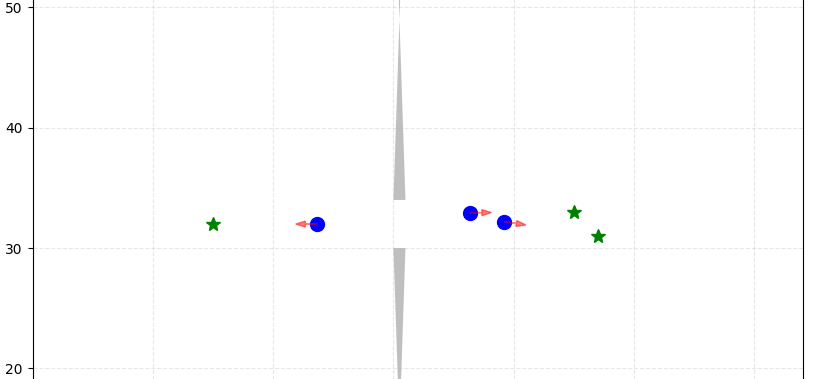}\label{fig:smglib_frame4}}
  \end{tabular}

  % Row 3 (single, centered)
  \subcaptionbox{}{\includegraphics[width=0.48\columnwidth]{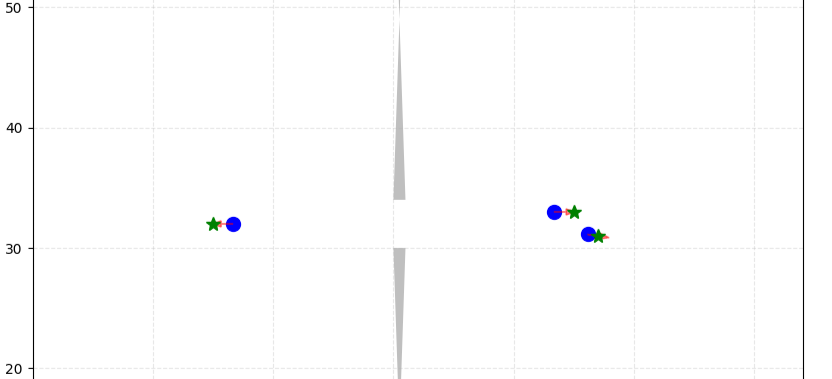}\label{fig:smglib_frame5}}

  \caption{Selected keyframes from the SMGlib simulation of a 3-robot doorway scenario showing progressive stages of multi-robot navigation. The green star denotes the respective goals of each of the robots.}
  \label{fig:smglib_phases}
\end{figure}

\end{document}